\documentclass{article}
\usepackage{iclr2027_conference,times}

\iclrfinalcopy

\usepackage[utf8]{inputenc} 
\usepackage[T1]{fontenc}    
\usepackage{hyperref}       
\hypersetup{hidelinks,pdfauthor={}} 
\usepackage{url}            
\usepackage{booktabs}       
\usepackage{amsfonts}       
\usepackage{nicefrac}       
\usepackage{microtype}      
\usepackage{xcolor}         

\usepackage{amsmath}
\usepackage{amssymb}
\usepackage{mathtools}
\usepackage{amsthm}
\usepackage{thm-restate}
\usepackage[capitalize,noabbrev]{cleveref}

\theoremstyle{plain}
\newtheorem{theorem}{Theorem}[section]
\newtheorem{proposition}[theorem]{Proposition}
\newtheorem{lemma}[theorem]{Lemma}
\newtheorem{corollary}[theorem]{Corollary}

\theoremstyle{definition}
\newtheorem{definition}[theorem]{Definition}

\newtheorem{example}[theorem]{Example}
\theoremstyle{remark}
\newtheorem{remark}[theorem]{Remark}

\title{Universal interpolation for deep residual self-attention networks}

\author{%
  Sibylle Marcotte
  \\
  Department of Computer Science\\
  New York University\\
  \texttt{slm9693@nyu.edu} \\
  \And
  Joan Bruna \\
  Department of Computer Science\\
  New York University\\
}

\begin{document}

\maketitle

\begin{abstract}
Universal approximation is a necessary qualitative property of learning architectures to benefit from scaling laws. 
While it is generically verified on a variety of neural architectures and random feature models, it typically involves infinite width limits. 
In this work, we focus on deep self-attention models and consider instead the `dual' regime, where  approximation power is enabled entirely by depth, and featuring strong parameter sharing across layers, motivated by recent models such as the Looped Transformers. More specifically, we ask whether one can find a predefined finite set of parameters, each defining an attention block, such that the resulting finite set of transformations can map any collection of $N$ sequences of $n$ tokens to any other collection of $N$ sequences of $n$ tokens. Crucially, these transformations are \emph{fixed independently of the input and output} collections: only the order in which the blocks are applied, their signs, and their durations depend on the particular interpolation task.
Our main result establishes it for residual softmax attention using only two frozen single-head blocks with Gaussian-initialized projection matrices. The result holds at both continuous and finite depth. We also characterize the restrictions imposed by causal masking and establish corresponding universal interpolation guarantees.
\end{abstract}

\section{Introduction}
Introduced by \citet{vaswani2017attention}, transformers have driven major advances in natural language processing~\citep{devlin2019bert}, computer vision~\citep{dosovitskiy2021image} and reasoning~\citep{wei2022chain}. This widespread success motivates several key questions regarding their architecture, including their expressive power, their `inductive bias' or feature-learning abilities, and ultimately their ability to generalize in the current pretrain/post-train paradigm.

Towards this long-term goal, the first stepping stone is to characterize a flexible class of transformer architectures with enough expressive power to interpolate any given dataset, as a basis for studying scaling laws. 
A standard approach to studying expressivity in neural architectures is to consider infinitely wide limits, starting with the classic universal approximation theory of \cite{cybenko1989approximation}, \citet{hornik1989multilayer} and others in the single-hidden layer setting. In this context, the design of simple hypothesis spaces (e.g. random feature models) with universal approximation has enabled the subsequent study of non-linear feature learning.
Given the prevalent role of depth in the success of recent deep learning models, it is natural to ask whether other mechanisms for universal approximation exist where depth is the instrumental variable; in other words, to consider the deep analogs of random feature models. 

In this work we focus on transformer models, and specifically their self-attention modules. Motivated by recent interest on looped transformers, we make an additional distinction: while traditional memorization asks whether, for any given task, it is possible to adjust the parameters of the architecture to produce prescribed outputs, we are interested in capturing the notion of parameter reuse. Specifically, we ask whether it is possible to predefine a finite family of attention blocks \emph{a priori}, without looking at the data, and then achieve exact interpolation only by choosing which block, and with which scalar gain, to use at each layer.
Following the fixed-generator formulations
in~\citet{cuchiero2020deep,agrachev2026generic}, we call this property \emph{universal interpolation}.

These questions are particularly relevant to recent Transformer architectures that reuse a small number of blocks across depth. Such recurrent-depth or looped architectures have recently been explored at scale in language models, including recurrent-depth models~\citep{geiping2025scaling}, Mixture-of-Recursions~\citep{bae2025mixture}, Nanbeige4.2-3B~\citep{nanbeige2026nanbeige}, and Hyperloop Transformers~\citep{zeitoun2026hyperloop}. Rather than introducing independent parameters at every layer, these models obtain additional effective depth by repeatedly applying the same transformations. This makes it natural to ask how much expressive power can arise from depth and composition alone when the available transformations are fixed.

\textbf{Exact interpolation.}
A large body of work studies expressivity of neural networks by asking whether, for each finite dataset, network parameters can be chosen to match the prescribed outputs exactly. Such results characterize the fitting capacity of an architecture, while allowing all of its parameters to depend on the dataset. 
 \citet{ruizbalet2023neural} establish exact interpolation of finite datasets of distinct points by ReLU neural ODEs in dimension at least two, using data-dependent controls and biases; one active hidden unit suffices.  
For transformers, \citet{kajitsuka2024transformers} construct contextual representations with one attention head and rank-one projections, followed by a feed-forward decoder whose size grows with the data. \citet{dana2025memorization} prove next-token memorization for attention-only models with constructed embeddings and a readout, with capacity controlled by the number and dimension of heads. \citet{alcalde2026exact} obtain exact interpolation of output token sets, without prescribed token correspondences, using attention and ReLU feed-forward blocks with explicit depth bounds. \citet{geshkovski2026measure} instead study interpolation of token distributions on the sphere using attention, perceptrons, and spherical layer normalization. In all of these settings, the transformations used by the network may be adapted to the particular interpolation problem.

\textbf{Universal approximation.}
A related notion of expressivity is universal approximation, which asks whether a network class can approximate every mapping in a prescribed function class to arbitrary accuracy, under an appropriate topology. Here too, the network parameters may depend on the target function; universality concerns the network class rather than a fixed collection of transformations. Classical results establish universal approximation for sufficiently wide networks with a single hidden layer and a continuous sigmoidal activation \citep[e.g.,][]{cybenko1989approximation,hornik1989multilayer}. \citet{lin2018resnet} establish \(L^1\) universal approximation for deep ReLU residual networks with one-neuron hidden layers and target-dependent weights and biases. From a control-theoretic perspective, \citet{tabuada2021universal} derive approximation guarantees through ensemble controllability and monotonicity. For transformers, \citet{yun2020transformers} establish $L^p$ approximation of continuous permutation-equivariant mappings with compact support at fixed sequence length, using two attention heads of width one and feed-forward blocks of width four. \citet{furuya2025transformers} obtain uniform approximation of continuous in-context mappings on compact token domains, with token dimension and head count bounded independently of context length and accuracy while retaining tokenwise MLPs. \citet{hu2026universal} establish approximation using two attention layers supplemented with linear transformations. These results address approximation of function classes rather than exact interpolation under a fixed set of reusable transformations.

\textbf{Universal interpolation with fixed generators.}
Beyond classical exact interpolation, one can ask whether the transformations available to the network can be fixed in advance and reused across datasets. \citet{cuchiero2020deep} construct five fixed vector fields that interpolate every finite dataset of distinct points in a bounded connected domain of dimension at least two, with only the scalar controls depending on the data. Related fixed-family interpolation results have been obtained for general neural ODEs~\citep{grong2026controllability} and for smooth-activation residual networks~\citep{tabuada2023universal}. 
For ReLU dynamics, \citet{cai2025achieving} establish
exact interpolation using arbitrary affine vector fields together
with a fixed ReLU field. Combining their result with an exact
factorization of affine flow maps into elementary linear flows
and translations yields a finite family of fixed generators,
independent of the dataset and its size, with biases allowed. 
Closest to our work, \citet{agrachev2026generic} prove that two generic equivariant vector fields suffice for simultaneous interpolation of arbitrarily many compatible point clouds. Their genericity is formulated over an infinite-dimensional space of vector fields, and they explicitly leave extensions to standard softmax attention as an open direction. Establishing such a result is one focus of our work.

\textbf{Our contributions.}
We formalize exact and universal interpolation
(\Cref{def:exact-interpolation,def:universal-interpolation}), two related
notions used in the literature, and give simple criteria that follow
directly from classical theorems in control theory
(\Cref{sec:interpolation-criteria}).  We also provide a \texttt{PyTorch}
implementation to investigate these criteria numerically
(\Cref{sec:computing-lie-rank}).  Our main contributions are:
\begin{itemize}
    \item to transfer exact (resp.\ universal) interpolation from
    continuous to finite depth under the corresponding Lie-rank
    conditions (\Cref{prop:finite-depth-local-lie-rank-proposal}, resp.\
    \Cref{prop:universal-finite-depth-lie-rank});
    \item to establish universal interpolation with frozen random
    noncausal attention blocks for any prescribed number of heads,
    using a number of blocks that grows linearly with the batch size
    (\Cref{thm:attention-uniform-finite-generation});
    \item to show that only two frozen random single-head noncausal
    attention blocks suffice for universal interpolation on a connected
    domain of full measure (\Cref{thm:two-gaussian-connected-domain});
    \item to rule out exact and universal interpolation on the full
    configuration space for causal attention when the batch size exceeds
    the token dimension, by characterizing the exact dimension of the
    associated Lie-algebra trace (\Cref{prop:causal-rank});
    \item to recover universal interpolation for causal attention on
    each connected component of a class of batches sharing the same
    linear relations among their first tokens
    (\Cref{cor:causal-random-dictionary}).
\end{itemize}

\section{Exact interpolation and universal interpolation}\label{sec:interpolation-framework}

In this section, we formulate exact and universal interpolation in continuous depth and show how the resulting criteria transfer to finite-depth residual networks. We begin by recalling the standard continuous-depth formulation of residual architectures.

\subsection{From residual layers to continuous-depth networks}
Given a partition $0=s_0<\cdots<s_L=: S$, a residual layer takes the form
\[
x_{k+1}=x_k+h_k f_{\theta_k}(x_k),
\qquad h_k=s_{k+1}-s_k.
\]
Taking $\theta(s)=\theta_k$ on $[s_k,s_{k+1})$, this update is a forward Euler discretization of the continuous-depth dynamics~\citep{lu2018beyond,haber2018stable,chen2018neural}
\begin{equation}\label{eq}
\frac{d x(s)}{ds}=f_{\theta(s)}(x(s)),
\qquad x(0)=x_0,
\end{equation}
where $s\in[0,S]$ represents continuous depth and the parameters $\theta(s)$ may vary with depth but are shared across inputs.

For self-attention, we represent each sequence by its ordered token
coordinates $x=\operatorname{vec}(x_1,\ldots,x_n)\in\mathbb{R}^{nd}$
and use the finite-dimensional ODE formulation~\eqref{eq}.
This representation retains the prescribed tokenwise input--output
correspondences, which are not encoded by an empirical-measure
description~\citep{sander2022sinkformers}.

We consider two classes of vector fields, corresponding to residual ReLU networks and residual self-attention networks.

\begin{example}[Continuous-depth ResNets] \label{ex:resnet}
 A bias-free residual block with one ReLU hidden layer has update
 \begin{equation} \label{eq:vectorfields_ResNets}
f_{U, V}: x \in \mathbb{R}^d \mapsto U \sigma (V^\top x) \in \mathbb{R}^d,
\end{equation}
where $\sigma(t)\coloneqq\max\{t,0\}$ is applied componentwise and $U,V\in\mathbb{R}^{d\times r}$, with $r$ the hidden width.
\end{example}

\begin{example}[Continuous-depth self-attention]\label{ex:attlayers}
For $x=\operatorname{vec}(x_1,\ldots,x_n)\in\mathbb R^{nd}$,
let $\mathcal J=(J_i)_{i=1}^n$ specify the visible key positions.
An $H$-head block updates token $i$ by
\begin{equation}\label{eq:vectorfields_att}
    [f^{\mathcal J}_{\mathbf W,\mathbf V}(x)]_i
    \coloneqq
    \sum_{h=1}^H
    \frac{\sum_{j\in J_i}e^{\langle W_hx_i,x_j\rangle}V_hx_j}
         {\sum_{q\in J_i}e^{\langle W_hx_i,x_q\rangle}},
    \qquad i=1,\ldots,n.
\end{equation}
Here $\mathbf W=(W_h)_{h=1}^H$ and $\mathbf V=(V_h)_{h=1}^H$,
with the standard parametrization~\citep{vaswani2017attention}
\[
    W_h=\frac{(W_h^K)^\top W_h^Q}{\sqrt{r_h}},
    \qquad V_h=W_h^O W_h^V,
\]
where $W_h^Q,W_h^K,W_h^V\in\mathbb R^{r_h\times d}$ and
$W_h^O\in\mathbb R^{d\times r_h}$.
The head widths $r_h$ are arbitrary positive integers,
including the standard choice $r_h=d/H$ when $H$ divides $d$.
We write $f_{\mathbf W,\mathbf V}$ for noncausal attention
($J_i=\{1,\ldots,n\}$) and $f^{\mathrm c}_{\mathbf W,\mathbf V}$
for causal attention ($J_i=\{1,\ldots,i\}$).
\end{example}

\subsection{Exact and universal interpolation}

Let $\mathcal{F}=\{f_\theta:\mathbb{R}^p\to\mathbb{R}^p: \theta\in\Theta\}$ be the family of available updates, with each $f_\theta$ locally Lipschitz. 
We represent a dataset of $N$ samples by its vectorisation
$X=\operatorname{vec}(x^1,\ldots,x^N)\in\mathbb{R}^{pN}$,
where $(x^1,\ldots,x^N)$ is an ordered batch. 
For residual ReLU networks, each sample is a vector in \(\mathbb R^d\), so \(p=d\). For self-attention, each sample is a sequence of \(n\) tokens vectorized in \(\mathbb R^{nd}\), so \(p=nd\) and \(N\) counts sequences rather than tokens.  Applying
the same weights to the whole batch gives the \emph{lifted} update
\[
    f^{(N)}_\theta(X)
    \coloneqq \bigl(f_\theta(x^1),\ldots,f_\theta(x^N)\bigr).
\]
A depth-dependent choice of parameters $\theta(\cdot)$ is called a \emph{control}; throughout, controls are piecewise constant with finitely many depth intervals. 

Architectural symmetries and preserved relations constrain which
input--output pairs can be interpolated. We study interpolation
on a chosen configuration domain
$\mathcal{M}_N\subseteq \mathbb{R}^{pN}$, specified for each architecture
below. Both input and target configurations must lie in this domain,
with targets represented in the same space as the inputs.

\begin{definition}[Exact interpolation]\label{def:exact-interpolation}
    Fix $N\geq1$ and a nonempty, open, connected set
    $\mathcal{M}_N\subseteq \mathbb{R}^{pN}$.  For $S>0$, a piecewise-constant
    control $\theta:[0,S]\to\Theta$ with finitely many depth intervals
    \emph{exactly interpolates} a pair
    $X,Y\in\mathcal{M}_N$ if the solution of
    \[
        \dot Z(s)=f^{(N)}_{\theta(s)}(Z(s)),\qquad Z(0)=X,
    \]
    remains in $\mathcal{M}_N$ and satisfies $Z(S)=Y$.  We say that
    $\mathcal{F}$ has the \emph{exact interpolation property on
    $\mathcal{M}_N$} if every pair $X,Y\in\mathcal{M}_N$ admits such
    a terminal depth $S>0$ and a control.
\end{definition}

Universal interpolation additionally requires a finite dictionary of
updates to be fixed before the input--target dataset is given.
For the next definition, the architecture must allow the negative of
each selected update.  Both architectures above have this property by
negating the output matrices.

\begin{definition}[Universal interpolation]\label{def:universal-interpolation}
    Fix $N\geq1$ and a nonempty, open, connected set
    $\mathcal{M}_N\subseteq \mathbb{R}^{pN}$.  The architecture described
    by $\mathcal{F}$ has \emph{universal interpolation on
    $\mathcal{M}_N$} if there exist an integer $m\geq1$ and parameters
    $\theta_1,\ldots,\theta_m\in\Theta$ such that, for every
    $X,Y\in\mathcal{M}_N$, there exist a terminal depth $S>0$ and
    a piecewise-constant control
    \[
        g:[0,S]\to\mathcal{D}_N,\qquad
        \mathcal{D}_N\coloneqq\{\pm f_{\theta_j}^{(N)}:1\leq j\leq m\},
    \]
    with finitely many depth intervals, such that
    \[
        \dot Z(s)=g(s)(Z(s)),
        \qquad Z(0)=X,\qquad Z(S)=Y,
    \]
    with $Z(s)\in\mathcal{M}_N$ for all $s\in[0,S]$.
\end{definition}

For fixed $N$, the parameters are chosen once for all admissible
input--target datasets.  Fitting a dataset changes only the selected
signed updates, their order, and their durations, shared across all samples;
in particular, $S$ may depend on $X$ and $Y$.
The selected parameters and their number may depend on $N$, while $p$
and the architecture remain fixed.  We count the $m$ base fields;
their lifts and negatives form $\mathcal{D}_N$ without additional
parameter choices or random draws.

Our use of \emph{universal interpolation} follows the fixed-vector-field
formulations in~\citet{cuchiero2020deep,agrachev2026generic,grong2026controllability}.
Elsewhere, this term also denotes exact interpolation for every $N$
without requiring a fixed finite dictionary~\citep{cheng2025interpolation,cai2025achieving}.

\subsubsection{Criteria for exact and universal interpolation}\label{sec:interpolation-criteria}

The rank criteria below ask whether the lifted updates and their iterated
Lie brackets span the entire batch space, of dimension $D=Np$.
Composing flows in different orders can produce directions outside
the span of the instantaneous updates. For smooth fields, this
effect is captured by the Lie bracket
\[
    [g,h](X)\coloneqq Dh(X)g(X)-Dg(X)h(X),
\]
where $Dg$ denotes the Jacobian.  For a family $\mathcal{G}$ of smooth
fields on an open set $\mathcal{U}$, the Lie algebra
$\operatorname{Lie}(\mathcal{G})$ is the real linear span of its fields
and their iterated brackets.  Its \emph{trace} at $X\in\mathcal{U}$ is
\[
    \operatorname{Lie}(\mathcal{G})(X)
    \coloneqq
    \operatorname{span}\{h(X):h\in\operatorname{Lie}(\mathcal{G})\}.
\]

We use the control formulation of the classical Chow--Rashevskii
criterion~\citep[Theorem~5.9 and Corollary~5.11]{agrachev2004control},
a consequence of the orbit theorem~\citep{sussmann1973orbits}.
Closure under negation allows flow segments to be traversed in both
directions, with no prescribed terminal depth.

\begin{theorem}[Chow--Rashevskii, control formulation]\label{thm:chow-rashevskii}
    Let $D\geq1$, let $\mathcal{M}\subseteq\mathbb{R}^D$ be nonempty,
    open, and connected, and let $\mathcal{G}$ be a family of smooth
    vector fields on $\mathcal{M}$ closed under negation.  If
    \[
        \operatorname{Lie}(\mathcal{G})(X)=\mathbb{R}^D
        \qquad\text{for every }X\in\mathcal{M},
    \]
    then, for every $X,Y\in\mathcal{M}$, there exist $S>0$ and a
    piecewise-constant control $g:[0,S]\to\mathcal{G}$ with finitely
    many intervals such that
    \[
        \dot Z(s)=g(s)(Z(s)),\qquad Z(0)=X,\qquad Z(S)=Y,
    \]
    with $Z(s)\in\mathcal{M}$ for all $s\in[0,S]$.
\end{theorem}

Full rank makes every nearby configuration reachable.
Connectedness then extends this local property to the whole domain.
The two corollaries below (see \Cref{app:exact-rank-proof,app:universal-rank-proof} for the proofs) apply this theorem to all available updates
or to a fixed finite selection together with its negatives.  In both,
we assume that $\mathcal{F}$ is closed under negation.
To include ReLU updates, we only require the fields selected at each
configuration to be smooth in a neighborhood of that configuration.
In the universal criterion, the index set $I_X$ selects these fields
from the fixed dictionary and may vary with $X$.

\begin{corollary}[Criterion for exact interpolation]
\label{cor:exact-interpolation-criterion}
    Fix $N\geq1$ and a nonempty, open, connected configuration space
    $\mathcal{M}_N\subseteq \mathbb{R}^{pN}$. Suppose that, for every
    $X\in\mathcal{M}_N$, there are finitely many parameters
    $\theta_1,\ldots,\theta_m$ whose lifted fields are smooth near $X$
    and satisfy
    \[
        \operatorname{Lie}\bigl(f_{\theta_1}^{(N)},\ldots,
        f_{\theta_m}^{(N)}\bigr)(X)=\mathbb{R}^{Np}.
    \]
    Then $\mathcal{F}$ has the exact interpolation property on
    $\mathcal{M}_N$ in the sense of \Cref{def:exact-interpolation}.
\end{corollary}

\begin{corollary}[Criterion for universal interpolation]
\label{cor:universal-interpolation-criterion}
    Fix $N\geq1$ and a configuration space $\mathcal{M}_N$ as in
    \Cref{def:universal-interpolation}.  Suppose there exist finitely
    many parameters $\theta_1,\ldots,\theta_m$
    such that, at each $X\in\mathcal{M}_N$, an index set
    $I_X\subseteq\{1,\ldots,m\}$ selects lifted fields smooth near
    $X$ with
    \[
        \operatorname{Lie}\bigl(f_{\theta_j}^{(N)}:j\in I_X\bigr)(X)
        =\mathbb{R}^{Np}.
    \]
    Then $\mathcal{F}$ has universal interpolation on
    $\mathcal{M}_N$ in the sense of
    \Cref{def:universal-interpolation}.
\end{corollary}

In the exact criterion, the parameters may depend on the configuration
$X$.  In the universal criterion, the same finite parameter set must work
for every $X\in\mathcal{M}_N$. For globally smooth updates, one can take
$I_X=\{1,\ldots,m\}$.

\begin{remark}[Analytic converse]\label{rem:analytic-orbit}
For real-analytic fields, the rank conditions above are also
necessary, by the analytic orbit theorem%
~\citep[Corollary~5.17]{agrachev2004control}.
This applies in particular to linear networks and softmax attention.
For merely smooth fields, the conditions remain sufficient but
need not be necessary. See \Cref{app:analytic-orbit-proof}.
\end{remark}

A stronger sufficient condition is that the update fields themselves
span $\mathbb{R}^{Np}$ at every configuration, without using Lie
brackets. This direct spanning condition only requires local $C^1$
regularity of the selected fields; see
\Cref{cor:pointwise-full-rank-criterion} in
\Cref{app:pointwise-rank-proof}.
For a fixed dictionary, it requires at least $Np$ fields, whereas
Lie brackets may yield full rank with fewer fields.

\subsubsection{Computing the Lie-algebra rank}\label{sec:computing-lie-rank}

Following \citet[Section~3.3]{marcotte2023abide}, we investigate the
Lie-algebra rank condition by iteratively generating Lie brackets and
computing the rank of their evaluations.  For a finite family of smooth
lifted fields $g_1,\ldots,g_m$, set
$W_0\coloneqq\operatorname{span}\{g_1,\ldots,g_m\}$ and construct
\[
    W_k\coloneqq W_{k-1}+[W_0,W_{k-1}],\qquad k\geq1.
\]
Here $[W_0,W_{k-1}]$ is the real span of the corresponding brackets,
and $\operatorname{Lie}(W_0)=\bigcup_{k\geq0}W_k$.
At each step, the original generators are bracketed with the fields
constructed so far, and their evaluations at a configuration $X$ give
the rank $r_k(X)\coloneqq\dim W_k(X)$.  A rank $r_k(X)=D$
certifies full Lie rank at $X$.  More generally, the stopping criterion
of \citet[Lemma~3.2]{marcotte2023abide} applies if
$r_{k+1}(Y)=r_k(X)$ for every $Y$ in a neighborhood of $X$:
the traces $W_j(Y)$ then equal $W_k(Y)$ locally for all $j\geq k$. %
We adapt their \texttt{SageMath} implementation to \texttt{PyTorch} (code in \url{https://github.com/sibyllema/attention_universality}),
replacing symbolic differentiation with automatic differentiation. Lie brackets are evaluated through Jacobian--vector products, which
can themselves be differentiated to compute nested brackets without
forming full Jacobians or expanding symbolic expressions.
This makes it possible to explore larger configuration spaces and
higher bracket orders for computationally demanding vector fields,
such as the softmax-attention updates in \eqref{eq:vectorfields_att},
whose repeated symbolic differentiation becomes costly.

\subsection{Exact and universal interpolation at finite depth}\label{sec:finite-depth}

Euler discretization of an interpolating ODE generally leaves a nonzero
output error. For a fixed finite family of smooth fields, full Lie
rank nevertheless suffices for finite-depth interpolation using
residual layers with signed scalar coefficients; see
\Cref{lem:finite-depth-lie-rank}.
The stronger direct spanning condition gives an explicit correction
bound: under local $C^1$ regularity of the selected fields, at most
$D=Np$ additional residual layers make a sufficiently fine Euler
discretization exact.
Near the target, $D$ independent update directions make the endpoint
map locally invertible in the corresponding step sizes, allowing
them to cancel a sufficiently small discretization error exactly.

At finite depth, exact and universal interpolation are defined as in
\Cref{def:exact-interpolation,def:universal-interpolation}, respectively,
replacing the controlled ODEs with finite compositions of residual layers
on the batch.  In the exact case, we write these layers as
$\operatorname{Id}+f_\theta^{(N)}$: for both architectures considered here,
any scalar step size can be absorbed into the output projections.
In the universal case, the layers take the form
$\operatorname{Id}+c_k f_{\theta_{j_k}}^{(N)}$, where the parameters
$\theta_1,\ldots,\theta_m$ are fixed across all admissible datasets,
while the indices $j_k$ and scalar coefficients $c_k\in\mathbb{R}$
may depend on the dataset.
All intermediate configurations must remain in $\mathcal{M}_N$.

\begin{restatable}[Exact finite-depth interpolation under Lie rank]{proposition}{propfinitedepthlierank}
\label{prop:finite-depth-local-lie-rank-proposal}
\label{lem:finite-depth-lie-rank}
    Let $D\geq1$, let $\mathcal M\subseteq\mathbb R^D$ be nonempty,
    open, and connected, and let $\mathcal G$ be a family of locally
    Lipschitz vector fields on $\mathcal M$ closed under real scalar
    multiplication.  Suppose that, at each $Z\in\mathcal M$, there
    are finitely many fields $g_1,\ldots,g_m\in\mathcal G$ that are
    smooth on a neighborhood of $Z$ and satisfy
    \[
        \operatorname{Lie}(g_1,\ldots,g_m)(Z)=\mathbb R^D.
    \]
    Then, for every $X,Y\in\mathcal M$, there exist a finite integer
    $L\geq0$ and fields $h_0,\ldots,h_{L-1}\in\mathcal G$ such that
    \[
        Z^0=X,\qquad Z^{k+1}=Z^k+h_k(Z^k),\qquad Z^L=Y,
    \]
    with $Z^k\in\mathcal M$ for every $0\leq k\leq L$.
\end{restatable}

For a batch space $\mathcal{M}_N$, the dimension is $D=Np$.
The proof is given in \Cref{app:finite-depth-lie-rank-proposal}.
The direct-spanning version is stated and proved in
\Cref{app:finite-depth-direct-spanning}.

\begin{restatable}[Universal finite-depth interpolation under Lie rank]{proposition}{propuniversalfinitedepthlierank}
\label{prop:universal-finite-depth-lie-rank}
    Fix $N\geq1$ and a configuration space $\mathcal M_N$ as in
    \Cref{def:universal-interpolation}, and assume that $\mathcal F$
    allows real scalar multiplication.  Suppose there exist finitely
    many parameters $\theta_1,\ldots,\theta_m$ such that, at each
    $X\in\mathcal M_N$, an index set $I_X\subseteq\{1,\ldots,m\}$
    selects lifted fields smooth near $X$ with
    \[
        \operatorname{Lie}\bigl(f_{\theta_j}^{(N)}:j\in I_X\bigr)(X)
        =\mathbb R^{Np}.
    \]
    Then $\mathcal F$ has universal interpolation at finite depth
    on $\mathcal M_N$, using these same parameters.
\end{restatable}

The proof is given in \Cref{app:universal-finite-depth-lie-rank}.
The direct-spanning version is stated and proved in
\Cref{app:universal-finite-depth-direct-spanning}.
Under the direct-spanning assumption of
\Cref{prop:finite-depth-interpolation}, at most $D=Np$ layers are
needed for the final correction.  Neither construction gives a bound
on total depth or parameter norms in terms of dataset size and
architecture dimensions.

\section{Applications to continuous- and finite-depth networks}

We apply the criteria of \Cref{sec:interpolation-framework} to noncausal
and causal self-attention. The appendix also discusses
bias-free ReLU ResNets and the obstruction to interpolation in linear
networks (\Cref{sec:resnet,app:linear-proof}).

\subsection{Noncausal self-attention networks}

In this section, we show that noncausal self-attention networks achieve universal interpolation with a finite number of frozen Gaussian blocks. For single-head attention of width {$r\ge d$}, two blocks suffice for every batch size and sequence length (\Cref{thm:two-gaussian-connected-domain}).

For $N\geq1$, write $X=\operatorname{vec}(x^1,\ldots,x^N)\in\mathbb R^{Nnd}$,
where $x^\ell=\operatorname{vec}(x_1^\ell,\ldots,x_n^\ell)$ and
$x_i^\ell\in\mathbb R^d$. 
Our interpolation domain is
\[
    \mathcal B_{N,n,d}
    \coloneqq
    \left\{
        X\in\mathbb{R}^{Nnd}:
        \begin{array}{l}
        x_i^\ell\neq0,\quad x_i^\ell\notin\mathbb{R}x_{i'}^{\ell}\ \text{ for } i\neq i',\\[-1mm]
        x^{\ell'}\neq\pm P x^\ell\ \text{ for } \ell<\ell',\ P\in S_n
        \end{array}
    \right\},
\]
where $S_n$ acts by permuting tokens. Tokens are nonzero and pairwise non-collinear within each sequence, may be shared across sequences, and no two sequences are related by a signed permutation. We will also consider the generic subset
\[
    \mathcal{A}_{N,n,d}
    \coloneqq
    \left\{
        X\in\mathbb{R}^{Nnd}:
        \begin{array}{l}
        x_i^\ell\neq0,\ \text{and}\\[-1mm]
        x_i^\ell\notin\mathbb{R}x_{i'}^{\ell'}
        \text{ whenever }(\ell,i)\neq(\ell',i')
        \end{array}
    \right\}.
\]
For $d\geq3$, both domains are open, dense and connected, with
$\mathcal A_{N,n,d} \subseteq \mathcal B_{N,n,d}$; see
\Cref{app:attention-proof,app:attention-uniform-generation}. Noncausal attention is permutation-equivariant and odd, as are its
residual layers and controlled flows.  Thus, two input sequences
related by $x^{\ell'}=\varepsilon P x^\ell$, where $\ell\neq\ell'$,
$P$ permutes tokens, and $\varepsilon\in\{\pm1\}$, require targets
satisfying $y^{\ell'}=\varepsilon P y^\ell$.
Both domains exclude such input pairs.  These are sufficient restrictions:
symmetry-related sequences can also be interpolable when their targets
respect those symmetries.

\begin{restatable}[Pointwise spanning by noncausal self-attention]{lemma}{thmattentionuit}
\label{thm:attention-uit}\label{lem:attention-pointwise-spanning}
    Let $d\geq2$, $n\geq2$, and $N\geq1$.  At every
    $X\in\mathcal A_{N,n,d}$, the single-head batch fields satisfy
    \[
        \operatorname{span}\left\{
            f^{(N)}_{W,V}(X):
            \operatorname{rank}W,\operatorname{rank}V\leq1
        \right\}=\mathbb R^{Nnd}.
    \]
\end{restatable}

The proof is given in \Cref{app:attention-proof}.

\begin{remark}[Exact interpolation on $\mathcal A_{N,n,d}$]
For $d\geq3$, connectedness and the criteria of
\Cref{sec:interpolation-framework} turn this lemma into exact
interpolation on $\mathcal A_{N,n,d}$ at both continuous and finite
depth, with all intermediate configurations in this domain.
One head of width one suffices, so the conclusion holds for any
prescribed number of heads and positive head widths.
For $n=1$, attention is linear and exact interpolation fails for
$N>d$ by \Cref{lem:linear-obstruction}.
\end{remark}

The following theorem (see \Cref{app:attention-uniform-generation} for a proof) shows that a finite number of random attention blocks suffices to achieve universal interpolation on $\mathcal B_{N,n,d}$, and hence exact interpolation, with the required number of blocks growing with  $N$.

\begin{restatable}[Universal interpolation with Gaussian-initialized attention blocks]{theorem}{thmuniformattentiongeneration}
\label{thm:attention-uniform-finite-generation}
    Let $d\geq3$, $n\geq2$, and $N\geq1$, and put $D\coloneqq Nnd$.
    Fix an $H$-head architecture with positive widths $r_1,\ldots,r_H$.
    Initialize $m_N=2D$ noncausal attention blocks by drawing every entry of
    all their query, key, value, and output projection matrices
    independently from $\mathcal N(0,1)$.
    The effective matrices are defined by the factorizations in
    \Cref{ex:attlayers}.  With probability one over this single draw,
    these frozen blocks yield universal interpolation at continuous
    depth on $\mathcal{B}_{N,n,d}$ in the sense of
    \Cref{def:universal-interpolation}, simultaneously for all input
    and target configurations in this domain.
    Only the block indices, signs, and durations depend on the
    prescribed interpolation problem; the terminal depth $S>0$
    may depend on the endpoints.
\end{restatable}

The proof has two steps. First, we prove a uniform version of Lemma~3.1: almost surely, $2D$ Gaussian blocks drawn independently of $X$ span the tangent space at every $X \in \mathcal{A}_{N,n,d}$ simultaneously. This follows from the pointwise full-rank witness given by Lemma~3.1 and an analyticity argument (see \Cref{lem:attention-uniform-direct-generation}). Second, we extend full Lie rank to $\mathcal{B}_{N,n,d}$. By contextual separation, a short flow along one of the sampled blocks moves any configuration into $\mathcal{A}_{N,n,d}$ (see \Cref{lem:attention-uniform-separation}). The system is analytic, so the Lie rank stays the same along this flow and is therefore full at the starting point. Universal interpolation then follows from the connectedness of $\mathcal{B}_{N,n,d}$ by \Cref{cor:universal-interpolation-criterion}.

\begin{corollary}[Universal interpolation at finite depth]
\label{cor:attention-dictionary-finite-depth}
    Under the assumptions of \Cref{thm:attention-uniform-finite-generation},
    with probability one, the same $2Nnd$ sampled fields yield universal
    interpolation at finite depth on $\mathcal{B}_{N,n,d}$, in the sense
    of \Cref{sec:finite-depth}.  All nonendpoint configurations can be
    chosen in $\mathcal{A}_{N,n,d}$.
\end{corollary}

See \Cref{app:attention-dictionary-finite-depth} for a proof. The $2Nnd$ sampled blocks form a reusable dictionary; the number of layers
may depend on the interpolation problem.
In \Cref{thm:attention-uniform-finite-generation}, the number of frozen blocks grows with $N$. Our main result removes this dependence: for single-head attention of width {$r\ge d$}, a dictionary of only two frozen blocks suffices for every batch size and sequence length.

\begin{restatable}[Universal interpolation with two Gaussian attention blocks]{theorem}{thmtwogaussianattention}
\label{thm:two-gaussian-connected-domain}
    Fix $d\geq3$.  Consider two noncausal single-head attention blocks
    with head width $r\ge d$ and query, key, value, and
    output projection matrices
   whose entries are all
    sampled independently from $\mathcal N(0,1)$.
    With probability one, simultaneously for every $N\geq1$ and
    $n\geq2$, there exists an open and connected subset
    \[
        \mathcal R_{N,n}\subseteq\mathcal B_{N,n,d},
    \]
    depending on the sampled matrices, whose complement in
    $\mathcal B_{N,n,d}$ has Lebesgue measure zero.
    On $\mathcal R_{N,n}$, these two frozen blocks yield universal
    interpolation at continuous depth in the
    sense of \Cref{def:universal-interpolation}, with $m_N=2$.
    The same sampled matrices work for every $N$ and $n$; only the block
selection, signs, and durations depend on the interpolation problem.
\end{restatable}

For almost every draw of the blocks, input and target batches with
i.i.d.\ standard Gaussian coordinates, sampled independently of the
blocks, belong to $\mathcal R_{N,n}$ with probability one.

We define $\mathcal R_{N,n}$ as the configurations where the two sampled
fields and their iterated Lie brackets span the full batch space.
The main step is to show that the analytic rank-deficient set has
codimension at least two.  Removing this set preserves connectedness,
allowing the interpolation criterion in
\Cref{cor:universal-interpolation-criterion} to connect any two
configurations in $\mathcal R_{N,n}$.  The full proof and the precise definition of the domain are given in \Cref{app:two-gaussian-attention}, which also contains the proof of the finite-depth corollary.

\begin{remark}[Head width and dictionary size]
The theorem covers the standard single-head choice $r=d$.
For $r<d$, our numerical experiments {(\Cref{app:attention-width-threshold})} suggest that
$m\ge\lceil d/r\rceil$ independent Gaussian-initialized frozen
blocks suffice for generic full lifted Lie rank, independently
of $N$ and $n${, with $n\ge2$}.
With the standard width $r=d/H$ ($H\ge2$), this gives the
conjectured threshold $m\ge H$, using one active head per block
and setting the other heads to zero.
A corresponding universal interpolation theorem remains open.
\end{remark}

\begin{corollary}[Universal interpolation at finite depth with two blocks]
\label{cor:two-gaussian-finite-depth}
    Under the assumptions of \Cref{thm:two-gaussian-connected-domain},
    with probability one, the same two frozen blocks yield universal
    interpolation at finite depth on the same domains
    $\mathcal R_{N,n}$, in the sense of \Cref{sec:finite-depth},
    simultaneously for every $N\geq1$ and $n\geq2$.
    Every layer uses one of the two sampled fields with a real scalar
    gain, and every intermediate batch configuration remains in
    $\mathcal R_{N,n}$.
\end{corollary}

\subsection{Causal self-attention networks}\label{sec:causal-attention}

In this section, we show that causal masking makes the first-token update linear, so every linear relation among the first tokens of a batch is preserved at every depth. As a result, exact interpolation fails for $N>d$, whatever the number of heads (\Cref{prop:causal-rank}). Positive results are recovered once input and target batches share the same first-token relations: on each such class, finitely many frozen Gaussian blocks achieve universal interpolation (\Cref{cor:causal-random-dictionary}).

We consider the causal attention architecture of \Cref{ex:attlayers}
with any prescribed number of heads and positive head widths.
Causal masking makes the first-token update linear: the first token
attends only to itself.
We retain the globally noncollinear domain $\mathcal{A}_{N,n,d}$;
the extension allowing shared tokens in Section~3.1 concerns noncausal
attention.  The interpolation constructions below require only one
active head; all other value matrices can be set to zero.
We denote the batch lift of a single-head causal update by
$F^{\mathrm c}_{W,V}$.
For a general $H$-head block,
$[f^{\mathrm c}_{\mathbf W,\mathbf V}(x)]_1=V_\Sigma x_1$,
where $V_\Sigma\coloneqq\sum_{h=1}^H V_h$.
For any batch $X$, denote its first-token matrix by
\[
    X_1\coloneqq
    \begin{bmatrix}x_1^1&\cdots&x_1^N\end{bmatrix}\in\mathbb{R}^{d\times N}.
\]
Along continuous and discrete batch trajectories $Z(s)$ and $Z^k$
starting at $X$, the first-token matrices satisfy
$\dot Z_1=V_\Sigma(s)Z_1$ and
$Z_1^{k+1}=(I+V_{\Sigma,k})Z_1^k$, respectively.
In both settings, the first-token matrix at any depth has the form
$MX_1$ for some $M\in\mathbb{R}^{d\times d}$.
Hence every initial linear relation among the first tokens remains
valid at every depth.
For example, an input relation $x_1^3=x_1^1+x_1^2$ forces
every reachable target to satisfy $y_1^3=y_1^1+y_1^2$, at both continuous
and finite depth.  The following proposition shows that later token
positions admit independent update directions.

\begin{proposition}[Exact rank of causal attention]\label{prop:causal-rank}
\label{rem:causal-obstruction}
    Let $d\geq2$, $n\geq2$, and $N\geq1$.
    At a fixed batch $X\in\mathcal{A}_{N,n,d}$, let
    $\mathcal{S}^{\mathrm c}_X$ denote the linear span of the batch updates
    $F^{\mathrm c}_{W,V}(X)$ with
    $\operatorname{rank}W,\operatorname{rank}V\leq1$.  Then
    \begin{equation}\label{eq:causal-span}
        \mathcal{S}^{\mathrm c}_X
        =\left\{Z\in\mathbb{R}^{Nnd}:
        Z_1=A X_1
        \text{ for some }A\in\mathbb{R}^{d\times d}\right\}.
    \end{equation}
    Consequently,
    \begin{equation}\label{eq:causal-rank}
        \dim\mathcal{S}^{\mathrm c}_X
        =N(n-1)d+d\,\operatorname{rank}X_1.
    \end{equation}
    The same span and rank hold for any prescribed number of heads
    and positive head widths.  Lie brackets do not increase this dimension.
    Full ambient rank
    $Nnd$ holds exactly when the $N$ first tokens are linearly independent,
    which requires $N\leq d$.
\end{proposition}

The proof is given in \Cref{app:causal-rank}.
The $N(n-1)d$ directions associated with later tokens are unrestricted;
the first-token directions are exactly those obtained by applying a common
linear map to the first tokens.
By the rank criterion and its analytic converse in
\Cref{cor:exact-interpolation-criterion,rem:analytic-orbit}, causal
attention has the continuous-depth exact interpolation property on a
nonempty connected open subset of $\mathcal{A}_{N,n,d}$ if and only if
the first tokens are linearly independent at every configuration in
that subset.  For $N>d$, this is impossible; the preserved first-token relations also rule out arbitrary
interpolation at finite depth.  This obstruction concerns interpolation
of the full target sequences, including their first tokens.
Pairwise noncollinearity alone does not ensure linear
independence when $N\geq3$.  

To obtain positive interpolation results for arbitrary $N$, we restrict
input and target batches to have exactly the same linear relations among
their first tokens.  For a fixed subspace $K\subseteq\mathbb{R}^N$, define
\[
    \mathcal{A}^{K}_{N,n,d}
    \coloneqq\{X\in\mathcal{A}_{N,n,d}:\ker X_1=K\},
    \qquad r\coloneqq N-\dim K.
\]
Here $c\in K$ means that $\sum_{\ell=1}^N c_\ell x_1^\ell=0$.
This class is an open subset of the linear space
\[
    E_K\coloneqq\{Z\in\mathbb{R}^{Nnd}:Z_1c=0\text{ for every }c\in K\},
    \qquad \dim E_K=R\coloneqq N(n-1)d+dr.
\]
We apply the interpolation criteria of \Cref{sec:interpolation-framework}
within each connected component, using $E_K$ as the ambient space.
The rank in \eqref{eq:causal-rank} is full in this space.

\begin{corollary}[Exact interpolation by causal attention]
\label{cor:causal-interpolation}
    Let $d\geq2$, $n\geq2$, $N\geq1$, and fix
    $K\subseteq\mathbb{R}^N$ such that $\mathcal{A}^{K}_{N,n,d}$ is
    nonempty.  The causal attention family has exact interpolation on
    each connected component of $\mathcal{A}^{K}_{N,n,d}$, both at
    continuous depth on $[0,S]$ for a suitable $S>0$ and at finite depth.
    One head of width one suffices, with effective parameter matrices
    $W,V$ of rank at most one shared across all sequences.
    The continuous trajectory and all
    intermediate residual configurations remain in the chosen component.
\end{corollary}

The proof is given in \Cref{app:causal-interpolation}.  Continuous flows
keep the first-token kernel fixed, since their linear action is invertible.
Their trajectories within $\mathcal{A}^{K}_{N,n,d}$ remain in a single
connected component. 

\begin{corollary}[Universal interpolation by causal attention]
\label{cor:causal-random-dictionary}
    Under the assumptions of \Cref{cor:causal-interpolation}, fix $K$
    before sampling and set $R=N(n-1)d+dr$.  Draw independent standard
    Gaussian vectors $a_j,b_j,u_j,v_j\in\mathbb{R}^d$, $1\leq j\leq2R$,
    and set $
        W_j=a_jb_j^\top$, $V_j=u_jv_j^\top.$
    These matrices specify one active head in each block; any
    additional heads have zero value matrices.
    With probability one, these $2R$ frozen blocks yield universal
    interpolation on each connected component of $\mathcal{A}^{K}_{N,n,d}$,
    at both continuous and finite depth.  The blocks may be used with
    either sign and, at finite depth, with real scalar gains.
    A single draw works simultaneously for all components of this fixed
    relation class, with each interpolation remaining in its chosen component.
\end{corollary}

\section*{Conclusion}
In this paper, we establish universal interpolation for residual self-attention networks with frozen attention parameters, at both continuous and
finite depth. In the single-head setting, two randomly sampled fixed blocks are sufficient, independently of the dataset size.
A natural direction for future work is to better understand the trade-off between the number of fixed generators and the depth required for interpolation.

\section*{Acknowledgements}
We wish to thank Margot Bruneaux, Frédéric Marbach, Domènec Ruiz-Balet and Federico Scavia for fruitful discussions, and we are particularly grateful to Cyril Letrouit, who introduced us to his article \cite{agrachev2026generic} and the open question it raises.

\section*{AI disclosure section}
We used LLMs to aid or polish our writing. We primarily used OpenAI GPT-6 Astra, and all parts of this paper were polished by an LLM at some point, following the instructions provided by the authors.
We used LLMs for retrieval and discovery, particularly to find related work both within the ML community and in other fields that helped us prove certain results, such as \cite{amir2023neural} and \cite{ax1971schanuel}.
We used LLMs for research execution both to assist with code development and to develop proofs (see our final point below).
We used LLMs to draft sections of the paper. Many sections were initially drafted by Astra. The authors systematically refined these drafts through extensive iteration, providing precise feedback at each step, and editing the text manually when needed. 
Finally, we used LLMs to prove mathematical claims at three different levels: i) proving direct consequences of known / previously proven theorems (see e.g. \Cref{app:exact-rank-proof}--\Cref{app:pointwise-rank-proof}), which LLMs can do easily and faster than the authors; ii) refining proofs starting from our own drafts (e.g., \Cref{app:finite-depth-proof}, \Cref{sec:resnet}, \Cref{app:attention-proof}, \Cref{app:causal-rank}); and iii) finding ideas for difficult proofs that required tools from other research communities (e.g., parts of \Cref{app:attention-generic-generation}, parts of \Cref{app:two-gaussian-algebraic-barriers}--{\Cref{app:two-gaussian-conclusion}}). We used Astra for all three purposes. All proofs have been checked by the authors. For iii), we took particular care to explain the tools and arguments drawn from other research communities in a way that is accessible to the ML community. We take responsibility for the final content of this work, including text, claims or artifacts produced with the aid of generative AI.

\section*{Reproducibility statement}
We provide code that can reproduce our results as supplementary materials, and describe experimental
details and results in \Cref{app:attention-width-threshold}.

\bibliography{biblio}
\bibliographystyle{iclr2027_conference}


\clearpage
\appendix

\section{Proof of \texorpdfstring{\Cref{cor:exact-interpolation-criterion}}{the exact interpolation criterion}}\label{app:exact-rank-proof}

\begin{proof}
    Write $\mathcal{M}=\mathcal{M}_N$, $D=Np$, and
    $\mathcal{G}=\{f_\theta^{(N)}:\theta\in\Theta\}$.  Fix
    $X\in\mathcal{M}$ and choose the locally smooth lifted fields
    $g_1,\ldots,g_m$ from the hypothesis.  Full Lie rank at $X$ gives
    $D$ Lie-algebra elements whose evaluations form a basis.  Their
    determinant stays nonzero on a smaller connected neighborhood
    $\mathcal{U}$ of $X$.  Apply \Cref{thm:chow-rashevskii} on
    $\mathcal{U}$ to the family
    $\{\pm g_j|_{\mathcal{U}}:1\leq j\leq m\}$.
    This joins every pair in $\mathcal{U}$ by a piecewise-constant
    control using the selected fields and their negatives.  Thus a
    neighborhood of $X$ belongs to its reachability class.

    Reachability by finite concatenations of flows of fields in $\mathcal{G}$,
    with every segment contained in $\mathcal{M}$, is an equivalence relation:
    transitivity follows by concatenation, while symmetry follows by reversing
    the segments and replacing each field by its negative.  Applying the
    preceding local argument at every point shows that every equivalence class
    is open.  Since $\mathcal{M}$ is connected, it has only one reachability
    class.  Thus any distinct $X,Y\in\mathcal{M}$ can be joined by
    finitely many positive-time flow segments:
    \[
        Y=\Phi_{h_L}^{\tau_L}\circ\cdots\circ
        \Phi_{h_1}^{\tau_1}(X),
        \qquad h_j\in\mathcal{G},\quad \tau_j>0,
    \]
    with every segment contained in $\mathcal{M}$, where $\Phi_h^t$
    denotes the local flow of $h$.  Taking $S\coloneqq\sum_j\tau_j$
    gives the required piecewise-constant control on $[0,S]$.
    When $X=Y$, follow any field in $\mathcal{G}$ for a sufficiently
    short positive time and then its negative for the same duration.
    Local existence and openness of $\mathcal{M}$ ensure that this
    trajectory stays in $\mathcal{M}$ and returns to $X$ with $S>0$.

    Conversely, suppose all lifted fields are analytic and exact
    interpolation holds.  Each controlled trajectory is a finite
    concatenation of their flows, so every orbit is $\mathcal{M}$.
    By \Cref{rem:analytic-orbit},
    $\operatorname{Lie}(\mathcal{G})(X)=\mathbb{R}^D$ at every $X$.
    A basis of this trace involves only finitely many Lie brackets
    and hence finitely many fields of $\mathcal{G}$, giving the
    required finite parameter selection at $X$.
\end{proof}

\section{Proof of \texorpdfstring{\Cref{cor:universal-interpolation-criterion}}{the universal interpolation criterion}}
\label{app:universal-rank-proof}

\begin{proof}
    Fix the finite parameter selection from the hypothesis.
    Apply \Cref{cor:exact-interpolation-criterion} to
    \[
        \widetilde{\mathcal{F}}_N
        =\{\pm f_{\theta_j}:1\leq j\leq m\}.
    \]
    This family is closed under negation, and at
    each configuration the indices $I_X$ supply a locally smooth
    subfamily with full Lie rank.  On the batch, the resulting controls
    select one signed lifted update on each of finitely many intervals, with a terminal
    depth $S>0$, exactly as in \Cref{def:universal-interpolation}.
    The parameters were chosen before the endpoints, so
    $\mathcal{F}$ has universal interpolation on $\mathcal{M}_N$.

    Conversely, suppose the lifted fields are analytic and choose
    the finite parameters supplied by universal interpolation on
    $\mathcal{M}_N$.  These fields and their negatives have one orbit on
    $\mathcal{M}_N$.  By \Cref{rem:analytic-orbit} their Lie algebra
    has rank $Np$ everywhere.  Adding negatives does not
    change the generated Lie algebra, so the fixed lifted fields
    themselves have full Lie rank; take $I_X=\{1,\ldots,m\}$.
\end{proof}

\section{Justification of \texorpdfstring{\Cref{rem:analytic-orbit}}{the analytic converse}}
\label{app:analytic-orbit-proof}

\begin{proof}
    Let $\mathcal{M}\subseteq\mathbb{R}^D$ be nonempty, open, and
    connected, and let $\mathcal{G}$ be a family of real-analytic
    fields on $\mathcal{M}$ closed under negation.
    Let $\mathcal{O}(X)$ denote the orbit
    through $X$ generated by their local flows.  The analytic orbit
    theorem gives
    \[
        T_X\mathcal{O}(X)=\operatorname{Lie}(\mathcal{G})(X)
    \]
    for every $X\in\mathcal{M}$~\citep[Corollary~5.17]{agrachev2004control}
    \citep[see also][Theorem~5]{sussmann1998analytic}.
    Closure under negation allows flow segments to be reversed.  Thus
    $\mathcal{O}(X)$ is precisely the set reachable from $X$ by
    piecewise-constant controls on some finite interval $[0,S]$.
    The point $X$ itself is reached by a short flow segment followed
    by its reverse.  If the conclusion of
    \Cref{thm:chow-rashevskii} holds, then
    $\mathcal{O}(X)=\mathcal{M}$, so
    $\operatorname{Lie}(\mathcal{G})(X)=T_X\mathcal{M}=\mathbb{R}^D$.
    Sufficiency is already given by \Cref{thm:chow-rashevskii}.
\end{proof}

\section{Pointwise full-rank criterion}\label{app:pointwise-rank-proof}

\begin{corollary}[Pointwise full-rank criterion]\label{cor:pointwise-full-rank-criterion}
    Let $D\geq1$, let $\mathcal{M}\subseteq\mathbb{R}^D$ be nonempty,
    open, and connected, and let
    $\mathcal{G}$ be a family of locally Lipschitz vector fields on
    $\mathcal{M}$.  Suppose that:
    \begin{enumerate}
        \item for every $X\in\mathcal{M}$, there are $D$ vector fields
        $g_1,\ldots,g_D\in\mathcal{G}$ that are $C^1$ on a neighborhood of
        $X$ and whose values $g_1(X),\ldots,g_D(X)$ form a basis of
        $\mathbb{R}^D$; and
        \item $g\in\mathcal{G}$ implies $-g\in\mathcal{G}$.
    \end{enumerate}
    Then piecewise-constant controls with values in $\mathcal{G}$ exactly
    interpolate every pair $X,Y\in\mathcal{M}$ on $[0,S]$ for some
    $S>0$ depending on the pair.
\end{corollary}

\begin{proof}
    Fix $X\in\mathcal{M}$ and choose $g_1,\ldots,g_D$ as in the first
    assumption.  If $\varphi_j^t$ denotes the local flow of $g_j$, the endpoint
    map
    \[
        E(t_1,\ldots,t_D)
        \coloneqq
        \varphi_D^{t_D}\circ\cdots\circ\varphi_1^{t_1}(X)
    \]
    has derivative at the origin whose columns are
    $g_1(X),\ldots,g_D(X)$.  The inverse function theorem therefore shows that
    the reachability class of $X$ contains a neighborhood of $X$; negative
    times are realized using the fields $-g_j$.  The symmetry,
    concatenation, and connectedness arguments in the proof of
    \Cref{cor:exact-interpolation-criterion} then give the conclusion.
\end{proof}

\section{Finite-depth interpolation}\label{app:finite-depth-proof}

\subsection{Direct spanning}
\label{app:finite-depth-direct-spanning}

\begin{proposition}[Exact finite-depth interpolation]\label{prop:finite-depth-interpolation}
    Let $D\geq1$, let $\mathcal{M}\subseteq\mathbb{R}^D$ be nonempty,
    open, and connected, and let $\mathcal{G}$ be a family of locally
    Lipschitz vector fields on $\mathcal{M}$ closed under real scalar
    multiplication.  Suppose that, at each $X\in\mathcal{M}$,
    $\mathcal{G}$ contains $D$ fields that are $C^1$ near $X$ and whose
    values at $X$ form a basis of $\mathbb{R}^D$.
    Then, for every $X,Y\in\mathcal{M}$, there exist a finite integer $L\geq0$ and
    fields $g_0,\ldots,g_{L-1}\in\mathcal{G}$ such that
    \begin{equation}\label{eq:finite-depth-recurrence}
        Z^0=X,\qquad Z^{k+1}=Z^k+g_k(Z^k),\qquad Z^L=Y,
    \end{equation}
    with $Z^k\in\mathcal{M}$ for every $0\leq k\leq L$.
\end{proposition}

\begin{proof}
    Fix $X,Y\in\mathcal M$.

    \emph{Step 1: construct an exact correction near the target.}
    We first find a neighborhood of $Y$ such that each point in this
    neighborhood can be sent exactly to $Y$ by at most $D$ admissible
    residual layers, with all intermediate states remaining in
    $\mathcal M$.  The layer coefficients may depend on the point
    being corrected.  Choose
    $g_1,\ldots,g_D\in\mathcal{G}$ that are $C^1$ on a neighborhood of $Y$
    and whose values at $Y$ form a basis of $\mathbb{R}^D$.  For
    $a=(a_1,\ldots,a_D)\in\mathbb{R}^D$, set
    \[
        R_j^{a_j}(Z)\coloneqq Z+a_jg_j(Z),\qquad
        E(Z,a)\coloneqq
        R_D^{a_D}\circ\cdots\circ R_1^{a_1}(Z).
    \]
    This map is $C^1$ near $(Y,0)$.  By shrinking this neighborhood, all
    intermediate states in the composition remain in $\mathcal{M}$ and in
    the common $C^1$ neighborhood of the selected fields.  At $(Y,0)$,
    \begin{equation}\label{eq:finite-depth-endpoint-derivative}
        E(Y,0)=Y,\qquad
        D_aE(Y,0)=
        \begin{bmatrix}g_1(Y)&\cdots&g_D(Y)\end{bmatrix}.
    \end{equation}
    The latter matrix is invertible.  Applying the implicit function
    theorem to $E(Z,a)-Y=0$ gives an open neighborhood
    $\mathcal{U}_Y\subseteq\mathcal{M}$ of $Y$ and a $C^1$ map
    $a:\mathcal{U}_Y\to\mathbb{R}^D$, with $a(Y)=0$, such that
    \begin{equation}\label{eq:finite-depth-exact-correction}
        E(Z,a(Z))=Y\qquad\text{for every }Z\in\mathcal{U}_Y.
    \end{equation}
    Each nonzero coefficient $a_j(Z)$ can be absorbed into its field,
    since closure under real scalar multiplication gives
    $a_j(Z)g_j\in\mathcal{G}$.  A zero coefficient gives an identity layer
    and can be omitted.  Thus every $Z\in\mathcal U_Y$ can be mapped
    exactly to $Y$ by at most $D$ admissible residual layers.

    \emph{Step 2: reach the correction neighborhood by Euler discretization.}
    We next approximate a controlled trajectory from $X$ to $Y$ by
    admissible residual layers.  We show that a sufficiently fine
    mesh keeps every iterate in $\mathcal M$ and places the endpoint
    in the neighborhood $\mathcal U_Y$ constructed in Step~1.
    By \Cref{cor:pointwise-full-rank-criterion}, a piecewise-constant
    controlled trajectory $\gamma:[0,S]\to\mathcal{M}$ joins $X$ to $Y$
    for some $S>0$.
    Fix any such trajectory.  Its image $\Gamma$ is compact, so there is
    $\rho>0$ such that the compact tube
    \[
        K_\rho\coloneqq
        \{Z\in\mathbb{R}^D:\operatorname{dist}(Z,\Gamma)\leq\rho\}
        \subseteq\mathcal{M}.
    \]
    Let $0=s_0<\cdots<s_q=S$ be the switching times of the fixed
    control, and let $h_i\in\mathcal G$ denote the vector field selected
    on $[s_{i-1},s_i)$, so that
    \[
        \dot\gamma(t)=h_i(\gamma(t))
        \qquad\text{for }t\in(s_{i-1},s_i).
    \]
    The fields $h_1,\ldots,h_q$ need not be distinct. By compactness
    and local Lipschitz continuity, these finitely many fields have
    a common norm bound $B\geq1$ and a common Lipschitz constant
    $\Lambda\geq1$ on $K_\rho$.

    Indeed, each $h_i$ is continuous, so compactness of $K_\rho$ gives
    \[
        B\coloneqq
        \max\left\{1,\max_{1\leq i\leq q}
        \max_{Z\in K_\rho}\|h_i(Z)\|\right\}<\infty.
    \]
    To obtain a common Lipschitz constant, local Lipschitz continuity
    and finiteness of the family imply that, for every $x\in K_\rho$,
    there exist $r_x>0$ and $L_x\geq1$ such that
    $\mathbb B(x,2r_x)\subseteq\mathcal M$ and all the $h_i$ are
    $L_x$-Lipschitz on this ball. Here $\mathbb B(x,r)$ denotes the
    open Euclidean ball of radius $r$ centered at $x$.

    By compactness, finitely many balls
    $\mathbb B(x_\ell,r_\ell)$, $\ell=1,\ldots,J$, cover $K_\rho$.
    Here $r_\ell=r_{x_\ell}$. Set
    \[
        \eta\coloneqq\min_{1\leq\ell\leq J}r_\ell>0,
        \qquad
        L_{\mathrm{loc}}\coloneqq\max_{1\leq\ell\leq J}L_{x_\ell}.
    \]
    If $Z,W\in K_\rho$ satisfy $\|Z-W\|<\eta$, choose $\ell$ with
    $Z\in\mathbb B(x_\ell,r_\ell)$. Then both $Z$ and $W$ belong to
    $\mathbb B(x_\ell,2r_\ell)$, and hence
    \[
        \|h_i(Z)-h_i(W)\|\leq L_{\mathrm{loc}}\|Z-W\|
        \qquad\text{for every }i.
    \]
    If instead $\|Z-W\|\geq\eta$, the norm bound gives
    \[
        \|h_i(Z)-h_i(W)\|
        \leq 2B
        \leq \frac{2B}{\eta}\|Z-W\|.
    \]
    Thus
    \[
        \Lambda\coloneqq\max\{1,L_{\mathrm{loc}},2B/\eta\}
    \]
    is a Lipschitz constant on $K_\rho$ shared by all the fields
    $h_1,\ldots,h_q$.

    Choose a finite mesh $0=t_0<\cdots<t_m=S$ containing every switching
    time.  Write $\Delta_k=t_{k+1}-t_k$,
    $h=\max_k\Delta_k$, and let $H_k\in\{h_1,\ldots,h_q\}$ be the field active
    on $[t_k,t_{k+1})$.  Forward Euler gives
    $Z^0=X$ and $Z^{k+1}=Z^k+\Delta_k H_k(Z^k)$.
    On each mesh interval, the integral equation for $\gamma$ yields
    \[
        \left\|\gamma(t_{k+1})-\gamma(t_k)
        -\Delta_k H_k(\gamma(t_k))\right\|
        \leq \tfrac12\Lambda B\Delta_k^2.
    \]
    Consequently, while $Z^k\in K_\rho$, the errors
    $e_k\coloneqq\|Z^k-\gamma(t_k)\|$ satisfy
    \[
        e_{k+1}\leq(1+\Lambda\Delta_k)e_k
        +\tfrac12\Lambda B\Delta_k^2.
    \]
    Since $e_0=0$ and $\Delta_k^2\leq h\Delta_k$, induction gives
    \begin{equation}\label{eq:euler-global-error}
        e_k\leq\frac{Bh}{2}
        \left(\prod_{j<k}(1+\Lambda\Delta_j)-1\right)
        \leq\frac{Bh}{2}(e^{\Lambda t_k}-1)
        \leq Ch,\qquad C\coloneqq\frac B2(e^{\Lambda S}-1).
    \end{equation}
    Choose $\varepsilon>0$ with
    $\{Z:\|Z-Y\|<\varepsilon\}\subseteq\mathcal{U}_Y$, and refine the
    mesh until $Ch<\min\{\rho,\varepsilon\}$.  This can be done with
    finitely many steps while retaining every switching time.  There can
    be no first iterate outside $K_\rho$: the bound for that iterate,
    computed from its predecessor inside the tube, would give distance
    less than $\rho$ from $\Gamma$.  Thus all iterates lie in
    $\mathcal{M}$, \eqref{eq:euler-global-error} holds throughout, and
    $\widetilde Y\coloneqq Z^m$ lies in the fixed neighborhood
    $\mathcal{U}_Y$.  Each Euler step is an admissible residual layer,
    since $\Delta_k H_k\in\mathcal{G}$ by positive scaling.

    \emph{Step 3: combine the Euler and correction layers to obtain exact interpolation.}
    Step~2 gives an Euler endpoint $\widetilde Y\in\mathcal U_Y$.
    We now apply the correction from Step~1 at this point to reach
    $Y$ exactly with at most $D$ additional layers.
    With the controlled trajectory and Euler mesh now fixed, set
    $c_j\coloneqq a_j(\widetilde Y)$ for $j=1,\ldots,D$, and define
    the correction fields
    \[
        \widehat g_j:W\longmapsto c_jg_j(W).
    \]
    Each $c_j$ is a real number, so closure under real scalar
    multiplication gives $\widehat g_j\in\mathcal G$.  Append the
    residual layers $\operatorname{Id}+\widehat g_j$ to the Euler
    network.  Their intermediate states satisfy
    \[
        W_0=\widetilde Y,\qquad
        W_j=W_{j-1}+c_jg_j(W_{j-1}),\qquad j=1,\ldots,D.
    \]
    By \eqref{eq:finite-depth-exact-correction}, their final output is
    $W_D=E(\widetilde Y,c)=Y$, where $c=(c_1,\ldots,c_D)$, and
    all intermediate states remain in $\mathcal M$.  Coefficients
    $c_j=0$ give identity layers and may be omitted.  Together with
    the Euler layers, this gives \eqref{eq:finite-depth-recurrence}
    with at most $m+D$ layers.
\end{proof}

\subsection{Extension under a Lie-rank condition}
\label{app:finite-depth-lie-rank-proposal}

We now prove \Cref{prop:finite-depth-local-lie-rank-proposal},
which replaces the direct-spanning assumption of
\Cref{prop:finite-depth-interpolation} by a local Lie-rank condition.
The proof follows the same three steps, with a new local correction
in Step~1.

\propfinitedepthlierank*

\begin{proof}
    Fix $X,Y\in\mathcal M$.  We construct a local correction at
    $Y$, reach its domain by Euler discretization, and concatenate
    the two networks.  If $X=Y$, no layers are needed to interpolate.

    \emph{Step 1: construct an exact correction near the target.}
    We seek a fixed finite sequence of residual layers whose
    coefficients can be adjusted to send any point near $Y$ to $Y$.
    Choose a finite family
    $g_1,\ldots,g_m\in\mathcal G$ that is smooth on an open
    neighborhood $U\subseteq\mathcal M$ of $Y$ and has full Lie
    rank at $Y$.  Write $\varphi_j^t$ for the local flow of $g_j$.

    \emph{A flow correction with independent parameter directions.}
    Let $\Psi$ range over finite compositions of the local flows
    $\varphi_i^t$, with signed times, defined near $Y$ and with all
    their flow segments contained in $U$, including the identity
    composition.  For a smooth vector field $h$, its pullback by
    $\Psi$ is the vector field
    \[
        (\Psi^*h)(Z)
        \coloneqq\bigl(D\Psi(Z)\bigr)^{-1}h(\Psi(Z)).
    \]
    The Jacobian $D\Psi(Z)$ is invertible since $\Psi$ is a
    composition of local flow maps.  Let $\mathcal V_Y$ be the
    linear span of $(\Psi^*g_j)(Y)$ over all such $\Psi$ and
    $1\leq j\leq m$.  We write
    $\operatorname{ad}_g h\coloneqq[g,h]$.
    The standard differentiation formula for pullbacks
    \citep[Section~2.5, Eq.~(2.21)]{agrachev2004control} gives
    \[
        \left.\frac{d}{dt}(\varphi_i^t)^*h\right|_{t=0}
        =[g_i,h]=\operatorname{ad}_{g_i}h.
    \]
    Mixed derivatives of pullbacks by successive flows therefore
    give the evaluations at $Y$ of all brackets of the form
    $\operatorname{ad}_{g_{i_1}}\cdots
    \operatorname{ad}_{g_{i_k}}g_j$.  Every such derivative is a
    limit of finite linear combinations of vectors in
    $\mathcal V_Y$.  Since this subspace of $\mathbb R^D$ is closed,
    the derivative also belongs to $\mathcal V_Y$.
    These brackets span the generated Lie algebra by the Jacobi
    identity.  The rank assumption consequently implies
    $\mathcal V_Y=\mathbb R^D$.

    Choose compositions $\Psi_1,\ldots,\Psi_D$ and indices
    $i_1,\ldots,i_D$ such that
    \[
        v_\ell=(\Psi_\ell^*g_{i_\ell})(Y),
        \qquad 1\leq\ell\leq D,
    \]
    form a basis of $\mathbb R^D$.  For $a=(a_1,\ldots,a_D)$ near
    zero, define
    \[
        C_\ell^{a_\ell}
        \coloneqq\Psi_\ell^{-1}\circ\varphi_{i_\ell}^{a_\ell}
                  \circ\Psi_\ell,
        \qquad
        E(Z,a)\coloneqq C_D^{a_D}\circ\cdots\circ C_1^{a_1}(Z).
    \]
    Each $C_\ell^{a_\ell}$ follows $\Psi_\ell$, flows along
    $g_{i_\ell}$ for time $a_\ell$, and follows $\Psi_\ell^{-1}$.
    The inverse uses the same flows in reverse order with negated
    times.  Thus $E$ uses only the $g_j$, and $E(Z,0)=Z$.
    Differentiating with respect to $a$ gives
    \begin{equation}\label{eq:proposed-lie-correction-flow-derivative}
        E(Y,0)=Y,\qquad
        A\coloneqq D_aE(Y,0)
        =\begin{bmatrix}v_1&\cdots&v_D\end{bmatrix}
    \end{equation}
    with $A$ invertible.  On a sufficiently small compact product
    neighborhood of $(Y,0)$, all trajectories defining $E(Z,a)$
    lie in a common compact subset of $U$.

    \emph{Replace each flow segment by residual layers.}
    Number the $J$ successive flow segments in $E$ as
    \[
        E(Z,a)=\varphi_{j_J}^{\tau_J(a)}\circ\cdots\circ
               \varphi_{j_1}^{\tau_1(a)}(Z).
    \]
    Here $j_\ell$ is the field index and $\tau_\ell(a)$ is the
    signed time of segment $\ell$.  Each $\tau_\ell$ is either a
    constant, from a fixed composition $\Psi_k$ or its inverse,
    or a coordinate $a_k$.  For fixed $a$, segment $\ell$ starting
    at an incoming state $\zeta$ is the solution, on $s\in[0,1]$, of
    \[
        \frac{d z_\ell}{ds}(s)
        =\tau_\ell(a)g_{j_\ell}(z_\ell(s)),
        \qquad z_\ell(0)=\zeta.
    \]
    Its endpoint is $z_\ell(1)=\varphi_{j_\ell}^{\tau_\ell(a)}(\zeta)$.
    This rescaling also covers negative or zero values of
    $\tau_\ell(a)$.  Choose an integer $q\geq1$ and approximate
    each such ODE by $q$ Euler steps on the grid $s_r=r/q$:
    \[
        z_\ell^0=\zeta,\qquad
        z_\ell^{r+1}=z_\ell^r+
        \frac{\tau_\ell(a)}q g_{j_\ell}(z_\ell^r),
        \qquad r=0,\ldots,q-1.
    \]
    Thus the exact flow of segment $\ell$ is approximated by
    $q$ residual layers with coefficient $\tau_\ell(a)/q$.
    Starting from $Z$ and feeding each segment's Euler endpoint
    into the next defines
    \[
        E_q(Z,a)=
        \left(\operatorname{Id}+\frac{\tau_J(a)}q g_{j_J}\right)^{\circ q}
        \circ\cdots\circ
        \left(\operatorname{Id}+\frac{\tau_1(a)}q g_{j_1}\right)^{\circ q}(Z).
    \]
    We use the same $q$ for all $J$ segments, giving $qJ$ layers.
    Increasing $q$ refines every segment's approximation; for each
    fixed $q$, the starting state $Z$ and parameters $a$ can vary.

    On a sufficiently small compact product neighborhood of $(Y,0)$,
    \begin{equation}\label{eq:proposed-lie-correction-c1-convergence}
        E_q\longrightarrow E\qquad\text{in }C^1\quad(q\to\infty),
    \end{equation}
    where derivatives are taken with respect to $(Z,a)$.
    All intermediate residual states remain in $U$ for large $q$.
    To control derivatives with respect to $a$ as well as the
    incoming state, adjoin the constant parameter $a$ to each
    segment's ODE.  The augmented state $u=(z,a)\in\mathbb R^{2D}$
    evolves under
    \[
        H_\ell(z,a)\coloneqq(\tau_\ell(a)g_{j_\ell}(z),0).
    \]
    Its time-one flow is
    $(z,a)\mapsto(\varphi_{j_\ell}^{\tau_\ell(a)}(z),a)$,
    and its Euler scheme is the one above with $a$ unchanged.
    The Jacobian $P(s)=D_{u(0)}u(s)$ satisfies
    \[
        \dot u=H_\ell(u),\qquad
        \dot P=DH_\ell(u)P,\qquad P(0)=I_{2D}.
    \]
    Differentiating the Euler updates gives exactly the Euler
    scheme for this variational system.  Smoothness on the compact
    neighborhoods above provides uniform bounds, so the Euler
    estimate in the preceding proof applies to both $u$ and $P$.
    It gives uniform convergence in $(z,a)$ of the maps and their
    Jacobians, while keeping the state iterates in $U$.  The chain
    rule for the finitely many segments then gives
    \eqref{eq:proposed-lie-correction-c1-convergence}.

    \emph{Recover an exact residual correction.}
    Although $E(Y,0)=Y$, generally $E_q(Y,0)\neq Y$.
    We first adjust $a$ to recover an exact return from $Y$ to $Y$.
    Choose $r>0$
    small enough that the closed parameter ball
    $\overline B_r(0)$ lies in the preceding neighborhood and
    \[
        \sup_{a\in\overline B_r(0)}
        \|I-A^{-1}D_aE(Y,a)\|\leq\tfrac14.
    \]
    By \eqref{eq:proposed-lie-correction-c1-convergence}, for all
    sufficiently large $q$,
    \[
        \sup_{a\in\overline B_r(0)}
        \|I-A^{-1}D_aE_q(Y,a)\|\leq\tfrac12,
        \qquad
        \|A^{-1}(E_q(Y,0)-Y)\|\leq\tfrac r4.
    \]
    Fix one such $q$, also large enough to keep every intermediate
    state in $U$.  The map
    \[
        T_q(a)\coloneqq a-A^{-1}(E_q(Y,a)-Y)
    \]
    is a contraction with constant at most $1/2$ on
    $\overline B_r(0)$ and satisfies
    \[
        \|T_q(a)\|\leq\|T_q(0)\|+\tfrac12\|a\|
        \leq\tfrac{3r}{4}.
    \]
    The contraction mapping theorem supplies a fixed point
    $a_q\in B_r(0)$.  Thus $E_q(Y,a_q)=Y$, and the derivative bound
    implies that $D_aE_q(Y,a_q)$ is invertible.

    Now allow the starting point $Z$ to vary near $Y$.
    The implicit function theorem at $(Y,a_q)$ gives a neighborhood
    $\mathcal U_Y\subseteq U$ and a $C^1$ map
    $\alpha:\mathcal U_Y\to B_r(0)$, with $\alpha(Y)=a_q$, such that
    \begin{equation}\label{eq:proposed-lie-residual-correction}
        E_q(Z,\alpha(Z))=Y
        \qquad\text{for every }Z\in\mathcal U_Y.
    \end{equation}
    Shrinking $\mathcal U_Y$ keeps all intermediate states in $U$.
    The correction uses $L_Y\coloneqq qJ$ layers: in segment $\ell$,
    each coefficient is $\tau_\ell(\alpha(Z))/q$.  These are fixed
    real scalars once $Z$ is specified, so scalar closure makes
    every layer admissible.  The integer $q$, the layer sequence,
    and $\mathcal U_Y$ are fixed before the next step.

    \emph{Step 2: reach the correction neighborhood by Euler discretization.}
    We now approximate a trajectory from $X$ to $Y$ closely enough
    to reach the fixed neighborhood $\mathcal U_Y$.
    By the local rank and connectedness argument proving
    \Cref{cor:exact-interpolation-criterion}, a piecewise-constant
    controlled trajectory $\gamma:[0,S]\to\mathcal{M}$ joins $X$ to $Y$
    for some $S>0$.
    Fix any such trajectory.  Its image $\Gamma$ is compact, so there is
    $\rho>0$ such that the compact tube
    \[
        K_\rho\coloneqq
        \{Z\in\mathbb{R}^D:\operatorname{dist}(Z,\Gamma)\leq\rho\}
        \subseteq\mathcal{M}.
    \]
    Let $0=s_0<\cdots<s_\nu=S$ be the switching times of the fixed
    control, and let $h_i\in\mathcal G$ denote the vector field selected
    on $[s_{i-1},s_i)$, so that
    \[
        \dot\gamma(t)=h_i(\gamma(t))
        \qquad\text{for }t\in(s_{i-1},s_i).
    \]
    These finitely many locally Lipschitz fields have a common
    norm bound $B\geq1$ and Lipschitz constant $\Lambda\geq1$ on
    $K_\rho$.  Indeed, continuity and compactness give
    \[
        B\coloneqq
        \max\left\{1,\max_{1\leq i\leq\nu}
        \max_{Z\in K_\rho}\|h_i(Z)\|\right\}<\infty.
    \]
    The finite-cover argument in the preceding proof gives
    $\eta>0$ and a common $L_{\mathrm{loc}}$ for pairs of points
    in $K_\rho$ at distance less than $\eta$.  For more distant
    pairs, the norm bound gives
    $\|h_i(Z)-h_i(W)\|\leq2B\leq(2B/\eta)\|Z-W\|$.
    Hence $\Lambda=\max\{1,L_{\mathrm{loc}},2B/\eta\}$ works
    on all of $K_\rho$.

    Choose a finite mesh $0=t_0<\cdots<t_{L_E}=S$ containing every switching
    time.  Write $\Delta_k=t_{k+1}-t_k$,
    $h=\max_k\Delta_k$, and let $H_k\in\{h_1,\ldots,h_\nu\}$ be the field active
    on $[t_k,t_{k+1})$.  Forward Euler gives
    $Z^0=X$ and $Z^{k+1}=Z^k+\Delta_k H_k(Z^k)$.
    On each mesh interval, the integral equation for $\gamma$ yields
    \[
        \left\|\gamma(t_{k+1})-\gamma(t_k)
        -\Delta_k H_k(\gamma(t_k))\right\|
        \leq \tfrac12\Lambda B\Delta_k^2.
    \]
    Consequently, while $Z^k\in K_\rho$, the errors
    $e_k\coloneqq\|Z^k-\gamma(t_k)\|$ satisfy
    \[
        e_{k+1}\leq(1+\Lambda\Delta_k)e_k
        +\tfrac12\Lambda B\Delta_k^2.
    \]
    Since $e_0=0$ and $\Delta_k^2\leq h\Delta_k$, induction gives
    \begin{equation}\label{eq:proposed-lie-euler-global-error}
        e_k\leq\frac{Bh}{2}
        \left(\prod_{j<k}(1+\Lambda\Delta_j)-1\right)
        \leq\frac{Bh}{2}(e^{\Lambda t_k}-1)
        \leq Ch,\qquad C\coloneqq\frac B2(e^{\Lambda S}-1).
    \end{equation}
    Choose $\varepsilon>0$ with
    $\{Z:\|Z-Y\|<\varepsilon\}\subseteq\mathcal{U}_Y$, and refine the
    mesh until $Ch<\min\{\rho,\varepsilon\}$.  This can be done with
    finitely many steps while retaining every switching time.  There can
    be no first iterate outside $K_\rho$: the bound for that iterate,
    computed from its predecessor inside the tube, would give distance
    less than $\rho$ from $\Gamma$.  Thus all iterates lie in
    $\mathcal{M}$, \eqref{eq:proposed-lie-euler-global-error} holds throughout, and
    $\widetilde Y\coloneqq Z^{L_E}$ lies in the fixed neighborhood
    $\mathcal{U}_Y$.  Each Euler step is an admissible residual layer,
    since $\Delta_k H_k\in\mathcal{G}$ by positive scaling.

    \emph{Step 3: concatenate the Euler and correction layers.}
    Step~2 gives an Euler endpoint $\widetilde Y\in\mathcal U_Y$.
    Set $a=\alpha(\widetilde Y)$ and append the $L_Y$ residual
    layers defining $E_q(\cdot,a)$ to this Euler network.  Their
    fields all belong to $\mathcal G$: each is a fixed real scalar
    multiple of one of the fields selected in Step~1.  By
    \eqref{eq:proposed-lie-residual-correction}, the final output is
    $E_q(\widetilde Y,a)=Y$, and all intermediate configurations
    remain in $\mathcal M$.  Together with the $L_E$ Euler layers,
    this gives a finite admissible network with at most $L_E+L_Y$
    layers.  All its scalar coefficients are fixed for the
    prescribed interpolation problem.  In a batch application,
    these scalars are shared across all samples.
\end{proof}

\section{Universal finite-depth interpolation}
\label{app:universal-finite-depth-proof}

\subsection{Direct spanning}
\label{app:universal-finite-depth-direct-spanning}

\begin{proposition}[Universal finite-depth interpolation]
\label{prop:universal-finite-depth-interpolation}
    Fix $N\geq1$ and a configuration space $\mathcal{M}_N$ as in
    \Cref{def:universal-interpolation}, and assume that $\mathcal{F}$
    allows real scalar multiplication.  Suppose there exist finitely
    many parameters $\theta_1,\ldots,\theta_m$ such that, at each
    $X\in\mathcal{M}_N$, one can select $Np$ of their lifted fields
    that are $C^1$ near $X$ and whose values at $X$ form a basis of
    $\mathbb{R}^{Np}$.  Then $\mathcal{F}$ has universal interpolation
    at finite depth on $\mathcal{M}_N$, using these same parameters.
\end{proposition}

\begin{proof}
    Fix the parameters $\theta_1,\ldots,\theta_m$ in the hypothesis
    and set
    \[
        \mathcal{G}_N
        \coloneqq\{a f_{\theta_j}^{(N)}:a\in\mathbb{R},\ 1\leq j\leq m\}.
    \]
    Every field in $\mathcal G_N$ is locally Lipschitz on
    $\mathbb R^{Np}$, hence also on $\mathcal M_N$.
    The family is closed under real scalar multiplication and,
    on $\mathcal M_N$, satisfies the pointwise basis condition of
    \Cref{cor:pointwise-full-rank-criterion} with $D=Np$.  Thus
    \Cref{prop:finite-depth-interpolation} joins every pair of
    configurations by residual layers
    $\operatorname{Id}+c_k f_{\theta_{j_k}}^{(N)}$, with all intermediate
    configurations in $\mathcal{M}_N$.  The parameters were fixed
    before the endpoints; only the depth, indices, and scalar
    coefficients depend on them.  This is universal interpolation
    at finite depth as defined in \Cref{sec:finite-depth}.
\end{proof}

\subsection{Extension under a Lie-rank condition}
\label{app:universal-finite-depth-lie-rank}

\propuniversalfinitedepthlierank*

\begin{proof}
    Fix the parameters $\theta_1,\ldots,\theta_m$ in the hypothesis
    and set
    \[
        \mathcal G_N
        \coloneqq\{c f_{\theta_j}^{(N)}:c\in\mathbb R,\ 1\leq j\leq m\}.
    \]
    This family is locally Lipschitz and closed under real scalar
    multiplication.  At each $X\in\mathcal M_N$, the fields indexed
    by $I_X$ form a finite subfamily that is smooth near $X$ and
    has full Lie rank there.  Thus
    \Cref{prop:finite-depth-local-lie-rank-proposal}, with $D=Np$,
    joins every pair $X,Y\in\mathcal M_N$ by finitely many residual
    layers
    \[
        \operatorname{Id}+c_k f_{\theta_{j_k}}^{(N)},
        \qquad c_k\in\mathbb R,\quad j_k\in\{1,\ldots,m\},
    \]
    with all intermediate configurations in $\mathcal M_N$.
    The parameters were fixed before the endpoints; only the depth,
    indices, and scalar coefficients depend on them.  This is
    universal interpolation at finite depth using the same dictionary.
\end{proof}

\section{Residual networks}\label{sec:resnet}

For $N\geq1$ and $d\geq2$, define the generic configuration space
\[
    \mathcal{C}_{N,d}
    \coloneqq
    \left\{(x_1,\ldots,x_N)\in(\mathbb{R}^d)^N:
    x_i\neq 0\ \text{and}\ x_i\notin\mathbb{R}x_j\ \text{for all }i\neq j\right\}.
\]
It is nonempty and open: distinct nonzero projective directions give
examples, while the excluded zero vectors and collinear pairs form a
finite union of closed sets.

\begin{theorem}[Exact interpolation by bias-free ResNets]\label{thm:resnet-uit}
    Let $d\geq 3$ and $r\geq 1$.  For every $N\geq1$, the family of
    vector fields in \eqref{eq:vectorfields_ResNets} has the exact
    interpolation property on $\mathcal{C}_{N,d}$, both at continuous
    depth in the sense of \Cref{def:exact-interpolation} and at finite
    depth in the sense of \Cref{sec:finite-depth}.  One active hidden
    unit per layer suffices, and every intermediate configuration
    remains in $\mathcal{C}_{N,d}$.
\end{theorem}

The proof is given in \Cref{app:resnet-proof}.

\begin{remark}
    The assumption $d\geq3$ in \Cref{thm:resnet-uit} ensures that
    $\mathcal{C}_{N,d}$ is connected.  In dimension two,
    $\mathcal{C}_{N,2}$ is not connected when $N\geq2$, but exact
    interpolation at both continuous and finite depth holds on each
    connected component.  These claims are proved in
    \Cref{app:resnet-proof}.
\end{remark}

With identity activation, both continuous- and finite-depth networks
apply a common linear map to all samples and preserve their linear
relations.  For $N>d$, the lifted Lie algebra has rank at most $d^2<Nd$,
so the analytic converse in \Cref{cor:exact-interpolation-criterion}
rules out continuous-depth exact interpolation on any nonempty open
subset of $(\mathbb{R}^d)^N$.  The preserved linear relations give the
same obstruction at finite depth, thereby also excluding universal
interpolation at both depths.  The proof is given in \Cref{app:linear-proof}.
The following theorem establishes universal interpolation for bias-free
ReLU ResNets on $\mathcal{C}_{N,d}$, using a fixed family of updates
that works for every $N$.

\begin{theorem}[Universal interpolation by bias-free ResNets]
    \label{thm:resnet-universal-interpolation}
    Let $d\geq3$ and $r\geq1$, and let $e_1,\ldots,e_d$ be the standard
    basis of $\mathbb{R}^d$.
    Define the fixed family
    \begin{equation}\label{eq:resnet-fixed-family}
        \mathcal{R}_d
        \coloneqq
        \left\{g_{ij}^{\varepsilon}:x\mapsto
        e_i\sigma(\varepsilon x_j):
        1\leq i,j\leq d,\ \varepsilon\in\{-1,1\}\right\}.
    \end{equation}
    For every $N\geq1$, the ResNet family in
    \eqref{eq:vectorfields_ResNets} has universal interpolation on
    $\mathcal{C}_{N,d}$, both at continuous depth in the sense of
    \Cref{def:universal-interpolation} and at finite depth in the sense
    of \Cref{sec:finite-depth}, using the same $2d^2$ generators from
    $\mathcal{R}_d$.  At continuous depth, each generator may be used
    in either sign.  These generators are independent of $N$ and
    require only one active hidden unit per layer.
\end{theorem}

The proof in \Cref{app:resnet-universal-proof} realizes the flows used in
\Cref{thm:resnet-uit} by composing flows from the fixed family
\eqref{eq:resnet-fixed-family}.  Each such flow is itself exactly a
residual layer with a suitable scalar gain.

Fixed families giving exact interpolation are known for general neural
ODEs~\citep{cuchiero2020deep,grong2026controllability} and for certain
smooth-activation ResNets~\citep{tabuada2023universal}.
For ReLU, \citet{duan2025minimal} prove universal
approximation using a finite family of linear fields, translations,
and ReLU, while \citet{cai2025achieving} prove exact
interpolation using affine fields and ReLU.  \citet{cai2024vocabulary}
also constructs a finite vocabulary of maps for universal approximation.
\Cref{thm:resnet-universal-interpolation} gives an explicit family of
bias-free, width-one ReLU updates, reusable for every $N$, with exact
interpolation at both continuous and finite depth.

\begin{remark}[Comparison with interpolation using biases]
    \label{rem:resnet-biased-interpolation}
    \citet{ruizbalet2023neural}
    prove exact interpolation with biases for $d\geq2$, using one active
    neuron at a time.  Their admissible configurations require only
    pairwise distinct points, allowing zero vectors and collinear pairs.
    Their construction chooses the weights according to the dataset
    and does not assert interpolation with a fixed finite family.
    It appears to adapt directly to universal interpolation by adding
    coordinate translations to \eqref{eq:resnet-fixed-family} and
    replacing its data-dependent flows by affine conjugations of fixed
    ReLU flows.  These translations are unavailable in the bias-free
    architecture considered here.
\end{remark}

\subsection{Linear activation and the obstruction to interpolation}\label{app:linear-proof}

The following lemma also applies to the shared linear dynamics of
attention with sequence length $n=1$.

\begin{lemma}[Linear-relation obstruction]\label{lem:linear-obstruction}
    Let $d\geq1$, $N>d$, and let
    $\mathcal{M}\subseteq(\mathbb{R}^d)^N$ be nonempty and open.
    For every $X\in\mathcal{M}$, there are arbitrarily close targets
    $Y\in\mathcal{M}$ in the same connected component that cannot be
    obtained from $X$ by a shared linear map.  Thus neither controlled
    linear ODEs nor finite compositions of linear residual layers can
    interpolate every pair of configurations in $\mathcal{M}$.
\end{lemma}

\begin{proof}
    For $A\in\mathbb{R}^{d\times d}$, define $f_A(x)=Ax$ and its lift
    $F_A(X)=(Ax^1,\ldots,Ax^N)$.  The bracket convention gives
    \[
        [F_A,F_B]=F_{BA-AB}.
    \]
    Therefore, for any family of linear fields,
    \[
        \operatorname{Lie}\{F_{A_\theta}:\theta\in\Theta\}(X)
        \subseteq
        \{(Ax^1,\ldots,Ax^N):A\in\mathbb{R}^{d\times d}\}.
    \]
    The space on the right is the image of a linear map from
    $\mathbb{R}^{d\times d}$, so its dimension is at most $d^2<Nd$.

    To establish the obstruction for both depth conventions, observe that
    a continuous linear network applies a common fundamental matrix to
    all samples, and a finite linear residual network applies the common
    matrix $(I+A_{L-1})\cdots(I+A_0)$.  In either case, write $A_*$
    for the common output matrix.  Every input linear relation is
    retained at the output:
    \[
        \sum_{i=1}^N c_i x^i=0
        \quad\Longrightarrow\quad
        \sum_{i=1}^N c_i A_*x^i= A_* \sum_{i=1}^N c_i x^i =0.
    \]
    Fix any $X=(x^1,\ldots,x^N)\in\mathcal{M}$.
    Since $N>d$, choose a nonzero relation $\sum_i c_i x^i=0$ and an
    index $j$ with $c_j\neq0$.  Choose any nonzero
    $u\in\mathbb{R}^d$ and set $y_j=x_j+\varepsilon u$ and $y_i=x_i$
    for $i\neq j$.  For sufficiently small $\varepsilon\neq0$, openness
    of $\mathcal{M}$ ensures that the segment from $X$ to $Y$ lies in
    $\mathcal{M}$.  Thus $Y$ is in the same connected component as $X$,
    but
    \[
        \sum_i c_i y_i=\varepsilon c_j u\neq0.
    \]
    No shared linear map can send $X$ to $Y$.  Taking arbitrarily small
    nonzero $\varepsilon$ proves the lemma.
\end{proof}

For ResNets with identity activation, the vector fields take the form
\[
    f_{U,V}(x)=U\sigma(V^\top x)=UV^\top x,
    \qquad U,V\in\mathbb{R}^{d\times r}.
\]
Since \Cref{lem:linear-obstruction} applies to arbitrary linear fields
$x\mapsto Ax$, it applies in particular to $A=UV^\top$, for every hidden
width $r$.  Thus, when $N>d$, increasing the hidden width cannot remove
the obstruction to interpolation, at either continuous or finite depth.

\subsection{Proof of \texorpdfstring{\Cref{thm:resnet-uit}}{the ResNet theorem}}\label{app:resnet-proof}

\begin{proof}
    We first prove the stronger statement that exact interpolation holds
    at both continuous and finite depth on every connected component of
    $\mathcal{C}_{N,d}$ for all $d\geq2$.  For $X=(x^1,\ldots,x^N)$, write
    \[
        f^{(N)}_{U,V}(X)
        \coloneqq
        \bigl(U\sigma(V^\top x^1),\ldots,U\sigma(V^\top x^N)\bigr).
    \]
    The width-one subfamily
    \[
        G_{u,v}(X)
        \coloneqq
        \bigl(u\sigma(v^\top x^1),\ldots,u\sigma(v^\top x^N)\bigr),
        \qquad u,v\in\mathbb{R}^d,
    \]
    is contained in the width-$r$ family for every $r\geq 1$: take
    $U=(u,0,\ldots,0)$ and $V=(v,\ldots,v)$.

    Fix $X\in\mathcal{C}_{N,d}$ and set
    \[
        \mathcal{H}_i\coloneqq\{v\in\mathbb{R}^d:v^\top x^i=0\},
        \qquad
        \mathcal{O}_X\coloneqq
        \mathbb{R}^d\setminus\bigcup_{j=1}^N\mathcal{H}_j.
    \]
    The hyperplanes $\mathcal{H}_1,\ldots,\mathcal{H}_N$ are pairwise distinct.  Hence, for each $i$, we may choose
    \[
        w\in\mathcal{H}_i\setminus\bigcup_{j\neq i}\mathcal{H}_j.
    \]
    For sufficiently small $\delta>0$, the four vectors
    \[
        v_k\coloneqq w+k\delta x^i,
        \qquad k\in\{-2,-1,1,2\},
    \]
    all belong to $\mathcal{O}_X$, and, for $j\neq i$, the signs of $v_k^\top xj$ do not depend on $k$.  Therefore, for every $u\in\mathbb{R}^d$,
    \begin{equation}\label{eq:resnet-finite-difference}
    \begin{split}
        &G_{u,v_2}(X)-G_{u,v_1}(X)
        -G_{u,v_{-1}}(X)+G_{u,v_{-2}}(X) \\
        &\hspace{2cm}
        =\delta\lVert x^i\rVert^2
        (0,\ldots,0,u,0,\ldots,0),
    \end{split}
    \end{equation}
    where $u$ occupies the $i$-th block on the right-hand side.  Indeed, in every block $j\neq i$, the ReLU is either zero or affine as a function of $k$, so its finite difference vanishes.  In the $i$-th block, $v_k^\top x^i=k\delta\lVert x^i\rVert^2$, and the finite difference is $\delta\lVert x^i\rVert^2u$.

    Since $i$ and $u$ are arbitrary, \eqref{eq:resnet-finite-difference} gives
    \begin{equation}\label{eq:resnet-full-rank}
        \operatorname{span}
        \left\{f^{(N)}_{U,V}(X):U,V\in\mathbb{R}^{d\times r}\right\}
        =(\mathbb{R}^d)^N.
    \end{equation}
    In fact, the width-one subfamily $\{G_{u,v}\}$ already has this full
    pointwise span.  Moreover, every field used in
    \eqref{eq:resnet-finite-difference} is smooth on a neighborhood of $X$,
    because $v_k^\top x^j\neq 0$ for all $j$.  Thus, at each
    $X\in\mathcal{C}_{N,d}$, we may select $Nd$ locally smooth width-one
    fields whose values form a basis of $(\mathbb{R}^d)^N$.

    The lifted ResNet family is closed under negation:
    \[
        -f^{(N)}_{U,V}=f^{(N)}_{-U,V}.
    \]
    Since $\mathcal{C}_{N,d}$ is open, \Cref{cor:pointwise-full-rank-criterion}
    and \eqref{eq:resnet-full-rank} give exact interpolation at continuous
    depth on every connected component of $\mathcal{C}_{N,d}$.
    For finite depth, the family is also closed under real scalar
    multiplication, since $af^{(N)}_{U,V}=f^{(N)}_{aU,V}$.
    The locally smooth width-one fields selected above therefore satisfy
    the hypotheses of \Cref{prop:finite-depth-interpolation}, with
    $D=Nd$.  It gives exact interpolation at finite depth, with every
    intermediate configuration in the same component.  Each scalar
    gain is absorbed into the output matrix $U$.

    It remains to prove that $\mathcal{C}_{N,d}$ is connected when $d\geq 3$.  Write $x^i=r_i u_i$, where $r_i>0$ and $u_i\in\mathbb{S}^{d-1}$.  This identifies $\mathcal{C}_{N,d}$ with
    \[
        (0,\infty)^N\times\mathcal{D}_{N,d},
        \qquad
        \mathcal{D}_{N,d}
        \coloneqq
        \left\{(u_1,\ldots,u_N)\in(\mathbb{S}^{d-1})^N:
        u_i\neq\pm u_j\text{ for }i\neq j\right\}.
    \]
    Let $u=(u_1,\ldots,u_N)$ and $\widetilde u=(\widetilde u_1,\ldots,\widetilde u_N)$ belong to $\mathcal{D}_{N,d}$.  Choose points $p_1,\ldots,p_N\in\mathbb{S}^{d-1}$ inductively, with each $p_i$ different from the finitely many points
    \[
        \{\pm u_j,\pm\widetilde u_j:1\leq j\leq N\}
        \cup
        \{\pm p_j:1\leq j<i\}.
    \]
    Because $d-1\geq 2$, a sphere with finitely many points removed is path connected.  We may consequently move $u_1,\ldots,u_N$, one at a time, to $p_1,\ldots,p_N$ while avoiding the directions and antipodal directions of the points kept fixed.  Moving the $p_i$ one at a time to $\widetilde u_i$ gives a path in $\mathcal{D}_{N,d}$ from $u$ to $\widetilde u$.  The factor $(0,\infty)^N$ is path connected as well.  Hence $\mathcal{C}_{N,d}$ is path connected for $d\geq 3$, and the componentwise result proved above yields the theorem.
    For $d=2$ and $N\geq2$, the continuous function
    $X\mapsto\det(x^1,x^2)$ never vanishes on $\mathcal{C}_{N,2}$ by definition of $\mathcal{C}_{N,2}$,
    and takes both signs, so this space is disconnected.
\end{proof}

\subsection{Proof of \texorpdfstring{\Cref{thm:resnet-universal-interpolation}}{the universal ResNet interpolation theorem}}\label{app:resnet-universal-proof}

\begin{proof}
    Fix $d\geq3$, $N\geq1$, and the family $\mathcal R_d$ in the
    statement.  Write $\phi_g^t$ for the flow of a field $g$, acting
    componentwise on configurations.  For
    $g(x)=u\sigma(v^\top x)$ and $\alpha=v^\top u$, this flow is defined
    for every $t\in\mathbb R$ and satisfies
    \begin{equation}\label{eq:resnet-gate-flow}
        \phi_g^t(x)=x+\beta_\alpha(t)g(x),
        \qquad
        \beta_\alpha(t)=
        \begin{cases}
            (e^{\alpha t}-1)/\alpha,&\alpha\neq0,\\
            t,&\alpha=0.
        \end{cases}
    \end{equation}
    Indeed, $q(t)=v^\top\phi_g^t(x)$ solves
    $\dot q=\alpha\sigma(q)$: it remains constant when $q(0)\leq0$
    and equals $e^{\alpha t}q(0)$ otherwise.  Integrating the equation
    for $\phi_g^t(x)$ gives \eqref{eq:resnet-gate-flow}.

    We first show that the fixed family can realize every linear map
    $A\in GL^+(d)\coloneqq\{A:\det A>0\}$ while keeping a prescribed
    configuration in $\mathcal C_{N,d}$.  Put $E_{ij}=e_i e_j^\top$.
    The flows of $g_{ij}^{+1}$ and $g_{ij}^{-1}$ commute: for $i\neq j$
    they preserve $x_j$, and for $i=j$ they act on separate invariant
    half-lines.  Since
    $g_{ij}^{+1}(x)-g_{ij}^{-1}(x)=E_{ij}x$, we have
    \begin{equation}\label{eq:resnet-coordinate-flow}
        \phi_{g_{ij}^{+1}}^t\circ\phi_{g_{ij}^{-1}}^{-t}(x)
        =e^{tE_{ij}}x.
    \end{equation}
    Every matrix in $GL^+(d)$ is a finite product of these exponentials.
    To see this, the derivative at zero of
    $(t_{ij})\mapsto\prod_{i,j}\exp(t_{ij}E_{ij})$, with any fixed
    ordering of the factors, is the isomorphism
    $(t_{ij})\mapsto\sum_{i,j}t_{ij}E_{ij}$.  The inverse function
    theorem shows that the subgroup generated by these exponentials
    contains a neighborhood of the identity.  It is therefore open,
    and connectedness of $GL^+(d)$ implies that it is the whole group.

    Although the endpoints in \eqref{eq:resnet-coordinate-flow} are
    related by a linear map, its two intermediate flows need not
    preserve noncollinearity.  This is resolved by subdivision.
    For $X\in\mathcal C_{N,d}$, the linear path
    $\Gamma=\{e^{stE_{ij}}X:0\leq s\leq1\}$ is a compact subset of
    $\mathcal C_{N,d}$.  Choose $\delta>0$ such that its
    $\delta$-neighborhood is contained in $\mathcal C_{N,d}$, and put
    $M=\max_{Z\in\Gamma}\|Z\|$.  Replace the pair in
    \eqref{eq:resnet-coordinate-flow} by $K$ repetitions with
    $h=t/K$.  Their successive endpoints are exactly
    $e^{khE_{ij}}X$.  Every lifted field in $\mathcal R_d$ satisfies
    $\|g^{(N)}(Z)\|\leq\|Z\|$.  Hence, during either subflow of a
    pair, Gronwall's inequality bounds the distance from the pair's
    initial point on $\Gamma$ by
    $M(e^{2|h|}-1)$.  For sufficiently large $K$, this is less than
    $\delta$.  All intermediate configurations then stay in
    $\mathcal C_{N,d}$.  Applying this construction to each exponential
    factor realizes $A$ with finitely many flows from the same fixed
    family, with the required constraint throughout.

    Next, every width-one field $g(x)=u\sigma(v^\top x)$ is conjugate,
    up to a real scalar, to one of
    \[
        H(x)=e_1\sigma(x_1),\qquad T(x)=e_2\sigma(x_1),
    \]
    which belong to $\mathcal R_d$.  Zero fields require no flow, so
    assume $u,v\neq0$.  If $\alpha=v^\top u\neq0$, choose
    $A\in GL^+(d)$ whose first column is $u/\alpha$ and whose remaining
    columns form a basis of $\ker v^\top$.  Reversing one of those
    columns if necessary makes the determinant positive.  Then
    $e_1^\top A^{-1}=v^\top$ and
    \[
        g(x)=\alpha A H(A^{-1}x),\qquad
        \phi_g^t=A\circ\phi_H^{\alpha t}\circ A^{-1}.
    \]
    If $\alpha=0$, choose the first column $a_1$ so that
    $v^\top a_1=1$, take the second column to be $u$, and complete
    these columns to a basis using vectors in $\ker v^\top$.
    Since $d\geq3$, the sign of a further column can be chosen to
    ensure $\det A>0$.  This gives
    \[
        g(x)=A T(A^{-1}x),\qquad
        \phi_g^t=A\circ\phi_T^t\circ A^{-1}.
    \]
    Suppose a segment of the flow of $g$ starting at $X$ remains in
    $\mathcal C_{N,d}$.  Realize $A^{-1}$ and $A$ by the admissible
    compositions just constructed.  Between them, the flow of $H$ or
    $T$ is the image under $A^{-1}$ of the prescribed segment, so it
    also stays in $\mathcal C_{N,d}$.  Thus that segment can be
    replaced exactly by finitely many flows from $\mathcal R_d$,
    without leaving the configuration space.

    By \Cref{thm:resnet-uit} at width one, any
    $X,Y\in\mathcal C_{N,d}$ are joined inside this set by a finite
    concatenation of flows of fields $x\mapsto u\sigma(v^\top x)$.  Replacing
    each segment as above gives a finite concatenation of flows from
    $\mathcal R_d$ with signed times $\tau_1,\ldots,\tau_L$.
    Discard zero-time segments and run each remaining segment on the batch with
    $\operatorname{sgn}(\tau_k)g_k^{(N)}$ for duration $|\tau_k|$, where
    $g_k\in\mathcal R_d$ is its generator.  With
    $S=\sum_k|\tau_k|$, this gives a piecewise-constant control in
    $\{\pm g^{(N)}:g\in\mathcal R_d\}$ on $[0,S]$, whose trajectory
    satisfies $Z(0)=X$, $Z(S)=Y$ and remains in
    $\mathcal C_{N,d}$.  If no segment remains, a sufficiently short
    flow of any generator followed by its reverse gives the identity
    with $S>0$ and stays in this open set.  The $2d^2$ generators were
    fixed before both $N$ and the endpoints, proving the
    continuous-depth conclusion in \Cref{def:universal-interpolation}.

    Finally, \eqref{eq:resnet-gate-flow} turns each of these flows
    exactly into a residual layer
    $\operatorname{Id}+\beta_{\varepsilon\delta_{ij}}(\tau_k)
    g_{ij}^{\varepsilon}$.  The resulting finite network uses the
    same family and has the same intermediate configurations at the
    layer endpoints.  This proves universal interpolation at finite
    depth directly, without a discretization error.  All blocks embed
    into every hidden width $r\geq1$ by setting unused output columns
    to zero.
\end{proof}

The preceding proof uses changes of activation patterns along the
trajectory.  The local condition of
\Cref{cor:universal-interpolation-criterion} is not satisfied by this
fixed family when $N>d$: when all inputs lie in one open orthant, its
fields are locally linear, so their lifted Lie algebra has rank at most
$d^2<Nd$.  More generally,
the same observation applies to any finite family of bias-free ReLU
blocks on an open cone where all their activation patterns are constant.
This does not contradict the analytic converse in
\Cref{rem:analytic-orbit}, since these fields are not analytic across
their activation hyperplanes.

\section{Proof of \texorpdfstring{\Cref{thm:attention-uit}}{the self-attention theorem}}\label{app:attention-proof}

\thmattentionuit*

\begin{proof}
    The pointwise rank argument below only requires $d\geq2$.  We will
    also prove exact interpolation at both continuous and finite depth
    on every connected component of $\mathcal{A}_{N,n,2}$.  Fix
    $N\geq1$ and $d\geq2$.  The set $\mathcal{A}_{N,n,d}$ is open: its
    complement is a finite union of the closed sets where one token is zero or
    where all $2\times2$ minors of a pair of token columns vanish.  Hence its
    connected components are open.  Take
    $X=(x_i^\ell)_{\ell,i}\in\mathcal{A}_{N,n,d}$.  It is enough to
    use one head: set $V_h=0$ for $h\geq2$, and denote the effective matrices
    of the remaining head by $W,V\in\mathbb{R}^{d\times d}$.  We will
    further restrict both matrices to have rank at most one.  Let
    $f^{(N)}_{W,V}$ be the corresponding lifted vector field and define its
    pointwise span
    \[
        \mathcal{S}_{X}
        \coloneqq
        \operatorname{span}
        \left\{f^{(N)}_{W,V}(X):
        \operatorname{rank}W,\operatorname{rank}V\leq1\right\}
        \subseteq(\mathbb{R}^d)^{Nn}.
    \]
    This is a finite-dimensional linear subspace and is therefore closed.

    For $b\in\mathbb{R}^d$ and $1\leq\ell\leq N$, introduce the softmax
    barycenter
    \begin{equation}\label{eq:attention-barycenter}
        z_\ell(b)
        \coloneqq
        \frac{\sum_{j=1}^n
        \exp\!\bigl(\langle b,x_j^\ell\rangle\bigr)x_j^\ell}
        {\sum_{j=1}^n
        \exp\!\bigl(\langle b,x_j^\ell\rangle\bigr)}.
    \end{equation}
    The $(i,\ell)$ block of the lifted field is then
    \[
        f^{(N)}_{W,V}(X)_{i,\ell}
        =V z_\ell(Wx_i^\ell).
    \]

    We show that an arbitrary block can be controlled independently.  Fix a
    target $(i_0,\ell_0)$ and put $q\coloneqq x_{i_0}^{\ell_0}$.  First choose
    $a\in\mathbb{R}^d$ such that
    \begin{equation}\label{eq:attention-separating-a}
        \langle a,x_j^\ell\rangle
        \neq
        \langle a,x_k^\ell\rangle
        \qquad
        (1\leq\ell\leq N,\ j\neq k).
    \end{equation}
    Such an $a$ exists: each forbidden equality defines the proper hyperplane
    $(x_j^\ell-x_k^\ell)^\perp$, and only finitely many hyperplanes are
    excluded.

    Next choose $\alpha\in q^\perp$ satisfying
    \begin{equation}\label{eq:attention-separating-alpha}
        \langle\alpha,x_i^\ell\rangle\neq0
        \qquad\text{whenever }(i,\ell)\neq(i_0,\ell_0).
    \end{equation}
    Indeed, every other token $r=x_i^\ell$ is not collinear with $q$, so
    $q^\perp\cap r^\perp$ is a proper linear subspace of $q^\perp$.  A finite
    union of proper linear subspaces cannot cover $q^\perp$.

    Let $V\in\mathbb{R}^{d\times d}$ have rank at most one; its value
    will be chosen below.  For $t>0$, define the rank-one score matrices
    \begin{equation}\label{eq:attention-selector-matrices}
        W_t^\pm
        \coloneqq
        a\left(t\alpha\pm\frac{q}{\lVert q\rVert^2}\right)^\top.
    \end{equation}
    Since $\alpha\perp q$, they satisfy $W_t^\pm q=\pm a$.  Consequently, at
    the target block,
    \[
        f^{(N)}_{W_t^+,V}(X)_{i_0,\ell_0}
        -f^{(N)}_{W_t^-,V}(X)_{i_0,\ell_0}
        =V c,
        \qquad
        c\coloneqq z_{\ell_0}(a)-z_{\ell_0}(-a).
    \]
    The vector $c$ is nonzero.  To see this, let
    \[
        \psi(s)
        \coloneqq
        \log\left(
        \sum_{j=1}^n\exp\!\bigl(s\langle a,x_j^{\ell_0}\rangle\bigr)
        \right).
    \]
    Then
$        \psi'(s)=\langle a,z_{\ell_0}(sa)\rangle,
$ and 
\begin{align*}
        \psi''(s)
        &= \sum_{j=1}^n
p_j(s)\,\langle a,x_j^{\ell_0}\rangle^2
-
\left(
\sum_{j=1}^n
p_j(s)\,\langle a,x_j^{\ell_0}\rangle
\right)^2 \\
&= \sum_{j=1}^n
p_j(s)
\left(
\langle a,x_j^{\ell_0}\rangle
-
\sum_{k=1}^n
p_k(s)\,\langle a,x_k^{\ell_0}\rangle
\right)^2 > 0,
    \end{align*}
    where $p_j(s) :=
\frac{
e^{s\langle a,x_j^{\ell_0}\rangle}
}{
\sum_{k=1}^n e^{s\langle a,x_k^{\ell_0}\rangle}
}. $ Strict positivity follows from
    $n\geq2$ and \eqref{eq:attention-separating-a}.  Hence
    \[
        \langle a,c\rangle=\psi'(1)-\psi'(-1)>0.
    \]

    Consider now a non-target query $r=x_i^\ell$.  From
    \eqref{eq:attention-selector-matrices},
    \[
        W_t^\pm r
        =\lambda_t^\pm(r)a,
        \qquad
        \lambda_t^\pm(r)
        \coloneqq
        t\langle\alpha,r\rangle
        \pm\frac{\langle q,r\rangle}{\lVert q\rVert^2}.
    \]
    By \eqref{eq:attention-separating-alpha}, both
    $\lambda_t^+(r)$ and $\lambda_t^-(r)$ tend to the same signed infinity.
    If the sign is positive, both softmax barycenters converge to the unique
    key in sequence $\ell$ maximizing $\langle a,x_j^\ell\rangle$; if it is
    negative, both converge to the unique minimizing key.  The uniqueness is
    guaranteed by \eqref{eq:attention-separating-a}.  Therefore
    \[
        z_\ell(W_t^+r)-z_\ell(W_t^-r)\longrightarrow0
        \qquad(t\to\infty)
    \]
    at every non-target block.

    Let $E_{i_0,\ell_0}(u)$ denote the vector in $(\mathbb{R}^d)^{Nn}$ whose
    $(i_0,\ell_0)$ block equals $u$ and whose other blocks vanish.  The preceding
    limits give
    \[
        f^{(N)}_{W_t^+,V}(X)
        -f^{(N)}_{W_t^-,V}(X)
        \longrightarrow E_{i_0,\ell_0}(Vc).
    \]
    Every vector on the left belongs to $\mathcal{S}_{X}$, and this
    subspace is closed.  Given any $u\in\mathbb{R}^d$, choose the value
    matrix of rank at most one
    \[
        V\coloneqq\frac{u c^\top}{\lVert c\rVert^2},
    \]
    for which $Vc=u$.  Thus $E_{i_0,\ell_0}(u)$ belongs to
    $\mathcal{S}_{X}$.  Since the target block and $u$ were arbitrary,
    \begin{equation}\label{eq:attention-full-rank}
        \mathcal{S}_{X}=(\mathbb{R}^d)^{Nn}.
    \end{equation}
\end{proof}

\begin{theorem}[Exact interpolation by noncausal self-attention]
\label{thm:attention-exact-interpolation}
    Let $d\geq3$, $n\geq2$, and $H\geq1$.  For every $N\geq1$, the
    noncausal self-attention family in \eqref{eq:vectorfields_att} has
    the exact interpolation property on $\mathcal{A}_{N,n,d}$, both at
    continuous depth in the sense of \Cref{def:exact-interpolation} and
    at finite depth in the sense of \Cref{sec:finite-depth}.
    The conclusion holds for any prescribed positive head widths.
    The token dimension remains $d$, and every intermediate
    configuration remains in $\mathcal{A}_{N,n,d}$.
\end{theorem}

\begin{proof}
    Fix $N\geq1$ and $X\in\mathcal A_{N,n,d}$.
    The attention vector fields are analytic, because every softmax denominator
    is strictly positive.  Negating all value matrices negates the
    field, so the family is closed under negation.
    Equation \eqref{eq:attention-full-rank} supplies a
    finite collection of actual rank-one attention fields whose values form
    a basis at $X$.  In particular, the limiting argument only
    establishes a full finite-dimensional span: the selected basis fields
    have finite parameter values, and no infinite-temperature layer is used.
    Since $X$ was arbitrary,
    \Cref{cor:pointwise-full-rank-criterion} gives exact interpolation
    at continuous depth on every connected component of $\mathcal{A}_{N,n,d}$.
    Multiplying the value matrix by any real scalar scales the field
    by that scalar, so the same rank-one family satisfies
    \Cref{prop:finite-depth-interpolation}, with $D=Nnd$, and therefore
    gives exact interpolation at finite depth, with every intermediate
    configuration in the same component.  Absorbing the scalar gains
    into $V$ preserves its rank.  Rank-one factorizations of $W,V$
    realize each layer with a head of width one; setting unused value
    matrices to zero embeds it into every prescribed $H$-head architecture.

    It remains to determine these components.  Put $m\coloneqq Nn$.  Writing
    each token as $x_k=r_ku_k$, with
    $r_k>0$ and $u_k\in\mathbb{S}^{d-1}$, gives a homeomorphism
    \[
        \mathcal{A}_{N,n,d}
        \simeq
        (0,\infty)^m\times\mathcal{D}_{m,d},
        \qquad
        \mathcal{D}_{m,d}
        \coloneqq
        \left\{(u_1,\ldots,u_m)\in(\mathbb{S}^{d-1})^m:
        u_k\neq\pm u_l\text{ for }k\neq l\right\}.
    \]
    For $d\geq3$, the path-connectedness argument in
    \Cref{app:resnet-proof} applies with $N$ replaced by $m$.
    Hence $\mathcal{A}_{N,n,d}=\mathcal{C}_{m,d}$ is path connected,
    proving the theorem.

\end{proof}

\section{Uniform direct spanning on the generic attention domain}
\label{app:attention-generic-generation}

Fix an $H$-head architecture with positive integer widths
$r_1,\ldots,r_H$.  Its full projection parameter space is
\begin{equation}\label{eq:attention-projection-parameter-space}
    \Theta=\prod_{h=1}^H
    \left((\mathbb R^{r_h\times d})^3\times
    \mathbb R^{d\times r_h}\right)\simeq\mathbb R^p,
    \qquad p=4d\sum_{h=1}^H r_h.
\end{equation}
Write $\theta=(Q_h,K_h,U_h,O_h)_{h=1}^H$, set
$W_h=K_h^\top Q_h/\sqrt{r_h}$ and $V_h=O_hU_h$, and denote the
corresponding batch field by $F_\theta=f^{(N)}_{\mathbf W,\mathbf V}$.
Thus $\Theta$ parametrizes the query, key, value, and output projections,
not independent effective matrices $W_h,V_h$.

\begin{lemma}[Uniform direct spanning on the generic domain]
\label{lem:attention-uniform-direct-generation}
    Let $d\geq3$, $n\geq2$, and $N\geq1$, and put $D\coloneqq Nnd$.
    Draw $(\theta_1,\ldots,\theta_{2D})$ from any jointly absolutely
    continuous law on $\Theta^{2D}$.
    Then, with probability one over this single draw,
    \begin{equation}\label{eq:attention-uniform-direct-rank}
        \operatorname{span}\left\{
        F_{\theta_k}(X):1\leq k\leq2D
        \right\}
        =\mathbb{R}^D
        \qquad
        \text{simultaneously for every }
        X\in\mathcal{A}_{N,n,d}.
    \end{equation}
    Thus, the same frozen finite dictionary is full rank at every generic
    configuration.  In particular, it satisfies the uniform Lie-rank
    condition without using any Lie brackets, with
    $m_N\leq2D=2Nnd$.
\end{lemma}

We first isolate the general principle behind this lemma.
\citet[Theorem~A.2]{amir2023neural} establish a more general
finite-witness theorem using tools from o-minimal geometry.  For
accessibility, we state only its consequence for jointly real analytic
maps on open Euclidean domains, which covers all applications below.
We apply their argument to covectors annihilating the sampled fields
and include the resulting dimension-counting proof.

\begin{theorem}[Uniform random spanning principle]
\label{thm:uniform-random-spanning-principle}
    Let $p,q,r\geq1$.  Let $\Theta\subseteq\mathbb{R}^p$ be a nonempty,
    connected open set, and let $\mathcal{X}\subseteq\mathbb{R}^q$ be
    a nonempty open set.  Suppose that
    $G:\Theta\times\mathcal{X}\to\mathbb{R}^r$ is jointly
    real analytic and that
    \begin{equation}\label{eq:general-pointwise-spanning}
        \operatorname{span}\{G(\theta,X):\theta\in\Theta\}
        =\mathbb{R}^r
        \qquad
        \text{for every }X\in\mathcal{X}.
    \end{equation}
    For every $m\geq q+r$, the exceptional set
    \[
        \mathcal{B}_m
        \coloneqq
        \left\{(\theta_1,\ldots,\theta_m)\in\Theta^m:
        \operatorname{span}\{G(\theta_k,X):1\leq k\leq m\}
        \neq\mathbb{R}^r
        \text{ for some }X\in\mathcal{X}\right\}
    \]
    satisfies
    \begin{equation}\label{eq:general-bad-bank-dimension}
        \dim_{\mathrm H}\mathcal{B}_m
        \leq mp-m+q+r-1
        \leq mp-1.
    \end{equation}
    Here $\dim_{\mathrm H}$ denotes Hausdorff dimension.  In particular,
    $\mathcal{B}_m$ has Lebesgue measure zero, so every jointly
    absolutely continuous random bank in
    $\Theta^m$ spans $\mathbb{R}^r$ simultaneously at every
    $X\in\mathcal{X}$ with probability one.
\end{theorem}

\begin{proof}
    For $(X,\lambda)\in\mathcal{X}\times\mathbb{S}^{r-1}$, define
    \[
        h_{X,\lambda}(\theta)
        \coloneqq\langle\lambda,G(\theta,X)\rangle.
    \]
    By \eqref{eq:general-pointwise-spanning}, this analytic function is not
    identically zero on the connected set $\Theta$.  The zero set
    $Z_{X,\lambda}$ of this nontrivial analytic function therefore satisfies
    \[
        \dim_{\mathrm H} Z_{X,\lambda}\leq p-1.
    \]
    Consider the incidence set
    \[
        \mathcal{I}_m
        \coloneqq
        \left\{(\theta_1,\ldots,\theta_m,X,\lambda)
        \in\Theta^m\times\mathcal{X}\times\mathbb{S}^{r-1}:
        \theta_k\in Z_{X,\lambda}\ \text{for every }1\leq k\leq m
        \right\}.
    \]
    Joint analyticity ensures that $\mathcal{I}_m$, like each
    $Z_{X,\lambda}$, is an analytic set and hence a countable union of
    smooth submanifolds; see \citet[Appendix~A]{amir2023neural}.
    For each fixed $(X,\lambda)$, the parameter tuples satisfying
    the incidence conditions form exactly the Cartesian product
    $Z_{X,\lambda}^m$, whose Hausdorff dimension is at most $m(p-1)$
    \citep[Lemma~A.17]{amir2023neural}.
    Since $(X,\lambda)$ ranges over
    $\mathcal{X}\times\mathbb{S}^{r-1}$, of dimension $q+r-1$,
    the dimension bound for countable unions of smooth submanifolds
    \citep[Lemma~A.18]{amir2023neural} gives
    \[
        \dim_{\mathrm H}\mathcal{I}_m
        \leq mp-m+q+r-1.
    \]
    A bank fails to span at some $X$ if and only if a unit covector
    $\lambda$ annihilates all its values there.  Hence $\mathcal{B}_m$ is
    exactly the projection of $\mathcal{I}_m$ onto $\Theta^m$.
    Coordinate projections are Lipschitz and do not increase Hausdorff
    dimension, which proves
    \eqref{eq:general-bad-bank-dimension}.  For $m\geq q+r$, this dimension
    is strictly smaller than $mp$, so $\mathcal{B}_m$ has Lebesgue measure
    zero.  Moreover, $\mathcal{I}_m$ is a countable union of compact
    sets, as is its projection, so $\mathcal{B}_m$ is Borel.
    Any jointly absolutely continuous law assigns probability zero
    to this exceptional set.
\end{proof}

\begin{proof}[Proof of \Cref{lem:attention-uniform-direct-generation}]
    Every rank-one single-head field belongs to the architecture family.
    Indeed, for $a,b,u,v\in\mathbb R^d$, choose a head of width $r$,
    let $e_1\in\mathbb R^r$ be its first coordinate vector, and set
    \begin{equation}\label{eq:attention-rank-one-embedding}
        Q=\sqrt r\,e_1b^\top,\qquad K=e_1a^\top,
        \qquad U=e_1v^\top,\qquad O=ue_1^\top.
    \end{equation}
    Set the output projections of all other heads to zero.
    The effective matrices of the active head are $W=ab^\top$ and
    $V=uv^\top$.  These parameter choices are witnesses inside $\Theta$;
    no such restriction is imposed on the random draw.
    Therefore the rank-one pointwise spanning result
    \eqref{eq:attention-full-rank} implies
    \[
        \operatorname{span}\{F_\theta(X):\theta\in\Theta\}
        =\mathbb R^D
        \qquad\text{for every }X\in\mathcal A_{N,n,d}.
    \]
    The map $(\theta,X)\mapsto F_\theta(X)$ is jointly
    real analytic: the projection factorizations are polynomial, and
    the attention formula involves only finite sums, products,
    exponentials, and division by strictly positive softmax denominators.
    Thus the analytic regularity assumption of
    \Cref{thm:uniform-random-spanning-principle} is satisfied.

    The configuration space $\mathcal A_{N,n,d}$ is nonempty and open
    in $\mathbb R^D$, so it has dimension $D$.
    Apply \Cref{thm:uniform-random-spanning-principle} with
    $\Theta$ as in \eqref{eq:attention-projection-parameter-space},
    $\mathcal X=\mathcal A_{N,n,d}$, $G(\theta,X)=F_\theta(X)$,
    and
    \[
        p=4d\sum_{h=1}^H r_h,\qquad q=D,\qquad r=D,\qquad m=2D.
    \]
    The exceptional banks form a subset of $\Theta^{2D}$ of Hausdorff
    dimension at most $2Dp-1$, hence of Lebesgue measure zero.
    Every jointly absolutely continuous draw, including independent
    Gaussian projection entries with prescribed positive variances,
    avoids this set almost surely and satisfies
    \eqref{eq:attention-uniform-direct-rank}.  The Lie algebra generated
    by the sampled fields contains their real linear span.
\end{proof}

\section{Proof of \texorpdfstring{\Cref{thm:attention-uniform-finite-generation}}{the Gaussian attention universal interpolation theorem}}
\label{app:attention-uniform-generation}

\thmuniformattentiongeneration*

Throughout this section, put $D=Nnd$, $\mathcal A=\mathcal A_{N,n,d}$,
$\mathcal B=\mathcal B_{N,n,d}$, and use the full projection space
$\Theta$ from
\eqref{eq:attention-projection-parameter-space}.
For $x,y\in\mathbb R^d$, we identify their exterior product
$x\wedge y$ with the vector
$(x_i y_j-x_j y_i)_{i<j}\in\mathbb R^{\binom d2}$.
It vanishes exactly when $x$ and $y$ are linearly dependent.
The set $\mathcal B$ is nonempty and open: its defining
conditions are open, and $\mathcal A\subseteq\mathcal B$.
For $d\geq3$, $\mathcal A$ is connected by
\Cref{app:attention-proof} and dense in $\mathcal B$,
so $\mathcal B$ is connected as well.

\begin{lemma}[Contextual separation]\label{lem:attention-contextual-separation}
    Let $d\geq2$, $n\geq2$, and $N\geq1$.  For every
    $X\in\mathcal B$, there exists $(a,b,u,v)\in(\mathbb R^d)^4$ such
    that, setting $W=ab^\top$ and $V=uv^\top$, one has
    \begin{equation}\label{eq:attention-contextual-transversality}
        f_{W,V}(x^\ell)_i\wedge x_j^m
        +x_i^\ell\wedge f_{W,V}(x^m)_j\neq0
    \end{equation}
    for every pair $(\ell,i)\neq(m,j)$ with $x_i^\ell\wedge x_j^m=0$.
    Consequently, one rank-one residual attention layer with sufficiently
    small nonzero scalar gain sends $X$ into $\mathcal A$.
\end{lemma}

\begin{proof}
    Fix $X\in\mathcal B$.  For each sequence, define its normalized
    exponential sum and softmax barycenter by
    \[
        M_\ell(\eta):=\frac1n\sum_{j=1}^n
            e^{\langle\eta,x_j^\ell\rangle},\qquad
        z_\ell(\eta):=\nabla\log M_\ell(\eta)
        =\frac{\sum_j e^{\langle\eta,x_j^\ell\rangle}x_j^\ell}
        {\sum_j e^{\langle\eta,x_j^\ell\rangle}}.
    \]
    The token update is $f_{W,V}(x^\ell)_i=Vz_\ell(Wx_i^\ell)$.

    \emph{Step 1: separate one initially collinear pair.}
    Consider a collinear pair $q=x_i^\ell$ and $x_j^m=cq$.
    Nonzero tokens imply $c\neq0$, and within-sequence noncollinearity
    implies $\ell\neq m$.  We claim that
    \begin{equation}\label{eq:attention-contextual-barycenter-difference}
        \delta(\eta)\coloneqq z_m(c\eta)-c z_\ell(\eta)
        \quad\text{is not identically zero.}
    \end{equation}
    Otherwise, $z_m(c\eta)=c z_\ell(\eta)$ for all $\eta\in\mathbb R^d$.
    Since $z_k=\nabla\log M_k$, the chain rule gives
    \[
        \nabla_\eta\bigl[\log M_m(c\eta)-c^2\log M_\ell(\eta)\bigr]
        =c z_m(c\eta)-c^2 z_\ell(\eta)=0.
    \]
    Thus the difference is constant on $\mathbb R^d$.
    Evaluating at $\eta=0$ and using $M_m(0)=M_\ell(0)=1$ shows that
    this constant is zero, hence
    \[
        \log M_m(c\eta)=c^2\log M_\ell(\eta)
        \qquad\text{for all }\eta\in\mathbb R^d.
    \]
    Choose $h\in\mathbb R^d$ such that
    \[
        \langle h,x_i^k\rangle\neq\langle h,x_j^k\rangle
        \qquad(k\in\{\ell,m\},\ i\neq j).
    \]
    Such an $h$ exists: each forbidden equality defines the proper hyperplane
    $(x_i^k-x_j^k)^\perp$, and only finitely many hyperplanes are excluded.
    Since $c\neq0$, each exponential sum has a unique dominant term along
    $\eta=th$ as $t\to+\infty$, which gives
    \[
        \log M_m(cth)=tA-\log n+o(1),\qquad
        \log M_\ell(th)=tB-\log n+o(1),
    \]
    where $A=\max_j c\langle h,x_j^m\rangle$ and
    $B=\max_i\langle h,x_i^\ell\rangle$.  This also holds when $c<0$.
    Dividing the identity by $t$ first yields $A=c^2B$; comparing the
    remaining constants yields $(c^2-1)\log n=0$.  Since $n\geq2$,
    $c=\pm1$, and the identity reduces to $M_m(c\eta)=M_\ell(\eta)$.
    Let $y_1,\ldots,y_R$ be the distinct vectors in
    $\{c x_j^m\}_j\cup\{x_i^\ell\}_i$, and let $a_r$ be the multiplicity
    of $y_r$ in the first multiset minus its multiplicity in the second.
    Multiplying the identity by $n$ and grouping equal vectors gives
    \[
        \sum_{r=1}^R a_r e^{\langle\eta,y_r\rangle}=0
        \qquad\text{for all }\eta\in\mathbb R^d.
    \]
    Choose $g\in\mathbb R^d$ outside the finitely many proper hyperplanes
    $(y_r-y_s)^\perp$, $r\neq s$, and relabel so that
    $\lambda_1<\cdots<\lambda_R$, where $\lambda_r=\langle g,y_r\rangle$.
    Taking $\eta=tg$ and dividing by $e^{t\lambda_R}$ yields
    \[
        a_R+\sum_{r<R}a_r e^{-t(\lambda_R-\lambda_r)}=0.
    \]
    Letting $t\to+\infty$ gives $a_R=0$; repeating with the remaining
    exponents gives $a_r=0$ for every $r$.  Thus the two multisets agree,
    so there is a permutation $\pi$ such that
    $c x_j^m=x_{\pi(j)}^\ell$ for all $j$.
    Since $c=\pm1$, the two sequences are related by a signed permutation,
    contradicting $X\in\mathcal B$.  This proves the claim.

    Choose $\eta$ with $\delta(\eta)\neq0$ and $u\notin\mathbb Rq$, and
    set
    \[
        W=\frac{\eta q^\top}{\|q\|^2},\qquad
        V=\frac{u\delta(\eta)^\top}{\|\delta(\eta)\|^2}.
    \]
    These matrices have rank at most one, $Wq=\eta$, and
    \[
        f_{W,V}(x^m)_j-c f_{W,V}(x^\ell)_i
        =V\delta(\eta)=u.
    \]
    By bilinearity and antisymmetry of the exterior product, the
    left-hand side of \eqref{eq:attention-contextual-transversality}
    for this pair is
    \begin{align*}
        f_{W,V}(x^\ell)_i\wedge x_j^m
            +x_i^\ell\wedge f_{W,V}(x^m)_j
        &=f_{W,V}(x^\ell)_i\wedge(cq)+q\wedge f_{W,V}(x^m)_j\\
        &=-c\,q\wedge f_{W,V}(x^\ell)_i+q\wedge f_{W,V}(x^m)_j\\
        &=q\wedge\bigl(f_{W,V}(x^m)_j-c f_{W,V}(x^\ell)_i\bigr)\\
        &=q\wedge u\neq0,
    \end{align*}
    where the last inequality follows from $u\notin\mathbb Rq$.

    \emph{Step 2: choose common parameters for all initially collinear pairs.}
    For each initially collinear pair, the squared norm of the left-hand
    side of \eqref{eq:attention-contextual-transversality} is a real
    analytic function of $(a,b,u,v)\in(\mathbb R^d)^4$, with
    $W=ab^\top$ and $V=uv^\top$.  The preceding construction shows that
    it is not identically zero.  Its zero set is therefore closed with
    empty interior, so its complement is open and dense.  There are only
    finitely many initially collinear pairs; hence the intersection of
    these open dense complements is itself open and dense, and in
    particular nonempty.  Any $(a,b,u,v)$ in this intersection satisfies
    all the required inequalities.

    \emph{Step 3: apply one residual layer to reach $\mathcal A$.}
    Fix such a common choice of $a,b,u,v$.  All token updates lie in the image of
    $V=uv^\top$, which has dimension at most one.  Hence, for every pair,
    \begin{align*}
        &(x_i^\ell+t f_{W,V}(x^\ell)_i)
            \wedge(x_j^m+t f_{W,V}(x^m)_j)\\
        &\qquad=x_i^\ell\wedge x_j^m
          +t\bigl(f_{W,V}(x^\ell)_i\wedge x_j^m
                   +x_i^\ell\wedge f_{W,V}(x^m)_j\bigr).
    \end{align*}
    Every initially collinear pair separates for all $t\neq0$.
    Previously noncollinear pairs remain noncollinear, and nonzero tokens
    remain nonzero, for all sufficiently small $|t|$.  There are finitely
    many tokens and pairs, giving one common neighborhood of $t=0$.
    The resulting batch therefore belongs to $\mathcal A$ for every
    sufficiently small nonzero $t$.
\end{proof}

For $\theta\in\Theta$, write $f_\theta=f_{\mathbf W,\mathbf V}$ for the
field on one sequence and $f_\theta^{(N)}$ for its corresponding batch field.

\begin{lemma}[A uniform separating dictionary]\label{lem:attention-uniform-separation}
    For the fixed attention architecture above, let $m\geq D+1$ and
    draw $(\theta_1,\ldots,\theta_m)$ from a jointly absolutely
    continuous law on $\Theta^m$.  With probability one, simultaneously
    for every $X\in\mathcal B$, there is $k\in\{1,\ldots,m\}$ such that
    \[
        f_{\theta_k}(x^\ell)_i\wedge x_j^{\ell'}
        +x_i^\ell\wedge f_{\theta_k}(x^{\ell'})_j\neq0
    \]
    for every pair $(\ell,i)\neq(\ell',j)$ with
    $x_i^\ell\wedge x_j^{\ell'}=0$.
    For this same $k$ and all sufficiently small nonzero $t$, both
    \[
        X+t f_{\theta_k}^{(N)}(X)\in\mathcal A,
        \qquad \Phi_{f_{\theta_k}^{(N)}}^t(X)\in\mathcal A.
    \]
    Both curves remain in $\mathcal B$ for sufficiently small $|t|$.
\end{lemma}

\begin{proof}
    \emph{Step 1: establish uniform separation by the sampled bank.}
    Define the scalar selector
    \begin{equation}\label{eq:attention-contextual-selector}
        \mathcal S(\theta,X)=
        \prod_{(\ell,i)<(\ell',j)}\left(
            \begin{aligned}
                &\|x_i^\ell\wedge x_j^{\ell'}\|^2\\
                &+\|f_\theta(x^\ell)_i\wedge x_j^{\ell'}
                    +x_i^\ell\wedge f_\theta(x^{\ell'})_j\|^2
            \end{aligned}
        \right).
    \end{equation}
    Here $<$ denotes lexicographic order on the token indices, so each
    pair of distinct token positions occurs once.  The function
    $\mathcal S$ is nonnegative and jointly real analytic on
    $\Theta\times\mathcal B$.
    By \Cref{lem:attention-contextual-separation} and the embedding
    \eqref{eq:attention-rank-one-embedding}, for every $X\in\mathcal B$
    there is $\theta\in\Theta$ with $\mathcal S(\theta,X)>0$.
    For a scalar-valued map, this is exactly the spanning assumption of
    \Cref{thm:uniform-random-spanning-principle}.

    Since $\Theta\simeq\mathbb R^p$ and $\mathcal B\subseteq\mathbb R^D$
    is nonempty and open, we can apply that theorem with
    $\mathcal X=\mathcal B$, $G=\mathcal S$, $q=D$, and $r=1$.
    Thus, for $m\geq D+1$, a jointly absolutely continuous draw
    $(\theta_1,\ldots,\theta_m)\in\Theta^m$ almost surely satisfies
    \[
        \forall X\in\mathcal B,\quad
        \exists k\in\{1,\ldots,m\}:\quad
        \mathcal S(\theta_k,X)>0.
    \]
    By definition of $\mathcal S$, this is precisely the required
    transversality for all initially collinear pairs at $X$.

    \emph{Step 2: reach $\mathcal A$ with a residual layer.}
    Fix such a bank and configuration, and choose $k$ with
    $\mathcal S(\theta_k,X)>0$.
    For an initially collinear pair $x_i^\ell,x_j^{\ell'}$, the wedge
    after a residual step is
    \begin{align*}
        &(x_i^\ell+t f_{\theta_k}(x^\ell)_i)
            \wedge(x_j^{\ell'}+t f_{\theta_k}(x^{\ell'})_j)\\
        &\qquad=t\bigl(f_{\theta_k}(x^\ell)_i\wedge x_j^{\ell'}
             +x_i^\ell\wedge f_{\theta_k}(x^{\ell'})_j\bigr)+t^2 f_{\theta_k}(x^\ell)_i
            \wedge f_{\theta_k}(x^{\ell'})_j.
    \end{align*}
    The linear coefficient is nonzero, so the pair separates for all
    sufficiently small nonzero $t$.  Previously noncollinear pairs
    and nonzero tokens persist by continuity.

    \emph{Step 3: reach $\mathcal A$ along the flow.}
    Keep the same $k$ and write $V_{k,h}=O_{k,h}U_{k,h}$.
    For $Z=\operatorname{vec}(z^1,\ldots,z^N)$, each softmax barycenter
    is a convex combination of the tokens in its sequence.  Hence,
    for every $\ell,i$,
    \[
        \|f_{\theta_k}(z^\ell)_i\|
        \leq\left(\sum_{h=1}^H\|V_{k,h}\|_{\mathrm{op}}\right)
            \|z^\ell\|.
    \]
    Summing the squared norms over all tokens and sequences gives
    \[
        \begin{aligned}
            \|f_{\theta_k}^{(N)}(Z)\|^2
            &=\sum_{\ell=1}^N\sum_{i=1}^n
                \|f_{\theta_k}(z^\ell)_i\|^2\leq n\left(\sum_{h=1}^H\|V_{k,h}\|_{\mathrm{op}}\right)^2
                \sum_{\ell=1}^N\|z^\ell\|^2.
        \end{aligned}
    \]
  Taking square roots yields
    \[
        \|f_{\theta_k}^{(N)}(Z)\|
        \leq\sqrt n\left(\sum_{h=1}^H\|V_{k,h}\|_{\mathrm{op}}\right)
              \|Z\|\qquad\text{on }\mathbb R^D.
    \]
    Analyticity gives local existence and uniqueness.  By Gronwall's
    inequality, this linear growth bound rules out finite-time blow-up
    for both the field and its negative, so the flow is defined for all
    $t\in\mathbb R$.

    Write $Z(t)=\Phi_{f_{\theta_k}^{(N)}}^t(X)$ and denote its tokens by
    $z_i^\ell(t)$.  Then $z_i^\ell(0)=x_i^\ell$ and
    $\dot z_i^\ell(0)=f_{\theta_k}(x^\ell)_i$.  For every initially
    collinear pair, bilinearity gives
    \[
        \left.\frac{d}{dt}\bigl(z_i^\ell(t)\wedge z_j^{\ell'}(t)\bigr)
        \right|_{t=0}
        =f_{\theta_k}(x^\ell)_i\wedge x_j^{\ell'}
            +x_i^\ell\wedge f_{\theta_k}(x^{\ell'})_j\neq0.
    \]
    Since $x_i^\ell\wedge x_j^{\ell'}=0$, Taylor's formula yields
    \begin{align*}
        z_i^\ell(t)\wedge z_j^{\ell'}(t)
        &=t\,f_{\theta_k}(x^\ell)_i\wedge x_j^{\ell'}\\
        &\quad+t\,x_i^\ell\wedge f_{\theta_k}(x^{\ell'})_j+O(t^2).
    \end{align*}
    The coefficient of $t$ is nonzero, so this product is nonzero for
    all sufficiently small nonzero $t$.  Initially noncollinear pairs
    and nonzero tokens persist by continuity.  There are finitely many
    tokens and pairs, so one common interval works for all of them.
    Finally, both the residual curve and the flow start at $X\in\mathcal B$;
    by continuity and openness of $\mathcal B$, they remain there for
    sufficiently small $|t|$.
\end{proof}

\begin{proof}[Proof of \Cref{thm:attention-uniform-finite-generation}]
    Let $(\theta_1,\ldots,\theta_{2D})\in\Theta^{2D}$ be the sampled
    parameters, with batch fields $f_{\theta_j}^{(N)}$, $1\leq j\leq2D$.
    Independent Gaussian projection entries have a joint density on
    $\Theta^{2D}$.
    By \Cref{lem:attention-uniform-direct-generation}, with probability
    one these fields span $\mathbb R^D$ simultaneously at every
    $X\in\mathcal A$.
    For separation, recall that the proof of
    \Cref{lem:attention-uniform-separation} applies
    \Cref{thm:uniform-random-spanning-principle} to the scalar selector
    $\mathcal S:\Theta\times\mathcal B\to\mathbb R$ from
    \eqref{eq:attention-contextual-selector}, with $q=D$ and $r=1$.
    Since our bank has $2D\geq q+r=D+1$ fields, it almost surely satisfies
    \[
        \forall X\in\mathcal B,\quad
        \exists k\in\{1,\ldots,2D\}:\quad
        \mathcal S(\theta_k,X)>0.
    \]
    Fix a bank in the intersection of these two probability-one events.
    Both properties then hold uniformly in the configuration for this
    same bank.

    We now prove that the sampled fields have full Lie rank on $\mathcal B$.
    Fix $X\in\mathcal B$ and choose $k\in\{1,\ldots,2D\}$ with
    $\mathcal S(\theta_k,X)>0$, as in
    \Cref{lem:attention-uniform-separation}.
    Write $\Phi_t=\Phi_{f_{\theta_k}^{(N)}}^t$ for its flow and put
    \[
        L_X
        =\operatorname{Lie}\bigl(f_{\theta_1}^{(N)},\ldots,
            f_{\theta_{2D}}^{(N)}\bigr)(X)
        \subseteq\mathbb R^D.
    \]
    For an analytic vector field $G$, its pullback by $\Phi_t$ is the
    vector field defined by
    \[
        (\Phi_t^*G)(Z)
        \coloneqq\bigl(D\Phi_t(Z)\bigr)^{-1}G(\Phi_t(Z)).
    \]
    The Jacobian $D\Phi_t(Z)$ is invertible since $\Phi_{-t}$ is the
    inverse of $\Phi_t$.  For each $j\in\{1,\ldots,2D\}$, the map
    \[
        t\longmapsto U_j(t)
        =\bigl(\Phi_t^*f_{\theta_j}^{(N)}\bigr)(X)
        \in\mathbb R^D
    \]
    is real analytic near $t=0$.  The standard differentiation
    formula for pullbacks
    \citep[Section~2.5, Eq.~(2.21)]{agrachev2004control}
    and its iterates give
    \begin{align*}
        \frac{d}{dt}\Phi_t^*G
        &=\Phi_t^*\bigl[f_{\theta_k}^{(N)},G\bigr],\\
        \frac{d^r}{dt^r}\bigl(\Phi_t^*G\bigr)
        &=\Phi_t^*\bigl(\operatorname{ad}_{f_{\theta_k}^{(N)}}^{\,r}
            G\bigr),\qquad r\geq0,
    \end{align*}
    where $\operatorname{ad}_{f_{\theta_k}^{(N)}}G
    =[f_{\theta_k}^{(N)},G]$.
    Taking $G=f_{\theta_j}^{(N)}$ and evaluating at $t=0$, where
    $\Phi_0^*$ is the identity, gives
    \[
        U_j^{(r)}(0)
        =\bigl(\operatorname{ad}_{f_{\theta_k}^{(N)}}^{\,r}
            f_{\theta_j}^{(N)}\bigr)(X)
        \in L_X,\qquad r\geq0.
    \]
    Every partial sum of the Taylor series of $U_j$ belongs to $L_X$.
    Since this series converges to $U_j(t)$ for sufficiently small
    $|t|$ and $L_X$ is closed, we obtain $U_j(t)\in L_X$ for every $j$
    and all sufficiently small $|t|$.

    Choose such a nonzero $t$ with $X'=\Phi_t(X)\in
    \mathcal A$, as guaranteed by
    \Cref{lem:attention-uniform-separation}.  Uniform direct spanning
    gives
    \[
        \operatorname{span}\bigl\{
            f_{\theta_j}^{(N)}(X'):1\leq j\leq2D
        \bigr\}=\mathbb R^D.
    \]
    Since $D\Phi_t(X)$ is invertible, their pullbacks $U_j(t)$
    also span $\mathbb R^D$.  Since they belong to $L_X$,
    \[
        \mathbb R^D
        =\operatorname{span}\{U_j(t):1\leq j\leq2D\}
        \subseteq L_X\subseteq\mathbb R^D.
    \]
    Thus $L_X=\mathbb R^D$.  This holds at every
    $X\in\mathcal B$ for the same bank of full multi-head blocks.
    The same proof applies to independent Gaussian entries with any
    prescribed positive variances, or more generally to a jointly
    absolutely continuous law on $\Theta^{2D}$.

    The sampled fields are analytic, and $\mathcal B$ is nonempty,
    open, and connected for $d\geq3$ by the domain argument above.
    Moreover, $-f_{\theta_j}$ belongs to the same architecture family:
    it is realized by negating all output-projection blocks of
    $\theta_j$.
    The draw is fixed before the endpoints, so
    \Cref{cor:universal-interpolation-criterion} proves universal
    interpolation on $\mathcal{B}_{N,n,d}$ with the same $2D$
    sampled generators, used in either sign over a horizon $[0,S]$
    that may depend on the endpoints.  Since $N$ was arbitrary, this
    holds for every $N$.
\end{proof}

\begin{remark}[Rank-one attention]\label{rem:attention-rank-one}
    \Cref{thm:attention-uniform-finite-generation} includes the case
    of one head of width one.  Taking $H=1$ and $r_1=1$, write the
    Gaussian projections as $Q_k=b_k^\top$, $K_k=a_k^\top$,
    $U_k=v_k^\top$, and $O_k=u_k$.  Then
    \[
        W_k=a_kb_k^\top,\qquad V_k=u_kv_k^\top,
        \qquad 1\leq k\leq2D,
    \]
    where all vectors $a_k,b_k,u_k,v_k$ are independent standard
    Gaussian vectors in $\mathbb R^d$.  Both effective matrices have
    rank one almost surely.  Thus the same theorem gives universal
    interpolation at continuous depth on $\mathcal B_{N,n,d}$ with
    a frozen bank of $2Nnd$ such fields.
\end{remark}

\paragraph{Dependence of the terminal depth on the endpoints.}
Allowing the terminal time to vary is essential for this formulation.
Each sampled field, or its negative, satisfies
$\lVert G(Z)\rVert\leq C\lVert Z\rVert$ with a common
$C<\infty$, since softmax takes convex combinations of token values.
Gronwall's inequality therefore bounds every trajectory using these
unscaled fields by
\[
    \lVert Z(S)\rVert\leq e^{CS}\lVert Z(0)\rVert.
\]
Yet $\lambda X\in\mathcal{B}_{N,n,d}$ for every
$X\in\mathcal{B}_{N,n,d}$ and $\lambda>0$.  For any fixed
horizon $S_0$, choosing $\lambda>e^{CS_0}$ gives an admissible target
$Y=\lambda X$ that cannot be reached by time $S_0$.
A horizon depending on the endpoints removes this obstruction
without introducing adjustable scalar gains.

\section{Proof of \texorpdfstring{\Cref{cor:attention-dictionary-finite-depth}}{the finite-depth universal interpolation corollary}}
\label{app:attention-dictionary-finite-depth}

\begin{proof}
    Fix $N\geq1$, put $D=Nnd$, and choose a draw
    $(\theta_1,\ldots,\theta_{2D})\in\Theta^{2D}$ in the intersection of
    the probability-one events in
    \Cref{lem:attention-uniform-direct-generation,lem:attention-uniform-separation}.
    These events hold simultaneously for all
    configurations in their respective domains, so the bank is fixed
    before the endpoints are chosen.  The batch fields
    $f_{\theta_1}^{(N)},\ldots,f_{\theta_{2D}}^{(N)}$ are analytic
    and span $\mathbb{R}^D$ at every point of
    $\mathcal{A}_{N,n,d}$ by
    \Cref{lem:attention-uniform-direct-generation}.  This set is open
    and connected for $d\geq3$, and the attention architecture allows
    real scalar multiplication since
    $c f_{\mathbf W,\mathbf V}=f_{\mathbf W,c\mathbf V}$.  Hence
    \Cref{prop:universal-finite-depth-interpolation} gives universal
    interpolation at finite depth on $\mathcal{A}_{N,n,d}$ using
    this same bank, with all intermediate
    configurations in $\mathcal{A}_{N,n,d}$.

    Now fix arbitrary endpoints
    $X,Y\in\mathcal{B}_{N,n,d}$.  By
    \Cref{lem:attention-uniform-separation}, choose an index
    $k_X\in\{1,\ldots,2D\}$ and set $F\coloneqq f_{\theta_{k_X}}^{(N)}$.
    For sufficiently small $s>0$,
    \[
        X^+\coloneqq X+sF(X)
        \in\mathcal{A}_{N,n,d}.
    \]
    This is one admissible residual layer.

    At the target, choose an index $k_Y\in\{1,\ldots,2D\}$ supplied
    by the same lemma and set $G\coloneqq f_{\theta_{k_Y}}^{(N)}$.
    For every pair $(\ell,i)\neq(\ell',j)$ with
    $y_i^\ell\wedge y_j^{\ell'}=0$, this choice gives
    \[
        f_{\theta_{k_Y}}(y^\ell)_i\wedge y_j^{\ell'}
        +y_i^\ell\wedge f_{\theta_{k_Y}}(y^{\ell'})_j\neq0.
    \]
    Consider the equation
    \[
        Z+tG(Z)-Y=0.
    \]
    Its derivative with respect to $Z$ at
    $(t,Z)=(0,Y)$ is the identity.  The implicit
    function theorem therefore gives an analytic curve
    $Z(t)$ for $t$ near zero satisfying
    \begin{gather*}
        Z(0)=Y,\qquad
        Z(t)+tG(Z(t))=Y,\\
        Z'(0)=-G(Y).
    \end{gather*}
    Taylor's formula therefore gives
    \[
        Z(t)=Z(0)+tZ'(0)+O(t^2)=Y-tG(Y)+O(t^2).
    \]
    Write $z_i^\ell(t)$ for the tokens of $Z(t)$.  For each initially
    collinear pair, its wedge along this curve is
    \begin{align*}
        z_i^\ell(t)\wedge z_j^{\ell'}(t)
        &=-t\,f_{\theta_{k_Y}}(y^\ell)_i\wedge y_j^{\ell'}\\
        &\quad-t\,y_i^\ell\wedge f_{\theta_{k_Y}}(y^{\ell'})_j+O(t^2).
    \end{align*}
    It is consequently nonzero for all sufficiently small nonzero
    $t$.  Initially noncollinear pairs remain noncollinear by
    continuity, and all tokens remain nonzero.  There are finitely
    many tokens and pairs, so we may choose one sufficiently small
    $t>0$ such that
    \[
        Y^-\coloneqq Z(t)
        \in\mathcal{A}_{N,n,d},\qquad
        Y^-+tG(Y^-)=Y.
    \]

    Apply the finite-depth interpolation result on
    $\mathcal{A}_{N,n,d}$ to connect $X^+$ to
    $Y^-$.  Prepend the layer
    $\operatorname{Id}+sF$ and append the layer
    $\operatorname{Id}+tG$.  This gives the required finite residual
    network.  All configurations
    other than the prescribed endpoints belong to
    $\mathcal{A}_{N,n,d}$, and hence every configuration belongs to
    $\mathcal{B}_{N,n,d}$.  The construction uses at most two additional
    layers around an interpolation inside $\mathcal{A}_{N,n,d}$.
    The implicit function theorem only defines the intermediate
    target $Y^-$; the last network layer is the admissible
    forward residual map $\operatorname{Id}+tG$.

    Every selected index and scalar gain is shared across the entire
    batch.  Finally, each multiple $c f_{\theta_j}$ is realized by
    a scalar gain on the frozen block output; the head
    widths and all stored projections remain unchanged.
    The same bank works for every endpoint pair.
    If $X=Y$, the depth-zero network also
    suffices.  Since the sampled fields are fixed before the endpoints,
    this is universal interpolation at finite depth on
    $\mathcal{B}_{N,n,d}$ in the sense of \Cref{sec:finite-depth}.
    The argument applies to every $N\geq1$.
\end{proof}

\section{Proof of \texorpdfstring{\Cref{thm:two-gaussian-connected-domain}}{the two-Gaussian-block interpolation theorem}}
\label{app:two-gaussian-attention}

This appendix proves \Cref{thm:two-gaussian-connected-domain} for
noncausal attention with unrestricted effective score and value matrices.
The admissible
interpolation domain here is allowed to depend on the sampled pair,
whereas the Gaussian multihead bank in
\Cref{thm:attention-uniform-finite-generation} works on the fixed domain
$\mathcal B_{N,n,d}$.
The same pair works for every finite batch size and sequence length.
For a single sequence $x=\operatorname{vec}(x_1,\ldots,x_n)$, we use
the single-head field
\begin{equation}\label{eq:two-gaussian-field}
    f_{W,V}(x)_i
    =V\frac{\sum_{j=1}^n x_j\exp(x_j^\top W x_i)}
             {\sum_{j=1}^n\exp(x_j^\top W x_i)},
    \qquad W,V\in\mathbb R^{d\times d},
\end{equation}
and its batch lift $f^{(N)}_{W,V}$.  One head of width at least $d$
realizes every such field; in a multihead block, the remaining heads
can be set to zero.  Under the standard equal-width convention
$r_h=d/H$, this single-active-head realization corresponds to $H=1$.

Write $\theta=(W_1,V_1,W_2,V_2)\in\mathbb R^{4d^2}$ and, for
$N\geq1$ and $n\geq2$, put $D=Nnd$ and
$f_r^{(N)}=f_{W_r,V_r}^{(N)}$.  We choose the interpolation domain as
\begin{align}
    \mathcal R_{\theta,N,n}
    &\coloneqq\left\{X\in\mathcal B_{N,n,d}:
    \operatorname{Lie}(f_1^{(N)},f_2^{(N)})(X)
    =\mathbb R^D\right\},
    \label{eq:two-gaussian-regular-domain}\\
    \Sigma_{\theta,N,n}
    &\coloneqq\mathcal B_{N,n,d}\setminus\mathcal R_{\theta,N,n}.
    \nonumber
\end{align}
Thus $\mathcal R_{N,n}=\mathcal R_{\theta,N,n}$ in the main statement.
We prove the stronger bound
\begin{equation}\label{eq:two-gaussian-codimension}
    \dim_{\mathbb R}\Sigma_{\theta,N,n}\leq Nnd-2,
\end{equation}
almost surely, simultaneously for every $N\geq1$ and $n\geq2$.
In particular, the domain also has full Lebesgue measure in
$\mathcal B_{N,n,d}$ and is path connected.
More generally, the same conclusions hold when $\theta$ is drawn from
\textcolor{black}{any joint law absolutely continuous with respect to
Lebesgue measure on $\mathbb R^{4d^2}$}.  Both value
matrices $V_1,V_2$ are then invertible almost surely.

\textcolor{black}{The Gaussian projection initialization in
\Cref{thm:two-gaussian-connected-domain} induces such a law.
Indeed, condition on the key and value projections, which almost surely
have full column rank when $r\ge d$.  The maps from the query and output
projections to the effective matrices are then surjective linear maps.
Since all projection entries are independent standard Gaussians, the
conditional joint law of $\theta$ is a nondegenerate Gaussian, and its
unconditional joint law is therefore absolutely continuous.
The arguments below only exclude Lebesgue-null parameter sets.}

The two frozen base fields, used with either sign on a freely chosen
interval $[0,S]$, give universal interpolation in the sense of
\Cref{def:universal-interpolation} on $\mathcal R_{\theta,N,n}$,
with $m_N=2$.  Their negatives are fixed by the same draw and require
no additional sampled matrices.  The matrices are independent
of $N$ and $n$, while the domain may depend on $\theta,N,n$.
The theorem does not assert full Lie rank on all of
$\mathcal B_{N,n,d}$, nor a width-one realization.

\subsection{A finite Lie-minor witness for every batch size}
\label{app:two-gaussian-rank-witness}

\begin{lemma}[A two-field witness for arbitrary batch size]
\label{lem:two-gaussian-witness}
For every $d\geq3$, $n\geq2$, and $N\geq1$, there exist invertible
$W_1,V_1,W_2,V_2\in\mathbb R^{d\times d}$ and
$X\in\mathcal A_{N,n,d}$ such that, with $D=Nnd$,
\[
 \operatorname{Lie}\bigl(f_{W_1,V_1}^{(N)},
 f_{W_2,V_2}^{(N)}\bigr)(X)=\mathbb R^D.
\]
\end{lemma}

\begin{proof}
Put $M=Nn$, $\eta=(4M)^{-1}$, and choose
\begin{equation}\label{eq:two-gaussian-witness-tokens}
 s_{\ell i}=\frac14+((\ell-1)n+i)\eta,
 \qquad
 x_i^\ell=s_{\ell i}e_1+
          \sqrt{1-s_{\ell i}^2}\,e_2.
\end{equation}
All $M$ tokens are distinct points in the open first quadrant of the unit
circle, so this batch belongs to $\mathcal A_{N,n,d}$.
For each token define the real-analytic function
\[
 h_{\ell i}(\beta)=
 \frac{\sum_{j=1}^n s_{\ell j}
                  e^{\beta s_{\ell i}s_{\ell j}}}
      {\sum_{j=1}^n e^{\beta s_{\ell i}s_{\ell j}}}.
\]
The largest and second-largest scalars in sequence $\ell$ differ by
$\eta$ (indeed $s_{\ell n}-s_{\ell,n-1}
=\bigl[n-(n-1)\bigr]\eta=\eta$).  Dividing the numerator and denominator by \(e^{\beta s_{\ell i}s_{\ell n}}\) gives
\begin{equation}\label{eq:two-gaussian-scalar-asymptotics}
 h_{\ell i}(\beta)
 =s_{\ell n}-\eta e^{-\eta s_{\ell i}\beta}
       +o\bigl(e^{-\eta s_{\ell i}\beta}\bigr),
 \qquad \beta\longrightarrow+\infty.
\end{equation}
The
functions $1,h_{\ell i}$ are linearly independent: indeed suppose that
\[
a_0+\sum_{\ell,i}a_{\ell i}h_{\ell i}(\beta)=0.
\]
Subtracting its limit as \(\beta\to+\infty\) gives
\[
\sum_{\ell,i}a_{\ell i}
\bigl(h_{\ell i}(\beta)-s_{\ell n}\bigr)=0.
\]By the asymptotic expansion,
\[
h_{\ell i}(\beta)-s_{\ell n}
=(-\eta+o(1))e^{-\eta s_{\ell i}\beta}.
\]If any coefficient \(a_{\ell i}\) is nonzero, choose the unique index (as the rates $\eta s_{\ell i}$ are pairwise distinct) \((\ell_*,i_*)\) with the smallest decay rate among those with nonzero coefficients. Dividing by \(e^{-\eta s_{\ell_*i_*}\beta}\) and letting \(\beta\to+\infty\), all other terms vanish, yielding
\[
-\eta a_{\ell_*i_*}=0,
\]a contradiction. Thus all \(a_{\ell i}\) vanish, and the original relation then gives \(a_0=0\), proving linear independence.

Consider the auxiliary pair consisting of the tokenwise linear field
$L_\Lambda$ and an attention field $H$, where
\[
 \Lambda=\operatorname{diag}(\lambda_1,\ldots,\lambda_d),
 \qquad \lambda_r=1+\frac{r-1}{d},\qquad
 v=(1,\ldots,1)^\top,\qquad
 H=f_{e_1e_1^\top,\,ve_1^\top}^{(N)}.
\]
Here $(L_\Lambda(X))_i^\ell=\Lambda x_i^\ell$.
The flow $\Phi_t$ of $L_\Lambda$ multiplies every token by
$e^{t\Lambda}$.  Its pushforward of $H$, evaluated at the batch
\eqref{eq:two-gaussian-witness-tokens}, has spatial component $r$
\begin{equation}\label{eq:two-gaussian-conjugated-field}
 \bigl((\Phi_t)_*H\bigr)_{\ell i,r}(X)
   =e^{(\lambda_r-1)t}h_{\ell i}(e^{-2t}).
\end{equation}
We claim that the evaluations of
$\operatorname{ad}_{L_\Lambda}^kH$, $k\geq0$, span $\mathbb R^D$.
If a covector $(c_{\ell i,r})$ annihilated them all, the real-analytic
function obtained by pairing it with \eqref{eq:two-gaussian-conjugated-field}
would have every derivative zero at $t=0$, and hence vanish for all real
$t$.  Indeed, these derivatives are the iterated brackets, up to signs.
Set
\[
 B_r(\beta)=\sum_{\ell,i}c_{\ell i,r}h_{\ell i}(\beta),
 \qquad z=e^{-t/d}.
\]
The resulting identity is
\begin{equation}\label{eq:two-gaussian-laurent-separation}
 \sum_{r=1}^d z^{-(r-1)}B_r(z^{2d})=0,
 \qquad z>0.
\end{equation}
Each $B_r$ is analytic near zero.  After multiplication by $z^{d-1}$,
its contribution to \eqref{eq:two-gaussian-laurent-separation} contains
only powers $d-r+2dk$, $k\geq0$.  These sets of exponents are disjoint
for different $r$.  Every Taylor coefficient of every $B_r$ therefore
vanishes.  Analyticity and the independence of the $h_{\ell i}$ imply
$c_{\ell i,r}=0$ for all indices, proving the claim.

Finite dimensionality now provides nonnegative integers
$k_1,\ldots,k_D$ for which
\begin{equation}\label{eq:two-gaussian-auxiliary-determinant}
 \det\bigl(
   (\operatorname{ad}_{L_\Lambda}^{k_j}H)(X)
       \bigr)_{j=1}^D\ne0.
\end{equation}
To replace the linear field by attention, put
$G_T=f_{TI,\Lambda}^{(N)}$.  At the fixed batch all tokens have norm one
and, within each sequence, $\langle x_i^\ell,x_j^\ell\rangle<1$ when
$i\ne j$.  Thus, for every fixed derivative order $a$,
\begin{equation}\label{eq:two-gaussian-finite-jet-limit}
 D^aG_T(X)\longrightarrow D^aL_\Lambda(X)
 \qquad (T\longrightarrow+\infty).
\end{equation}
For completeness, divide each softmax row by its diagonal exponential.
Every remaining exponential is $e^{-Tg_{ij}^\ell}$ at the witness, with
$g_{ij}^\ell=1-\langle x_i^\ell,x_j^\ell\rangle>0$.
Every fixed-order derivative is bounded by a polynomial in $T$ times
$e^{-Tg}$, where $g$ is the smallest of these finitely many gaps.
The denominators after division are bounded below by one.  This proves
\eqref{eq:two-gaussian-finite-jet-limit}, including derivatives transverse
to the plane containing the witness tokens.

A fixed Lie word evaluated at a point depends polynomially on finitely
many derivatives of its generators.  Hence the determinant in
\eqref{eq:two-gaussian-auxiliary-determinant} remains nonzero after replacing
$L_\Lambda$ by $G_T$, for some finite sufficiently large $T$.
Finally replace the second field's parameters by
\[
 W_2=e_1e_1^\top+\varepsilon I,
 \qquad V_2=ve_1^\top+\varepsilon I.
\]
Both determinants equal $\varepsilon^{d-1}(1+\varepsilon)$, and the
selected Lie determinant remains nonzero for sufficiently small
$\varepsilon>0$ by continuity.  Together with $W_1=TI$ and
$V_1=\Lambda$, this gives four invertible matrices and proves the lemma.
\end{proof}

\begin{lemma}[Generic rank simultaneously for all finite batches]
\label{lem:two-gaussian-generic-rank}
Fix $d\geq3$, and draw $(W_1,V_1,W_2,V_2)$ from \textcolor{black}{a joint law
absolutely continuous with respect to Lebesgue measure on
$\mathbb R^{4d^2}$}.  Almost surely, simultaneously for every
$n\geq2$ and $N\geq1$, the full Lie-rank locus in
$\mathcal B_{N,n,d}$ is open, dense, and has full Lebesgue measure.
\end{lemma}

\begin{proof}
For fixed $n,N$, \Cref{lem:two-gaussian-witness} provides a fixed batch
$X_0$ and finitely many Lie words whose determinant
$\Delta_{n,N}(\theta,X)$ is nonzero at one parameter choice,
where $\theta=(W_1,V_1,W_2,V_2)$.
The determinant is jointly real analytic in all parameters and state
coordinates: the softmax denominators are strictly positive on the
entire real domain.  Therefore
$\theta\mapsto\Delta_{n,N}(\theta,X_0)$ is a nonzero analytic
function, whose zero set has Lebesgue measure zero.
For almost every draw, $\Delta_{n,N}(\theta,\cdot)$ is consequently
nonzero analytic.  Its nonvanishing locus is open, dense, and of full
measure, also after restriction to $\mathcal B_{N,n,d}$.
The full Lie-rank locus contains it and is itself open, so it has the
same three properties.  Finally intersect these probability-one parameter
events over the countably many pairs $(n,N)$.  The resulting event still
has probability one and uses the same two generators for every pair.
\end{proof}

\subsection{Generic matrix conditions}
\label{app:two-gaussian-matrix-conditions}

Write
\[
    S_r=\frac{W_r+W_r^\top}{2},
    \qquad K_r=\frac{W_r-W_r^\top}{2},
    \qquad r=1,2.
\]
In addition to the event supplied by
\Cref{lem:two-gaussian-generic-rank}, we impose the following conditions:
\begin{enumerate}
    \item $\operatorname{Lie}_{\mathrm{mat}}(V_1,V_2)
    =\mathrm{gl}_d(\mathbb R)$;
    \item $S_1,S_2$ are invertible and nonproportional, and $K_1,K_2$
    are nonzero and nonproportional;
    \item for $r\neq s$,
    \begin{equation}\label{eq:two-gaussian-nonconformal}
        V_s^\top S_r+S_rV_s\notin\mathbb R S_r,
        \qquad
        V_s^\top K_r+K_rV_s\notin\mathbb R K_r;
    \end{equation}
    \item each $V_r$ has simple complex spectrum and no two of its
    eigenvalues, allowing repetition, sum to zero.
\end{enumerate}
These conditions hold almost surely and do not depend on $N,n$.
For the first, the matrices
\[
    D_0=\operatorname{diag}(1,2,4,\ldots,2^{d-1}),
    \qquad R_0=I+\mathbf1\mathbf1^\top
\]
generate $\mathrm{gl}_d$: their distinct nonzero adjoint eigenvalue
gaps isolate every off-diagonal matrix unit from $R_0$; brackets give
the traceless diagonals, and $D_0$ has nonzero trace.  A finite matrix
Lie determinant is therefore a nonzero polynomial in the two values.
The second and fourth conditions also exclude proper algebraic sets.
Here the skew-symmetric matrix space has dimension at least two
because $d\geq3$.

For fixed invertible symmetric $S$, the map
$V\mapsto V^\top S+SV$ is onto the symmetric matrices, so the inverse
image of $\mathbb R S$ is a proper linear subspace.  For fixed nonzero
skew-symmetric $K$, put $K$ in real block-diagonal canonical form with
$K_{12}\neq0$.  The choice $V=E_{13}$ gives
$V^\top K+KV$ a nonzero $(2,3)$ entry, whereas $K_{23}=0$.
Thus the skew condition also excludes a proper linear subspace.
Fubini's theorem proves \eqref{eq:two-gaussian-nonconformal} almost
surely.  Independence of the parameter blocks is unnecessary: the
joint law has a density.

Fix a parameter pair satisfying these conditions.  Enumerate the
$p=2Nn^2$ score quadratics as
\begin{equation}\label{eq:two-gaussian-score-quadratics}
    Q_{r,a,i,j}(X)=(x_j^a)^\top W_r x_i^a.
\end{equation}
They are linearly independent over $\mathbb C$.  Indeed, terms from
different sequences or different unordered token pairs have distinct
multidegrees.  For a diagonal token pair, independence follows from
that of $S_1,S_2$.  For an off-diagonal pair, a relation among
$W_1,W_1^\top,W_2,W_2^\top$ separates into a relation among $S_1,S_2$
and one among $K_1,K_2$, so all coefficients vanish.

Let $L_{V_r}$ denote the shared tokenwise linear field
$(L_{V_r}X)_i^a=V_rx_i^a$.  The fourth condition implies that
$L_{V_r}$, acting by differentiation, is invertible on the space of
homogeneous quadratic polynomials in all $D$ state coordinates.
In an eigenbasis, its eigenvalues on that space are precisely sums of
two spatial eigenvalues, with repetitions from the token indices.

\subsection{Exclusion of invariant algebraic hypersurfaces}
\label{app:two-gaussian-algebraic-barriers}

We first record the independence property needed for polynomial
exponents on a regular real hypersurface patch.
\begin{lemma}[Exponential independence on a hypersurface patch]
\label{lem:two-gaussian-exponential-independence}
    Let $h\in\mathbb R[x_1,\ldots,x_q]$, with $q\geq2$, and let
    $\mathcal M$ be a nonempty connected relatively open subset of
    \[
        \{x\in\mathbb R^q:h(x)=0,\ \nabla h(x)\neq0\}.
    \]
    Let $P_1,\ldots,P_s$ and $A_1,\ldots,A_s$ be real polynomials
    in these coordinates.  Suppose that $P_i-P_j$ is nonconstant
    on $\mathcal M$ whenever $i\neq j$.  If
    \[
        \sum_{j=1}^s A_j(x)e^{P_j(x)}=0
        \qquad\text{for every }x\in\mathcal M,
    \]
    then $A_j(x)=0$ for every $x\in\mathcal M$ and every $j$.
    The conclusion concerns the restrictions of the coefficients to
    $\mathcal M$, not their values outside this patch.
\end{lemma}
\begin{proof}
    We first explain a fact about functions of one real variable,
    and then use the implicit function theorem to restrict the
    proposed identity to such functions.

    \smallskip
    \noindent\emph{Algebraic functions of one variable.}
    A real analytic function $b(t)$ on an interval is called
    algebraic if it satisfies
    \[
        Q(t,b(t))=0
    \]
    for a nonzero polynomial $Q$ in two variables.  For example,
    $b(t)=\sqrt{1+t^2}$ satisfies $b(t)^2-1-t^2=0$.
    We use the elementary elimination properties of polynomial
    equations: sums, products, and quotients of algebraic functions
    are algebraic wherever defined; moreover, any two algebraic
    functions $b(t),c(t)$ satisfy a nonzero polynomial relation
    $R(b(t),c(t))=0$.  In the latter property the parameter $t$ is
    eliminated from their polynomial equations.
    Derivatives are algebraic as well.  Indeed, choose an equation
    $Q(t,b(t))=0$ of smallest possible degree in its second variable.
    Its partial derivative $Q_z(t,b(t))$ cannot vanish identically,
    since that would give an equation of smaller degree.  Implicit
    differentiation therefore gives, after restricting to a smaller
    interval if needed,
    \[
        b'(t)=-\frac{Q_t(t,b(t))}{Q_z(t,b(t))}.
    \]

    \smallskip
    \noindent\emph{An exponential of a nonconstant algebraic function
    is not algebraic.}
    Suppose that both $b$ and $e^b$ were algebraic, with $b$
    nonconstant.  The preceding elimination property would give
    $R(b(t),e^{b(t)})=0$ for some nonzero polynomial $R$.
    The range of $b$ contains a nonempty open interval, so
    $R(u,e^u)=0$ on that interval.  Since this expression is real
    analytic on all of $\mathbb R$, it vanishes for every real $u$.
    Writing $R(u,v)=\sum_{k=0}^m p_k(u)v^k$ with $p_m\neq0$,
    and dividing by $e^{mu}$, gives
    \[
        p_m(u)+\sum_{k<m}p_k(u)e^{-(m-k)u}=0.
    \]
    As $u\to+\infty$, every term in the sum tends to zero.
    This would force the nonzero polynomial $p_m(u)$ to tend to
    zero, which is impossible.

    \smallskip
    \noindent\emph{Independence in one variable.}
    Let $f_1(t),\ldots,f_s(t)$ be algebraic real analytic functions
    on an open interval $I$, with pairwise nonconstant differences.
   Suppose that their
    exponentials satisfy a nontrivial relation with algebraic analytic
    coefficients. A coefficient
    is counted as nonzero if it is not identically zero as a
    function, even though it may vanish at some individual points.

    Keep the functions $f_1,\ldots,f_s$ fixed, but consider all
    possible choices of algebraic real analytic coefficients giving
    a nontrivial relation, on $I$ or on a nonempty open subinterval.
    Among these relations, choose one with the smallest number $m$
    of nonzero coefficients.  Such a minimum exists because this
    number is an integer between $1$ and $s$.
    Discard the identically zero terms and relabel the remaining
    functions and coefficients, so that the chosen relation reads
    \[
        a_1(t)e^{f_1(t)}+\cdots+a_m(t)e^{f_m(t)}=0,
    \]
    with every $a_j$ not identically zero.  Necessarily $m\geq2$:
    a relation with only one term would force its coefficient to
    vanish identically, since an exponential never vanishes.

    Since $a_m$ is not identically zero, choose a point $t_0$ where
    $a_m(t_0)\neq0$.  By continuity, there is a smaller open interval
    $J$ containing $t_0$ on which $a_m(t)$ never vanishes.
    The product $a_m(t)e^{f_m(t)}$ is therefore nonzero throughout
    $J$, and division gives
    \[
        1+\sum_{j=1}^{m-1}
        \frac{a_j(t)}{a_m(t)}e^{f_j(t)-f_m(t)}=0
        \qquad(t\in J).
    \]
    Define
    \[
        r_j(t)=\frac{a_j(t)}{a_m(t)},
        \qquad b_j(t)=f_j(t)-f_m(t),\qquad j<m.
    \]
    Each $r_j$ is algebraic and real analytic on $J$, and remains
    not identically zero.  Each $b_j$ is algebraic and nonconstant;
    analyticity ensures that a nonconstant difference cannot become
    constant on a smaller open interval.  We thus obtain
    \[
        1+\sum_{j<m}r_j(t)e^{b_j(t)}=0.
    \]
    The purpose of this division is to turn the last term into the
    constant $1$, which disappears when we differentiate:
    \[
        \sum_{j<m}\bigl(r_j'+r_jb_j'\bigr)e^{b_j}=0.
    \]
    All the new coefficients are algebraic and real analytic.
    If any were nonzero,
    multiplying by $e^{f_m}$ would give a relation among the original
    exponentials with fewer than $m$ terms.  Minimality thus gives
    $r_j'+r_jb_j'=0$ for every $j$.  Consequently,
    \[
        (r_je^{b_j})'=0,
        \qquad r_j=c_je^{-b_j},\quad c_j\neq0.
    \]
    This contradicts the preceding paragraph: $-b_j$ is a
    nonconstant algebraic function, whereas $e^{-b_j}=r_j/c_j$
    would be algebraic.

    \smallskip
    \noindent\emph{Restriction of the hypersurface identity to one
    variable.}
    Suppose, toward a contradiction, that there exists $p\in\mathcal M$
    such that $A_\ell(p)\neq0$.  Since $\nabla h(p)\neq0$, we may
    permute coordinates so that $\partial_{x_q}h(p)\neq0$.
    The real analytic implicit function theorem then solves
    $h(x)=0$ locally for the last coordinate as a real analytic
    function $x_q=\varphi(u)$ of $u=(x_1,\ldots,x_{q-1})$.
    Thus a neighborhood of $p$ in $\mathcal M$ is parametrized by
    \[
        x=\Phi(u)=(u,\varphi(u)),\qquad
        h(u,\varphi(u))=0,
        \qquad u\in U\subseteq\mathbb R^{q-1},
    \]
    where $U$ is open. In particular, the coefficient on this graph is
    $u\mapsto A_\ell(\Phi(u))$.
    By shrinking $U$, continuity ensures that both
    $\partial_{x_q}h(\Phi(u))$ and $A_\ell(\Phi(u))$ remain nonzero
    throughout $U$.
    Each difference $(P_i-P_j)\circ\Phi$ is nonconstant on every
    open subset of $U$: otherwise analytic continuation on the
    connected patch $\mathcal M$ would make $P_i-P_j$ constant
    throughout $\mathcal M$.  Its gradient is therefore nonzero on
    an open dense subset of $U$.
    There are only finitely many differences, so we can choose
    $u_0\in U$ where all these gradients are nonzero and
    $A_\ell(\Phi(u_0))\neq0$.  Choose a direction $v$ such that
    \[
        \nabla\bigl((P_i-P_j)\circ\Phi\bigr)(u_0)\cdot v\neq0
        \qquad(i\neq j).
    \]
    The forbidden directions lie in finitely many proper linear
    hyperplanes, which cannot cover $\mathbb R^{q-1}$.

    For sufficiently small $t$, set $x(t)=\Phi(u_0+tv)$.
    Its first $q-1$ coordinates are affine functions of $t$.
    Its last coordinate satisfies the polynomial equation
    \[
        h(u_0+tv,x_q(t))=0.
    \]
    This equation is nonzero as a polynomial in $(t,x_q)$, because
    $\partial_{x_q}h$ is nonzero at the starting point.  Thus all
    coordinates of $x(t)$ are algebraic functions of $t$, and so
    are $f_j(t)=P_j(x(t))$ and $a_j(t)=A_j(x(t))$.
    Our choice of $v$ ensures $(f_i-f_j)'(0)\neq0$, while
    $a_\ell(0)\neq0$.  Restricting the assumed identity gives
    \[
        \sum_j a_j(t)e^{f_j(t)}=0,
    \]
    contradicting the one-variable independence just proved.
\end{proof}

\begin{lemma}[No algebraic hypersurface barrier]
\label{lem:two-gaussian-no-algebraic-barrier}
    Under conditions~1, 2, and~3 of
    \Cref{app:two-gaussian-matrix-conditions}, put $D=Nnd$ and
    write $X=(x_i^a)_{a,i}\in\mathbb R^D$ for the batch.
    There do not exist a nonconstant real polynomial $h$ in these
    $D$ scalar coordinates and a nonempty connected relatively open
    subset $\mathcal M$ of
    \[
        \{X\in\mathcal B_{N,n,d}:h(X)=0,\ \nabla h(X)\neq0\}
    \]
    such that both batch attention fields are tangent to $\mathcal M$
    at every point.
\end{lemma}
\begin{proof}
    Suppose that such a polynomial $h$ and set $\mathcal M$ exist.
    Thus $h=0$ is a polynomial equation for the hypothetical
    hypersurface, and $\mathcal M$ is a local portion of this
    hypersurface, called a \emph{patch}.

    \emph{Step 1: Polynomial divisibility on $\mathcal M$.}
    After shrinking $\mathcal M$, we may assume that $h$ is
    irreducible over $\mathbb R$.  Indeed, factor $h$ and fix a point
    of $\mathcal M$.  Since $\nabla h\neq0$ there, exactly one
    irreducible factor vanishes at that point, with multiplicity
    one.  The other factors remain nonzero in a sufficiently small
    neighborhood.  Replacing $h$ by the vanishing factor and
    restricting $\mathcal M$ to a connected portion of this
    neighborhood preserves the hypersurface and its tangent spaces.

    We will use the following polynomial divisibility fact: if a
    real polynomial $F$ vanishes on $\mathcal M$, then $h$ divides $F$.
    Fix a point $p\in\mathcal M$.  Since $\nabla h(p)\neq0$,
    we can relabel the coordinates as
    $(u,y)\in\mathbb R^{D-1}\times\mathbb R$ so that
    $\partial_y h(p)\neq0$.  The implicit function theorem gives
    a local parametrization $X=(u,\varphi(u))$ of $\mathcal M$,
    where $u$ ranges over a nonempty open set
    $U\subseteq\mathbb R^{D-1}$.  On this graph,
    \[
        h(u,\varphi(u))=F(u,\varphi(u))=0
        \qquad\text{for every }u\in U.
    \]

    Suppose that $h$ does not divide $F$.  Regard both polynomials
    as polynomials in the single variable $y$, and allow their
    coefficients to be rational functions of
    $u=(u_1,\ldots,u_{D-1})$.  These coefficients form the field
    $K=\mathbb R(u_1,\ldots,u_{D-1})$.
    The condition $\partial_y h(p)\neq0$ ensures that $h$ has
    positive degree in $y$.  Moreover, its coefficients in
    $\mathbb R[u]$ have no common nonconstant factor, since such
    a factor would contradict the irreducibility of $h$ in
    $\mathbb R[u,y]$.  Gauss's lemma therefore implies that $h$
    remains irreducible in $K[y]$, and that divisibility by $h$
    of a polynomial $F\in\mathbb R[u,y]$ is the same in these
    two polynomial rings.  Thus $h$ does not divide $F$ in $K[y]$
    either, so their greatest common divisor in $K[y]$ is $1$.

    The Euclidean algorithm for polynomials in one variable over
    a field now gives a B\'ezout identity
    \[
        \widetilde B(u,y)h(u,y)+\widetilde C(u,y)F(u,y)=1,
        \qquad \widetilde B,\widetilde C\in K[y].
    \]
    Choose a nonzero polynomial $c(u)$ that is a common denominator
    of the finitely many rational coefficients of
    $\widetilde B$ and $\widetilde C$.  Multiplying by $c(u)$ gives
    $B=c\widetilde B$ and $C=c\widetilde C$ in $\mathbb R[u,y]$,
    and hence the polynomial identity
    \[
        B(u,y)h(u,y)+C(u,y)F(u,y)=c(u),
        \qquad c\neq0.
    \]
    This identity is valid for every $(u,y)$, including points
    where $c(u)=0$.  Substituting $y=\varphi(u)$ makes both terms
    on the left zero, so $c(u)=0$ for every $u\in U$.
    A real polynomial vanishing on a nonempty open set is the zero
    polynomial: one can apply the corresponding one-variable fact
    successively to each coordinate on an open box contained in $U$.
    This contradicts the choice $c\neq0$ and proves that $h$ divides $F$.

    \emph{Step 2: At least one generator is resonant.}
    Since $T_{X}\mathcal M=\ker dh_{X}$, the assumed
    tangency of the two batch fields means precisely
    \[
        dh_{X}\bigl[f_r^{(N)}(X)\bigr]
        =\sum_{a=1}^N\sum_{i=1}^n
        \partial_{x_i^a}h(X)\cdot
        f_r^{(N)}(X)_i^a=0,
        \qquad X\in\mathcal M,\quad r=1,2,
    \]
    where $\partial_{x_i^a}h\in\mathbb R^d$ is the gradient with
    respect to the coordinates of token $x_i^a$.
    For one generator $(W,V)$, fix $X\in\mathcal M$ and write
    \[
        s_{a,i,j}=(x_j^a)^\top W x_i^a,
        \qquad Z_{a,i}=\sum_{j=1}^n e^{s_{a,i,j}},
        \qquad b_{a,i,j}=(Vx_j^a)\cdot\partial_{x_i^a}h(X).
    \]
    The tangency equation is
    \[
        0=\sum_{a,i}\frac{\sum_j b_{a,i,j}e^{s_{a,i,j}}}{Z_{a,i}}.
    \]
    Each $Z_{a,i}$ is positive.  Multiplying by
    $\prod_{a,i}Z_{a,i}$ cancels the denominator in each summand,
    giving
    \[
        0=\sum_{a,i}\left(\sum_j b_{a,i,j}e^{s_{a,i,j}}\right)
        \prod_{(b,k)\neq(a,i)}\left(\sum_\ell e^{s_{b,k,\ell}}\right).
    \]
    Expanding the products amounts to choosing one key index for each
    token position.  Such a choice is a map
    $\sigma:\{1,\ldots,N\}\times\{1,\ldots,n\}\to\{1,\ldots,n\}$,
    where $\sigma(a,i)$ selects a key in sequence $a$; different
    token positions may select the same key.  For a fixed $\sigma$, every
    summand has the common exponential factor
    $\prod_{a,i}e^{s_{a,i,\sigma(a,i)}}
    =e^{\sum_{a,i}s_{a,i,\sigma(a,i)}}$.
    Summing over the distinguished token position $(a,i)$ adds the coefficients
    $b_{a,i,\sigma(a,i)}$.  Thus we obtain
    \begin{equation}\label{eq:two-gaussian-tangency-expansion}
        \sum_\sigma e^{P_\sigma(X)}A_\sigma(X)=0
        \quad\text{on }\mathcal M,
    \end{equation}
    where
    \[
        P_\sigma=\sum_{a,i}(x_{\sigma(a,i)}^a)^\top W x_i^a,
        \qquad
        A_\sigma=\sum_{a,i}
        (Vx_{\sigma(a,i)}^a)\cdot\partial_{x_i^a}h.
    \]
    Call the generator nonresonant on $\mathcal M$ if every difference
    $P_\sigma-P_\tau$, $\sigma\neq\tau$, is nonconstant on $\mathcal M$.
    We first suppose that both generators are nonresonant on
    $\mathcal M$.  We will show that $T_X\mathcal M=\mathbb R^D$
    for every $X\in\mathcal M$, contradicting the fact that
    $\mathcal M$ is a hypersurface of dimension $D-1$.

    For either generator $(W,V)$,
    \Cref{lem:two-gaussian-exponential-independence} applies to
    \eqref{eq:two-gaussian-tangency-expansion} and gives
    $A_\sigma=0$ on $\mathcal M$ for every $\sigma$.
    Let $L_V$ denote the linear batch field defined by
    $(L_V(X))_i^a=Vx_i^a$, which applies the same matrix $V$
    to every token.  Taking $\sigma(a,i)=i$ for every token position gives
    \[
        A_\sigma(X)=\sum_{a,i}(Vx_i^a)\cdot\partial_{x_i^a}h(X)
        =dh_X[L_V(X)]=0
        \qquad\text{on }\mathcal M.
    \]
    Hence $L_V$ is tangent to $\mathcal M$.

    Now take two of the identities $A_\sigma=A_\tau=0$ whose
    index maps agree at every token position except $(a,i)$,
    where $\sigma(a,i)=j$ and $\tau(a,i)=k$.
    Keeping the same $X$, $W$, and $V$, subtracting these identities gives
    \[
        0=A_\sigma(X)-A_\tau(X)
        =V(x_j^a-x_k^a)\cdot\partial_{x_i^a}h(X).
    \]
    Thus the vector in $\mathbb R^D$ with component
    $V(x_j^a-x_k^a)$ at position $(a,i)$ and zero at every other
    position belongs to $T_X\mathcal M$.  Since this holds for every
    $X\in\mathcal M$, the corresponding linear field is tangent to
    $\mathcal M$.

    Fix $a,i,j,k$.  We have obtained tangent linear fields whose
    only nonzero component is $B(x_j^a-x_k^a)$ at position $(a,i)$,
    for $B=V_1$ and $B=V_2$.
    The fields $L_{V_1}$ and $L_{V_2}$ are also tangent to $\mathcal M$.
    Taking the bracket of $L_{V_r}$ with such a field, for $r=1,2$,
    gives a field of the same form, with $B$ replaced by $BV_r-V_rB$.
    Since linear combinations and Lie brackets preserve tangency,
    repeating these operations gives tangent fields for all matrices
    in the linear span of nested commutators of $V_1,V_2$, including
    the generators themselves.  This span is
    $\operatorname{Lie}_{\mathrm{mat}}(V_1,V_2)$, which equals
    $\mathrm{gl}_d(\mathbb R)$ by condition~1 of
    \Cref{app:two-gaussian-matrix-conditions}.
    Choose $j\neq k$.  Since the tokens within each sequence are
    distinct, $v=x_j^a-x_k^a\neq0$, and
    $\{Bv:B\in\mathrm{gl}_d(\mathbb R)\}=\mathbb R^d$.
    Consequently, $T_X\mathcal M$ contains every vector supported
    at a single token position.  These vectors span $\mathbb R^D$,
    contradicting $\dim T_X\mathcal M=D-1$.

    Consequently, for at least one of the two generators $(W,V)$,
    there exist distinct index maps $\sigma,\tau$ and a real constant
    $c$ such that $P_\sigma-P_\tau=c$ on $\mathcal M$.

    \emph{Step 3: At most one generator is resonant.}
    Fix such a generator, such maps, and this constant.
    Order all $M=Nn$ tokens and write the chosen exponent difference as
    \[
        \begin{aligned}
            Q_{C,W}(X)&\coloneqq P_\sigma(X)-P_\tau(X)\\
            &=\sum_{r,s=1}^M C_{rs}x_r^\top W x_s
            =X^\top(C\otimes W)X,
        \end{aligned}
    \]
    where $C\neq0$ is block diagonal by sequences, with key rows and
    query columns, and $X=(x_1^\top,\ldots,x_M^\top)^\top$.
    By construction,
    \[
        Q_{C,W}(X)=P_\sigma(X)-P_\tau(X)=c
        \qquad\text{for every }X\in\mathcal M.
    \]
    Let $H=H^\top$ be the symmetric matrix representing this
    quadratic form, so that
    \[
        Q_{C,W}(X)=X^\top H X.
    \]
    A matrix and its symmetric part define the same quadratic form.
    Thus, writing $S=(W+W^\top)/2$ and $K=(W-W^\top)/2$, we obtain
    \begin{equation}\label{eq:two-gaussian-tensor-decomposition}
        H=\frac{C\otimes W+C^\top\otimes W^\top}{2}
        =\operatorname{sym}(C)\otimes S
        +\operatorname{skew}(C)\otimes K.
    \end{equation}
    If $\operatorname{sym}(C)\neq0$, choose a fixed
    $t=(t_1,\ldots,t_M)\in\mathbb R^M$ with $t^\top Ct\neq0$.
    Restrict $Q_{C,W}$ to the linear subspace
    $X=(t_1z,\ldots,t_Mz)$, with $z\in\mathbb R^d$ variable:
    each token $x_r$ is replaced by the scalar multiple $t_rz$
    of the same vector $z$.  Substitution gives
    \[
        Q_{C,W}(t_1z,\ldots,t_Mz)
        =\sum_{r,s}C_{rs}t_rt_s\,z^\top Wz
        =(t^\top Ct)z^\top Sz.
    \]
    This restricted quadratic form has matrix $(t^\top Ct)S$,
    of rank $d$ because $t^\top Ct\neq0$ and $S$ is invertible.
    Restriction to a linear subspace cannot increase the rank of
    a quadratic form, so $\operatorname{rank}(H)\geq d> 2$.
    This is an algebraic restriction of the polynomial; the
    substituted configurations need not belong to $\mathcal M$
    or to $\mathcal B_{N,n,d}$.
    If $\operatorname{sym}(C)= 0$, then $C$ is skew and 
    $\operatorname{rank}(H) = \operatorname{rank}(C)\operatorname{rank}(K)\geq4> 2$. That means that in any case: $\operatorname{rank}(H)> 2$. Thus $Q_{C,W}-c$ is irreducible over $\mathbb C$ for every
    constant $c$: a product of two affine linear factors would have
    quadratic part of rank at most two.
    Since $Q_{C,W}-c$ vanishes on $\mathcal M$ by construction,
    the divisibility fact above gives
    $h\mid Q_{C,W}-c$, so irreducibility forces $h$ to be
    proportional to $Q_{C,W}-c$.

    Suppose now that both generators are resonant on $\mathcal M$.
    Their index pairs may differ, giving coefficient matrices
    $C_1,C_2\neq0$ and constants $c_1,c_2$.
    By the preceding argument, both $Q_{C_1,W_1}-c_1$ and
    $Q_{C_2,W_2}-c_2$ are proportional to the same polynomial $h$.
    Hence, for some $\lambda\in\mathbb R\setminus\{0\}$,
    \[
        Q_{C_1,W_1}-c_1=\lambda\bigl(Q_{C_2,W_2}-c_2\bigr).
    \]
    Comparing the terms of degree two gives
    \[
        Q_{C_1,W_1}=\lambda Q_{C_2,W_2},
        \qquad H_1=\lambda H_2,
    \]
    where $H_r$ is the symmetric matrix representing $Q_{C_r,W_r}$.
    The decomposition in \eqref{eq:two-gaussian-tensor-decomposition}
    then forces either $S_1,S_2$ or $K_1,K_2$ to be proportional,
    contradicting condition~2 of \Cref{app:two-gaussian-matrix-conditions}.

    \emph{Step 4: Excluding the remaining case.}
    Thus exactly one generator is resonant; denote it by $r$ and
    the nonresonant generator by $s$.
    Applying the nonresonance argument to generator $s$ and choosing
    $\sigma(a,i)=i$ shows that $L_{V_s}$ is tangent to $\mathcal M$.
    After multiplying $h$ by a nonzero constant, we may write
    $h=Q_{C,W_r}-c$.  Tangency means that the polynomial
    $F(X)\coloneqq dh_X[L_{V_s}(X)]$ vanishes on $\mathcal M$.
    Since $L_{V_s}$ applies $V_s$ to every token, differentiating
    each quadratic term in both token variables gives
    \[
        \begin{aligned}
            F(X)
            &=\sum_{p,q}C_{pq}\bigl[
                (V_sx_p)^\top W_rx_q
                +x_p^\top W_r(V_sx_q)\bigr]\\
            &=Q_{C,V_s^\top W_r+W_rV_s}(X).
        \end{aligned}
    \]
    The divisibility fact gives $h\mid F$.  As $h$ has degree two
    and $F$ has degree at most two, the quotient is a scalar:
    \[
        F=\gamma h=\gamma(Q_{C,W_r}-c),
        \qquad \gamma\in\mathbb R.
    \]
    Comparing the terms of degree two therefore yields
    \[
        Q_{C,V_s^\top W_r+W_rV_s}=\gamma Q_{C,W_r}
    \]
    (the constant terms also give $\gamma c=0$).

    The symmetric and skew-symmetric parts of
    $V_s^\top W_r+W_rV_s$ are respectively
    $V_s^\top S_r+S_rV_s$ and $V_s^\top K_r+K_rV_s$.
    Using \eqref{eq:two-gaussian-tensor-decomposition} and separating
    the symmetric and skew-symmetric factors in the token indices
    therefore gives
    \[
        \begin{aligned}
            \operatorname{sym}(C)\otimes
            (V_s^\top S_r+S_rV_s-\gamma S_r)&=0,\\
            \operatorname{skew}(C)\otimes
            (V_s^\top K_r+K_rV_s-\gamma K_r)&=0.
        \end{aligned}
    \]
    If $\operatorname{sym}(C)\neq0$, the first identity implies
    $V_s^\top S_r+S_rV_s=\gamma S_r$.
    Otherwise $\operatorname{skew}(C)=C\neq0$, and the second implies
    $V_s^\top K_r+K_rV_s=\gamma K_r$.
    Either relation contradicts condition~3 of
    \Cref{app:two-gaussian-matrix-conditions}, namely
    \eqref{eq:two-gaussian-nonconformal}.
\end{proof}

\subsection{Polynomial lifts and common invariant divisors}
\label{app:two-gaussian-newton}

\paragraph{Some notations and background.}
Write $X=(x_1,\ldots,x_D)\in\mathbb C^D$ for the complexified batch.
Enumerate the $p=2Nn^2$ score quadratics in
\eqref{eq:two-gaussian-score-quadratics} as $Q_1,\ldots,Q_p$.
For each $Q_\nu$, introduce a formal variable $z_\nu$ representing
the exponential $e^{Q_\nu(X)}$.
In the algebraic calculations, the $z_\nu$ are algebraically independent
of one another and of the state coordinates: the relation
$z_\nu=e^{Q_\nu(X)}$ is imposed only when evaluating the resulting
expressions at the original exponential functions.

Since exponentials never vanish, we also allow inverses of these
variables and work in the Laurent polynomial ring
\[
    A=\mathbb C[x_1,\ldots,x_D,z_1^{\pm1},\ldots,z_p^{\pm1}],
\]
where $z_\nu^{\pm1}$ means that both $z_\nu$ and its inverse
$z_\nu^{-1}=1/z_\nu$ are allowed.
Thus an element of $A$ has the form
\[
    P(X,z)=\sum_{\beta\in\mathbb Z^p}p_\beta(X)z^\beta,
    \qquad z^\beta=\prod_{\nu=1}^p z_\nu^{\beta_\nu},
\]
with only finitely many nonzero coefficients
$p_\beta\in\mathbb C[X]=\mathbb C[x_1,\ldots,x_D]$.
The exponents of the $z_\nu$ may therefore be negative, while the
coefficients remain polynomials in the state coordinates.

We also write $z_{r,a,i,j}$ for the variable corresponding to
$Q_{r,a,i,j}$; the variables with a fixed $r\in\{1,2\}$ form that
generator's score block.  A single subscript $\nu\in\{1,\ldots,p\}$
enumerates all these variables, and $z=(z_\nu)_{\nu=1}^p$.

For a linear state field $L(X)=(L_1(X),\ldots,L_D(X))$, we also
use $L$ to denote the associated differential operator: for a
polynomial $f$ in the state coordinates,
\[
    (Lf)(X)\coloneqq df_X[L(X)]
    =\sum_{k=1}^D L_k(X)\frac{\partial f}{\partial x_k}(X).
\]
In particular, $LQ_\nu$ is the derivative of $Q_\nu$ in the
direction of the field $L$.
Define the horizontal lift of $L$ by
\[
    \mathcal D_L
    =L+\sum_{\nu=1}^p z_\nu(LQ_\nu)\partial_{z_\nu}.
\]
For a key choice $\sigma$, put
$z_r^\sigma=\prod_{a,i}z_{r,a,i,\sigma(a,i)}$ and
$(L_{r,\sigma}(X))_i^a=V_rx_{\sigma(a,i)}^a$.
Multiplying the lifted attention field by the product of its row
denominators yields the exact polynomial derivation
\begin{equation}\label{eq:two-gaussian-polynomial-lift}
    \delta_r=\sum_\sigma z_r^\sigma\mathcal D_{L_{r,\sigma}}.
\end{equation}
Indeed, let $F_r(X,z)$ denote the attention field with each
$e^{Q_{r,a,i,j}(X)}$ replaced by $z_{r,a,i,j}$, and set
$H_r(z)=\prod_{a,i}(\sum_j z_{r,a,i,j})$.
Clearing denominators and expanding the product of row sums gives
\[
    H_rF_r=\sum_\sigma z_r^\sigma L_{r,\sigma}.
\]
The lift $F\mapsto F+\sum_\nu z_\nu dQ_\nu[F]\,\partial_{z_\nu}$
is linear in the field components.  Applying it to this identity
therefore gives \eqref{eq:two-gaussian-polynomial-lift}.
The identity key choice $\sigma(a,i)=i$ gives $L_{r,\sigma}=L_{V_r}$.

\begin{lemma}[Reduction to a single monomial in $z$]
\label{lem:two-gaussian-newton-rigidity}
    Suppose the score quadratics are linearly independent and
    $L_{V_1},L_{V_2}$ are invertible on homogeneous quadratics.
    If $0\neq P\in A$ satisfies
    $\delta_rP=B_rP$ with $B_r\in A$, $r=1,2$, then
    \[
        P=z^\beta p(X)
    \]
    for a Laurent monomial $z^\beta$ and a polynomial $p\in\mathbb C[X]$.
\end{lemma}
\begin{proof}
    Write $P=\sum_\beta p_\beta(X)z^\beta$ and
    $Q_\beta=\sum_\nu\beta_\nu Q_\nu$.  The horizontal lift preserves
    each exponent:
    \begin{equation}\label{eq:two-gaussian-horizontal-monomial}
        \mathcal D_L(p_\beta z^\beta)
        =z^\beta\bigl(Lp_\beta+(LQ_\beta)p_\beta\bigr).
    \end{equation}

    \emph{Step 1: Determine the possible monomials in $B_r$.}
    Fix $r\in\{1,2\}$ and first suppose that $B_r\neq0$.
    We use an elementary property of Laurent polynomials: the
    smallest and largest exponents of any one variable add under
    multiplication of nonzero polynomials.  The same holds for the
    smallest and largest sums of exponents over any fixed group of
    variables.  Indeed, group the terms by the exponent or sum in
    question.  The product of the two groups with largest values
    is nonzero and gives the largest value in the product; the
    argument for the smallest values is identical.

    By \eqref{eq:two-gaussian-horizontal-monomial}, every monomial
    of $\delta_rP$ comes from multiplying a monomial of $P$ by
    some $z_r^\sigma$.  Such multiplication never decreases any
    individual exponent.  Since $\delta_rP=B_rP$, additivity of
    the smallest exponents implies that every exponent in $B_r$
    is nonnegative.  The variables of the other score block do
    not occur in $z_r^\sigma$: their smallest exponents cannot
    decrease and their largest exponents cannot increase in
    $\delta_rP$.  Thus their smallest exponents in $B_r$ are at
    least zero and their largest at most zero, so they do not
    occur in $B_r$.

    For each position $(a,i)$, multiplication by $z_r^\sigma$
    increases the sum of the exponents of
    $z_{r,a,i,1},\ldots,z_{r,a,i,n}$ by exactly one.
    The same additivity argument shows that, among the monomials
    of $B_r$, the smallest such sum is at least one and the largest
    is at most one.  Every such sum therefore equals one.
    As the individual exponents are nonnegative integers, each
    monomial of $B_r$ contains exactly one of these variables,
    to the first power, for every position $(a,i)$.
    Moreover, \eqref{eq:two-gaussian-horizontal-monomial} increases
    degree in $X$ by at most two.  Since this degree adds in
    nonzero products, $\deg_X B_r\leq2$.  We have proved
    \begin{equation}\label{eq:two-gaussian-cofactor}
        B_r=\sum_\sigma b_{r,\sigma}(X)z_r^\sigma,
        \qquad \deg_X b_{r,\sigma}\leq2.
    \end{equation}
    If $B_r=0$, the same formula holds with all coefficients zero.

    \emph{Step 2: Compare a monomial of largest weight.}
    Assign a real weight $w_\nu$ to each variable $z_\nu$, so that
    the weight of $z^\beta$ is
    $w\cdot\beta=\sum_\nu w_\nu\beta_\nu$.
    Choose weights satisfying
    \[
        w_{r,a,i,i}>w_{r,a,i,j}
        \qquad\text{for every }a,i\text{ and }j\neq i.
    \]
    Then $z_r^{\mathrm{id}}$, where $\mathrm{id}(a,i)=i$, has
    strictly larger weight than every other $z_r^\sigma$.
    The weights in the other score block are unrestricted.
    We may also choose the weights so that distinct monomials of
    $P$ have different weights: ties require one of finitely many
    equalities $w\cdot(\alpha-\beta)=0$, which an arbitrarily small
    perturbation avoids while preserving the strict inequalities.

    Let $\beta$ be the exponent of the monomial of $P$ with largest
    weight.  In $\delta_rP=B_rP$, the coefficient of
    $z^\beta z_r^{\mathrm{id}}$ can only come from this monomial
    of $P$ and the identity choice: every other pair has smaller
    total weight.  Comparing this coefficient gives
    \[
        L_{V_r}p_\beta+
        (L_{V_r}Q_\beta-b_{r,\mathrm{id}})p_\beta=0.
    \]
    Since $L_{V_r}p_\beta$ has degree at most $\deg_X p_\beta$,
    comparison at degree $\deg_X p_\beta+2$, followed by cancelling
    the nonzero homogeneous part of highest degree of $p_\beta$,
    gives
    \begin{equation}\label{eq:two-gaussian-exposed-vertex}
        L_{V_r}Q_\beta=(b_{r,\mathrm{id}})_{[2]},
    \end{equation}
    where $[2]$ denotes the homogeneous quadratic part.
    The right side does not depend on the weights.  Invertibility
    of $L_{V_r}$ on homogeneous quadratics therefore determines
    $Q_\beta$ uniquely, and linear independence of the $Q_\nu$
    determines $\beta$ uniquely.  For this fixed $r$, the same
    exponent, denoted by $\beta_*$, thus has largest weight for
    every such choice without ties.

    \emph{Step 3: Show that all monomials have the same exponent.}
    Every weight vector satisfying the strict inequalities above
    can be approximated by choices without ties.  Passing to the
    limit gives $w\cdot(\alpha-\beta_*)\leq0$ for every exponent
    $\alpha$ appearing in $P$ and every such weight vector.
    Split the weights and exponents into their two score blocks.
    Fix any weights $w_r$ satisfying the strict inequalities and
    let $w_{3-r}$ vary freely.  We obtain
    \[
        w_r\cdot(\alpha_r-\beta_{*,r})
        +w_{3-r}\cdot(\alpha_{3-r}-\beta_{*,3-r})\leq0
        \quad\text{for every }w_{3-r}.
    \]
    If $\alpha_{3-r}-\beta_{*,3-r}\neq0$, taking
    $w_{3-r}=t(\alpha_{3-r}-\beta_{*,3-r})$ and letting
    $t\to+\infty$ contradicts this inequality.
    Hence every exponent appearing in $P$ has the same coordinates
    in block $3-r$.  Applying the argument for $r=1$ and $r=2$
    shows that all exponents coincide.  Thus
    $P=z^\beta p(X)$, as claimed.
\end{proof}

\subsection{Analytic hypersurfaces and the Ax--Schanuel theorem}
\label{app:two-gaussian-ax}

For functions $f_1,\ldots,f_\ell$, the notation
$\mathbb C(f_1,\ldots,f_\ell)$ denotes the field of rational
expressions in these functions with complex coefficients.
Its \emph{transcendence degree}, denoted by
$\operatorname{trdeg}_{\mathbb C}\mathbb C(f_1,\ldots,f_\ell)$,
is the largest number of these functions that are
\emph{algebraically independent}: no nonzero polynomial with complex
coefficients vanishes identically when evaluated on them.
For example, $\operatorname{trdeg}_{\mathbb C}\mathbb C(t,t^2)=1$,
since $t^2$ is already a polynomial in $t$.

We use the following theorem for complex-analytic functions.

\begin{theorem}[Ax--Schanuel; \citealp{ax1971schanuel}]
\label{thm:two-gaussian-ax}
    Let $g_1,\ldots,g_s$ be complex-analytic functions on a connected
    open neighborhood $U\subseteq\mathbb C^m$ of the origin.
    Suppose they are linearly independent over $\mathbb Q$ modulo
    constants: if $\sum_i q_i g_i$ is constant on $U$, with
    $q_i\in\mathbb Q$, then every $q_i$ is zero.  Then
    \[
        \operatorname{trdeg}_{\mathbb C}
        \mathbb C(g_1,\ldots,g_s,e^{g_1},\ldots,e^{g_s})
        \geq s+\max_{t\in U}\operatorname{rank}\left(
        \frac{\partial g_i}{\partial t_j}(t)\right).
    \]
    The rank on the right is the largest rank attained by the
    Jacobian on $U$.
\end{theorem}
Only the functions on this neighborhood are needed; no extension
to all of $\mathbb C^m$ is required.

\begin{lemma}[An analytic singular hypersurface gives a lifted divisor]
\label{lem:two-gaussian-ax-divisor}
    Assume the conditions of
    \Cref{app:two-gaussian-matrix-conditions}, and suppose a Lie
    determinant is nonzero at some state.  A smooth real-analytic
    hypersurface patch $H\subseteq\Sigma_{\theta,N,n}$ tangent to
    both fields would give a nonunit irreducible $P\in A$ satisfying
    $\delta_rP\in(P)$ for $r=1,2$.
    Moreover, no nonzero polynomial with complex coefficients
    vanishes on any nonempty relatively open subset of $H$.
\end{lemma}
\begin{proof}
    Suppose a nonzero real polynomial vanished on a nonempty
    relatively open subset of $H$.  Restrict to a connected open
    portion.  Factoring the polynomial and using the identity
    theorem for real-analytic functions shows that one real
    irreducible factor $h$ vanishes throughout this portion.
    The set $\{h=0,\nabla h=0\}$ has dimension at most $D-2$, so
    it cannot contain this smooth $(D-1)$-dimensional portion.
    Near a point where $\nabla h\neq0$, the portion is therefore
    relatively open in the smooth algebraic hypersurface $h=0$
    and remains tangent to both fields, contradicting
    \Cref{lem:two-gaussian-no-algebraic-barrier}.
    Separating real and imaginary parts gives the same conclusion
    for polynomials with complex coefficients.

    Extend a real-analytic local parametrization of $H$ to complex
    parameters using its convergent power series.  After restricting
    the parameter domain, this gives a complex-analytic map
    $X(t)=(x_1(t),\ldots,x_D(t))$ on a connected open neighborhood
    $U\subseteq\mathbb C^{D-1}$ of $0$, with
    $\operatorname{rank}DX(t)=D-1$; its restriction to real
    parameters parametrizes an open portion of $H$.
    The image $X(U)$ is \emph{Zariski dense} in $\mathbb C^D$:
    this means that the only polynomial in $\mathbb C[X]$
    vanishing throughout $X(U)$ is the zero polynomial.
    Indeed, any such identity would also hold on the real portion
    of $H$, which was just ruled out.
    Row denominators are nonzero at $X(0)$ and remain nonzero
    after shrinking $U$.
    The tangency identities likewise extend from real to complex
    parameters by their convergent power series.

    Apply \Cref{thm:two-gaussian-ax} to the $p+D$ functions
    \[
        Q_1(X(t)),\ldots,Q_p(X(t)),x_1(t),\ldots,x_D(t).
    \]
    They are linearly independent modulo constants even over
    $\mathbb C$: a relation would give a polynomial vanishing on
    $X(U)$, hence the zero polynomial.  Separating its homogeneous
    degrees and using independence of the score quadratics forces
    every coefficient to vanish.
    Their Jacobian has rank $D-1$ because it includes $DX$.
    Consequently,
    \[
        \operatorname{trdeg}_{\mathbb C}
        \mathbb C\bigl(X(t),e^{Q(X(t))},e^{X(t)}\bigr)
        \geq p+2D-1,
    \]
    where exponentials of tuples are taken componentwise.
    Removing the $D$ auxiliary coordinate exponentials lowers
    transcendence degree by at most $D$, giving
    \begin{equation}\label{eq:two-gaussian-ax-bound}
        \operatorname{trdeg}_{\mathbb C}
        \mathbb C\bigl(X(t),e^{Q(X(t))}\bigr)\geq p+D-1.
    \end{equation}
    These auxiliary exponentials are used only in the proof.

    Let $\mathcal O(U)$ denote the ring of complex-analytic functions
    on $U$.  The substitution homomorphism
    \[
        A\longrightarrow\mathcal O(U),
        \qquad x_i\mapsto x_i(t),\quad
        z_\nu\mapsto e^{Q_\nu(X(t))}
    \]
    has a prime kernel $I$, since on connected $U$ a product of
    two nonzero complex-analytic functions cannot vanish identically.
    The nonzero Lie determinant, after replacing exponentials by
    formal variables and clearing row denominators, gives a
    nonzero element of $I$: it vanishes on $H$, but is not
    identically zero under ambient analytic substitution.
    This relation shows that the $D+p$ functions
    $X(t),e^{Q(X(t))}$ are not algebraically independent.
    Together with \eqref{eq:two-gaussian-ax-bound}, it follows that
    their generated field has transcendence degree exactly $D+p-1$.

    Choose $D+p-1$ algebraically independent functions among these
    generators, and let $K$ be their field of rational expressions.
    The remaining function satisfies a nonzero polynomial equation
    with coefficients in $K$.  A nonzero polynomial of least degree that
    vanishes at this function divides every other such polynomial,
    by the Euclidean division algorithm.
    Clearing its denominators and removing the common factor of
    its coefficients gives a polynomial $P$ in the original variables.
    Gauss's lemma makes $P$ irreducible and shows that every polynomial
    relation is divisible by $P$.
    This remains true for Laurent polynomial relations after clearing
    negative powers of the $z_\nu$.  Hence $I=(P)$ in $A$.
    Inverting the $z_\nu$ preserves irreducibility unless $P$ becomes
    a unit.  This is impossible: $P$ vanishes under substitution,
    whereas all inverted variables become nonzero exponentials.

    The chain rule intertwines the horizontal lifts with
    differentiation on the exponential graph.
    Tangency to $H$ therefore makes $I$ invariant under the cleared
    derivations \eqref{eq:two-gaussian-polynomial-lift}.
    Hence $\delta_rP=B_rP$ for some $B_r\in A$, as required.
\end{proof}

\subsection{Connectedness of the interpolation domain}
\label{app:two-gaussian-conclusion}

\thmtwogaussianattention*

\begin{proof}[Proof of \Cref{thm:two-gaussian-connected-domain}]
    Fix a pair in the probability-one event of
    \Cref{lem:two-gaussian-generic-rank} and
    \Cref{app:two-gaussian-matrix-conditions}, and fix $N,n$.
    Put $D=Nnd$ and $\Sigma=\Sigma_{\theta,N,n}$.

    \smallskip
    \noindent\emph{Step 1: A finite rank test and its preservation along flows.}
    Write the reciprocal softmax denominators as
    \[
        u_{r,a,i}(X)
        \coloneqq\left(\sum_j e^{Q_{r,a,i,j}(X)}\right)^{-1}.
    \]
    Define
    \[
        \mathcal A
        =\mathbb R\bigl[x_1,\ldots,x_D,
        e^{Q_1(X)},\ldots,e^{Q_p(X)},
        (u_{r,a,i}(X))_{r,a,i}\bigr].
    \]
    This algebra is fixed for the chosen $\theta,N,n$; $X$ is the
    variable of its functions.  More precisely, its generators depend
    only on $W_1,W_2$ and $N,n$: the entries of $V_1,V_2$ appear
    only as real coefficients in the attention fields.
    Here square brackets denote all finite sums and products of the
    listed functions, with real coefficients.  No algebraic independence
    of these functions is assumed.
    Every function in $\mathcal A$ is real analytic because
    the softmax denominators are positive.
    The attention formula \eqref{eq:two-gaussian-field} shows that
    every component of the two batch fields belongs to $\mathcal A$.

    Partial derivatives also remain in $\mathcal A$.
    Indeed, derivatives of the state coordinates are constants, while
    \[
        \begin{aligned}
            \partial_{x_k}e^{Q_\nu}
            &=(\partial_{x_k}Q_\nu)e^{Q_\nu},\\
            \partial_{x_k}u_{r,a,i}
            &=-u_{r,a,i}^{2}\sum_j
            (\partial_{x_k}Q_{r,a,i,j})e^{Q_{r,a,i,j}}.
        \end{aligned}
    \]
    Since the $Q_\nu$ are quadratic, their derivatives are linear
    polynomials in the state coordinates.  Both right sides therefore
    belong to $\mathcal A$, and the sum and product rules give the
    same conclusion for every function in $\mathcal A$.
    Finally, each component of a Lie bracket is a sum of products
    of field components and their first derivatives.  Hence all
    iterated Lie brackets of the two fields have components in
    $\mathcal A$ as well.

    Since $\mathcal A$ is generated by finitely many functions over
    $\mathbb R$, Hilbert's basis theorem has the following consequence:
    any family of vectors with $D$ components in $\mathcal A$ is
    generated by finitely many of its members, using coefficients in
    $\mathcal A$ \citep[\S6, pp.~44--45]{shafarevich1990basic}.
    Applied to the two fields and all their iterated Lie brackets,
    this gives a finite list $Y_1,\ldots,Y_b$ among them such that
    every original field or iterated bracket $Z$ satisfies
    \[
        Z(X)=\sum_{\ell=1}^b a_{Z,\ell}(X)Y_\ell(X),
        \qquad a_{Z,\ell}\in\mathcal A.
    \]
    The same finite list $Y_1,\ldots,Y_b$ works for every $X$.
    Consequently, the columns of $Y(X)=(Y_1(X),\ldots,Y_b(X))$
    span the Lie algebra at $X$, and
    \[
        \Sigma=\{X\in\mathcal B_{N,n,d}:\operatorname{rank}Y(X)<D\}.
    \]
    Equivalently, all determinants formed by choosing $D$ columns
    of $Y(X)$ vanish.  This describes $\Sigma$ by finitely many
    real-analytic equations.  It has no interior by
    \Cref{lem:two-gaussian-generic-rank}.

    Each bracket $[f_r^{(N)},Y_\ell]$ is also such a linear combination,
    so $[f_r^{(N)},Y]=YC_r$ for an analytic $b\times b$ matrix $C_r$,
    with the bracket taken column by column.
    Let $\phi_t$ be the local flow of $f_r^{(N)}$ starting at $X$.
    Differentiating $D\phi_t(X)^{-1}Y(\phi_t(X))$ gives
    \[
        Y(\phi_t(X))=D\phi_t(X)Y(X)R(t),\qquad
        R'(t)=R(t)C_r(\phi_t(X)),\quad R(0)=I_b.
    \]
    Both $D\phi_t(X)$ and $R(t)$ are invertible, the latter by
    uniqueness for this linear matrix differential equation.
    Hence the rank of $Y$ is unchanged along either flow.
    In particular, a trajectory starting in $\Sigma$ remains in
    $\Sigma$ for as long as it stays in $\mathcal B_{N,n,d}$.

    \smallskip
    \noindent\emph{Step 2: The exceptional set has dimension at most $D-2$.}
    Suppose instead that $\dim_{\mathbb R}\Sigma=D-1$.
    A standard property of real-analytic zero sets then gives a
    neighborhood in which $\Sigma$ is a smooth real-analytic
    hypersurface $H$; see \citet[Section~2]{bierstone1988semianalytic}.
    The preceding flow property makes both fields tangent to $H$.
    The construction in \Cref{lem:two-gaussian-ax-divisor} gives a
    nonzero $P\in A$ such that
    \[
        P(X,e^{Q(X)})=0\quad\text{on }H,\qquad
        \delta_rP=B_rP\quad(r=1,2),\quad B_r\in A.
    \]
    By \Cref{lem:two-gaussian-newton-rigidity}, $P=z^\beta p(X)$
    for a nonzero polynomial $p\in\mathbb C[X]$.  Substitution yields
    \[
        0=e^{\sum_\nu\beta_\nu Q_\nu(X)}p(X)
        \qquad\text{on }H.
    \]
    Since the exponential factor never vanishes, $p$ vanishes on $H$.
    This contradicts the last assertion of
    \Cref{lem:two-gaussian-ax-divisor}.
    Thus $\dim_{\mathbb R}\Sigma\leq D-2$.

    \smallskip
    \noindent\emph{Step 3: The remaining domain is connected.}
    The sets excluded from $\mathcal B_{N,n,d}$ consist of proportional
    token pairs within a sequence, zero tokens, and pairs of sequences
    related by a permutation and an optional overall sign change.
    Their dimensions are, respectively, $D-d+1$, $D-d$, and $D-nd$,
    all at most $D-2$ because $d\geq3$.
    Together with $\Sigma$, they form the excluded set
    \[
        E=(\mathbb R^D\setminus\mathcal B_{N,n,d})\cup\Sigma.
    \]
    Real-analytic zero sets can be covered by countably many smooth
    pieces, each of dimension at most that of the set itself
    \citep[Corollary~2.11]{bierstone1988semianalytic}.
    Each smooth piece is a smooth submanifold of $\mathbb R^D$.
    If its dimension is $k$, it admits local parametrizations by
    $k$ real variables.
    Here $\Sigma$ is defined inside $\mathcal B_{N,n,d}$ by the
    analytic determinant equations of Step~1, and has dimension at
    most $D-2$ by Step~2.  The sets excluded from $\mathcal B_{N,n,d}$
    are defined by polynomial equations and satisfy the same dimension
    bound.  Applying the cited result separately to these sets therefore
    covers $E$ by smooth pieces of dimension at most $D-2$.
    We can then cover these pieces by countably many local
    parametrizations of the form
    \[
        Y=\psi(u),\qquad u\in U\subseteq\mathbb R^k,\qquad k\leq D-2,
    \]
    where $U$ is open and $\psi$ is smooth.

    Take $X_0,X_1\notin E$.  We seek an intermediate point $Z$ such
    that both segments $[X_0,Z]$ and $[Z,X_1]$ avoid $E$.
    Fix $i\in\{0,1\}$.  If $[X_i,Z]$ meets $E$ at $Y$, then
    $Y=X_i+\lambda(Z-X_i)$ for some $\lambda\in[0,1]$.
    Since $X_i\notin E$, we have $Y\neq X_i$ and thus $\lambda>0$.
    Solving for $Z$, setting $t=1/\lambda\geq1$, and writing
    $Y=\psi(u)$ gives
    \[
        Z=X_i+t(Y-X_i)=X_i+t(\psi(u)-X_i),
        \qquad u\in U,\quad t\geq1.
    \]
    Thus these forbidden choices of $Z$ lie on rays from $X_i$
    through points $Y\in E$, at or beyond $Y$.
    For each parametrization, this expression depends smoothly on
    $k+1\leq D-1$ parameters, namely $u$ and $t$, so its image has
    $D$-dimensional Lebesgue measure zero.  Taking the countable union
    over all parametrizations still gives a measure-zero set.
    We can therefore choose
    $Z$ outside the two measure-zero sets of forbidden intermediate
    points.  Both segments $[X_0,Z]$ and $[Z,X_1]$ then avoid $E$.
    This proves that $\mathcal R_{\theta,N,n}=\mathbb R^D\setminus E$
    is path connected.
    It is open by the rank test in Step~1, and dense with full
    Lebesgue measure in $\mathcal B_{N,n,d}$ by
    \Cref{lem:two-gaussian-generic-rank}.

    \smallskip
    \noindent\emph{Step 4: Interpolation with the same two sampled blocks.}
    At every point of this connected open domain, the two fields
    and their iterated Lie brackets span $\mathbb R^D$.
    The criterion in \Cref{cor:universal-interpolation-criterion}
    therefore joins any two configurations in the domain by finitely
    many successive flows of fields from
    $\{\pm f_1^{(N)},\pm f_2^{(N)}\}$, with the entire trajectory
    remaining in the domain.  The flow durations, and hence the
    terminal time $S>0$, may depend on the prescribed endpoints.

    Finally, the matrix conditions do not depend on $N,n$, and the
    probability-one event of \Cref{lem:two-gaussian-generic-rank}
    already holds simultaneously for every finite $N,n$.
    Hence all conclusions hold for the same sampled pair.
\end{proof}

\subsection{Exact interpolation at finite depth}
\label{app:two-gaussian-finite-depth}

The passage to residual layers follows from
\Cref{lem:finite-depth-lie-rank}, applied to the scalar multiples of
the two sampled fields.

\begin{proof}[Proof of \Cref{cor:two-gaussian-finite-depth}]
    Fix a draw in the probability-one event established in the proof of
    \Cref{thm:two-gaussian-connected-domain}.  For every $N\geq1$ and
    $n\geq2$, the two batch fields $f_1^{(N)},f_2^{(N)}$ are analytic
    on the open connected domain $\mathcal R_{\theta,N,n}$, and their
    Lie algebra has full rank there by
    \eqref{eq:two-gaussian-regular-domain}.
    Apply \Cref{lem:finite-depth-lie-rank} with $D=Nnd$,
    $\mathcal M=\mathcal R_{\theta,N,n}$, and
    \[
        \mathcal G
        =\{c f_r^{(N)}:c\in\mathbb R,\ r\in\{1,2\}\}.
    \]
    This family is closed under real scalar multiplication, and
    its two generators are smooth and have full Lie rank on
    $\mathcal M$.
    It joins every pair of configurations in this domain by finitely
    many residual layers
    \[
        \operatorname{Id}+c_k f_{W_{r_k},V_{r_k}}^{(N)},
        \qquad r_k\in\{1,2\},\quad c_k\in\mathbb R,
    \]
    with every intermediate configuration still in the domain.
    Each scalar gain and block index is shared across the batch.
    The matrices remain fixed; only the depth, indices, and gains
    depend on the prescribed endpoints.  This is universal
    interpolation at finite depth with $m_N=2$ in the sense of
    \Cref{sec:finite-depth}.
    The argument is deterministic for every $N,n$ once the draw is
    fixed, so no additional probability-one event is needed.
\end{proof}

\begin{remark}[Scope of the conclusion]
\label{rem:two-gaussian-scope}
    The exceptional set $\Sigma_{\theta,N,n}$ may be nonempty.
    The theorem gives two-field interpolation on the connected
    generic domain \eqref{eq:two-gaussian-regular-domain}, whereas
    \Cref{thm:attention-uniform-finite-generation} gives a bank of
    $2Nnd$ multihead blocks with Gaussian-initialized projections,
    valid at every point of $\mathcal B_{N,n,d}$ for any fixed number
    of heads and positive head widths.  Its rank-one special case is
    described in \Cref{rem:attention-rank-one}.
    {The two-block result requires a single head of width
    at least $d$.}
    \Cref{cor:two-gaussian-finite-depth} gives exact finite-depth
    interpolation on the same domain, with adjustable real scalar
    gains and the same two frozen fields.  No bound on the required
    depth is asserted. 
\end{remark}

\textcolor{black}{\section{Numerical tests of the head-width threshold}
\label{app:attention-width-threshold}}

\textcolor{black}{We test whether $m\ge t\coloneqq\lceil d/r\rceil$ frozen
noncausal single-head blocks of width $r<d$ suffice for generic full
lifted Lie rank in dimension $D=Nnd$.
The condition $mr\ge d$ is necessary: all bracket values lie tokenwise
in the sum of the output projection images, of dimension at most $mr$,
so the Lie rank is at most $Nn\min(d,mr)$.
A single generator has Lie rank at most one.}

\textcolor{black}{\textbf{Protocol.}
We use $(d,r)\in\{(3,1),(3,2),(4,1),(4,2),(4,3),(6,1),(6,2),(6,3)\}$,
$(N,n)\in\{(1,2),(2,2),(1,4),(2,3)\}$, and $m\in\{t-1,t,t+1\}$.
All query, key, value, and output projection entries are independent
centered Gaussians of variance $v\in\{1,1/d\}$.
Using seeds $0,1,2$ for each setting gives 576 cases with $6\le D\le36$.
The state $X$ is sampled independently with standard Gaussian coordinates.
Comparisons are paired: increasing $m$ takes prefixes of the same bank,
banks are shared across $(N,n)$, and the two variances rescale the same
base draws. Rescaling changes the fields.}

\textcolor{black}{We compute the spaces $W_k(X)$ of
\Cref{sec:computing-lie-rank} using nested PyTorch Jacobian-vector products
in float64, retaining all generated bracket fields.
Step $k$ includes brackets of length at most $k+1$.
Evaluated columns of norm above $10^{-14}$ are normalized, and singular
values exceeding $\max\{10^{-12},\varepsilon\sigma_1\}$ are counted.
We use $\varepsilon=10^{-8}$ for stopping and record $10^{-6},10^{-10}$
for sensitivity. Runs stop at full rank or the first pointwise rank
plateau, subject to $k\le6$ and 4096 words.
Analyticity justifies plateau stopping at generic points for exact
ranks; numerical plateaus can be premature.}

{\textbf{Results.}
The table reports full-rank counts at the final step of the original
runs, before any continuation. Each entry contains 96 cases.}
\begin{center}
\textcolor{black}{%
\begin{tabular}{ccrr}
\toprule
Projection variance $v$ & Blocks $m$ & Full rank at $10^{-8}$ & At all three tolerances \\
\midrule
$1$   & $t-1$ & $0/96$  & $0/96$  \\
$1$   & $t$   & $94/96$ & $89/96$ \\
$1$   & $t+1$ & $96/96$ & $96/96$ \\
\midrule
$1/d$ & $t-1$ & $0/96$  & $0/96$  \\
$1/d$ & $t$   & $96/96$ & $90/96$ \\
$1/d$ & $t+1$ & $96/96$ & $96/96$ \\
\bottomrule
\end{tabular}}
\end{center}

{All 192 below-threshold cases are structurally excluded
from full rank. At $m=t$, each of the 64 combinations $(v,d,r,N,n)$
has at least one seed reaching full rank at all three tolerances.
The two primary deficits occur at $v=1$.
For $(d,r,N,n,m)=(4,1,2,2,4)$, seed 0 stops at rank $15/16$;
continuing the same fields and state through $k=5$ gives rank 16 only
at $\varepsilon=10^{-10}$.
For $(6,3,2,3,2)$, seed 1 reaches rank $35/36$ at $k=6$ and
full primary rank at $k=7$.
These separate checks do not replace the original counts.}

{\textbf{Larger dimension.}
A separate pilot fixes $D=96$, $d=3$, $r=1$, $m=3=t$, $v=1/d$,
and seed 0, using the same projections for
$(N,n)=(16,2),(8,4),(4,8)$.
At $k=5$, the respective rank triplets, in tolerance order
$(10^{-6},10^{-8},10^{-10})$, are
$(93,95,96)$, $(94,96,96)$, and $(96,96,96)$.
Continuing the first case through $k=6$ gives $(92,95,96)$;
relative numerical rank need not increase when columns are added.
Thus full rank is observed for all three shapes at $10^{-10}$,
but only two at the primary tolerance.}

{These results support the proposed threshold across
the tested batch sizes and sequence lengths, with sensitivity to
conditioning at the boundary. They do not establish a universal
interpolation theorem, which would also require control of the
connectedness of the full-rank domain.
The experiment is reproduced by \texttt{run\_threshold.py}, using
\texttt{lie\_rank.py} and the helpers in
\texttt{experiment\_utils.py}; saved results include rank histories,
singular values, and stopping reasons.}

\section{Proof of \texorpdfstring{\Cref{prop:causal-rank}}{the causal rank proposition}}
\label{app:causal-rank}

\begin{proof}
    Fix $X\in\mathcal{A}_{N,n,d}$.
    Every single-head causal field has first-token block $VX_1$;
    for a multi-head field, this block is $(\sum_{h=1}^H V_h)X_1$.
    More generally, if
    two fields have first-token blocks $AX_1$ and $CX_1$ for constant matrices
    $A,C$, their Lie bracket has first-token block $(CA-AC)X_1$.
    Thus both the direct span and all iterated brackets lie in the space
    on the right of \eqref{eq:causal-span}.

    It remains to show that all directions in this space are attained.
    We adapt the argument from the proof of \Cref{thm:attention-uit}
    in \Cref{app:attention-proof} to causal prefixes.
    Define the prefix barycenters
    \[
        z_{\ell,i}(b)
        \coloneqq
        \frac{\sum_{j=1}^i e^{\langle b,x_j^\ell\rangle}x_j^\ell}
        {\sum_{j=1}^i e^{\langle b,x_j^\ell\rangle}},
        \qquad
        F^{\mathrm c}_{W,V}(X)_{i,\ell}
        =Vz_{\ell,i}(Wx_i^\ell).
    \]
    Fix a target block $(i_0,\ell_0)$ with $i_0\geq2$, and put
    $q=x_{i_0}^{\ell_0}$.  Choose $a$ separating the keys in each sequence
    as in \eqref{eq:attention-separating-a}, and choose $\alpha\in q^\perp$
    nonorthogonal to every other token as in
    \eqref{eq:attention-separating-alpha}.  These choices are possible
    because the tokens are nonzero and pairwise noncollinear.  Set
    \[
        W_t^\pm=a\left(t\alpha\pm\frac{q}{\|q\|^2}\right)^\top.
    \]
    At the target block their difference gives $Vc$, where
    $c=z_{\ell_0,i_0}(a)-z_{\ell_0,i_0}(-a)$.  This vector is nonzero:
    we apply the argument in \Cref{app:attention-proof} to the function
    \[
        \psi(s)\coloneqq\log\sum_{j=1}^{i_0}
        e^{s\langle a,x_j^{\ell_0}\rangle}
    \]
    to obtain $\psi''(s)>0$, since $i_0\geq2$ and the numbers
    $\langle a,x_j^{\ell_0}\rangle$ are distinct.  Hence
    $\langle a,c\rangle=\psi'(1)-\psi'(-1)>0$.

    At any other query, the two score vectors tend along $a$ to the same
    signed infinity.  Both prefix barycenters therefore converge to the
    same unique maximizing or minimizing key.  At a first-token block
    they are already identical, since $z_{\ell,1}(b)=x_1^\ell$ for every
    $b$.  Hence
    \[
        F^{\mathrm c}_{W_t^+,V}(X)
        -F^{\mathrm c}_{W_t^-,V}(X)
        \longrightarrow E_{i_0,\ell_0}(Vc),
    \]
    where $E_{i_0,\ell_0}(u)$ has block $u$ at the target and zeros
    elsewhere.  The pointwise span $\mathcal{S}^{\mathrm c}_X$ is
    finite-dimensional and closed, so it contains this limit.
    Thus the limiting direction is a finite linear combination of
    field values with finite parameters.
    Taking $V=uc^\top/\|c\|^2$ shows that it contains
    $E_{i_0,\ell_0}(u)$ for every $u\in\mathbb{R}^d$.  Thus every
    later-token direction can be varied independently.

    Finally, write any $A\in\mathbb{R}^{d\times d}$ as
    $A=\sum_r V_r$ with $\operatorname{rank}V_r\leq1$.
    Then $\sum_r F^{\mathrm c}_{0,V_r}(X)$ has first-token block $AX_1$.
    The directions constructed above allow all later-token blocks to be
    adjusted arbitrarily without changing $AX_1$.  This proves
    \eqref{eq:causal-span}.  The image of $A\mapsto AX_1$ has dimension
    $d\,\operatorname{rank}X_1$, giving \eqref{eq:causal-rank}; the bracket
    upper bound is therefore sharp as well.  Any head of positive width
    realizes these rank-at-most-one fields, with the remaining heads
    set to zero, so the same span and rank hold for the multi-head family.

    For completeness, the first-token constraint also obstructs global
    interpolation on $\mathcal{A}_{N,n,d}$ when $N>d$.  Choose a nonzero
    relation $X_1c=0$ and an index $\ell$ with $c_\ell\neq0$.  Scaling only
    $x_1^\ell$ by two breaks this relation while staying in the same
    component of $\mathcal{A}_{N,n,d}$ along the positive scaling path.
    Every causal flow and finite causal residual network preserves the
    original relation, so neither reaches this target.
\end{proof}

\section{Proof of \texorpdfstring{\Cref{cor:causal-interpolation}}{the causal interpolation corollary}}
\label{app:causal-interpolation}

\begin{proof}
    Let $r=N-\dim K$.  Choose a full-row-rank matrix
    $Q\in\mathbb{R}^{r\times N}$ with $\ker Q=K$.  A first-token
    matrix $X_1$ annihilates $K$ if and only if it has the unique form $X_1=PQ$
    for $P\in\mathbb{R}^{d\times r}$.  Thus $E_K$ is a linear space of
    dimension $R=N(n-1)d+dr$.  Within $E_K$, the condition
    $\ker X_1=K$ is equivalent to $P$ being injective, since $Q$ is
    surjective onto $\mathbb{R}^r$.  Equivalently,
    $\operatorname{rank}P=r$, an open condition because a nonzero
    $r\times r$ minor remains nonzero under small perturbations.
    Since $\mathcal{A}_{N,n,d}$ is open, it follows that
    $\mathcal{A}^{K}_{N,n,d}$ is relatively open in $E_K$.
    Every point therefore has a small ball in $E_K$ contained in this
    set.  This ball is connected and hence lies in the same connected
    component as its center, so each component is relatively open.

    Every causal field maps $E_K$ into $E_K$: for $X\in E_K$ and
    $c\in K$, its first-token block satisfies
    $(V_\Sigma X_1)c=V_\Sigma(X_1c)=0$.
    Now fix $X\in\mathcal{A}^{K}_{N,n,d}$ and $Z\in E_K$.
    The rule $A(X_1u)=Z_1u$ defines a linear map on
    $\operatorname{im}X_1$: if $X_1u=X_1v$, then
    $u-v\in\ker X_1=K$, so $Z_1u=Z_1v$.
    Extending $A$ linearly to $\mathbb{R}^d$ gives $Z_1=AX_1$.
    Thus every $Z\in E_K$ satisfies the condition in
    \eqref{eq:causal-span}, and \Cref{prop:causal-rank} yields
    \[
        \operatorname{span}\{F^{\mathrm c}_{W,V}(X):
        \operatorname{rank}W,\operatorname{rank}V\leq1\}=E_K.
    \]
    Choose linear coordinates identifying $E_K$ with $\mathbb{R}^R$.
    The restricted fields are analytic and closed under negation by
    replacing $V$ with $-V$.  On each connected component,
    \Cref{cor:pointwise-full-rank-criterion} gives the continuous-depth
    conclusion on a horizon $[0,S]$ depending on the endpoints.
    Closure under real scalar multiplication by scaling $V$ also
    permits applying \Cref{prop:finite-depth-interpolation} to obtain
    the finite residual network.  Both constructions stay in that component.
    The rank-at-most-one factorizations give one head of width one;
    additional prescribed heads can be set to zero.
\end{proof}

\section{Proof of \texorpdfstring{\Cref{cor:causal-random-dictionary}}{the universal causal interpolation corollary}}
\label{app:causal-random-dictionary}

\begin{proof}
    Fix $K$ as in the statement and a linear isomorphism
    $T:E_K\to\mathbb{R}^R$.  In these coordinates, define
    \[
        G(\theta,z)
        \coloneqq T (F^{\mathrm c}_{ab^\top,uv^\top}(T^{-1}z)),
        \qquad \theta=(a,b,u,v)\in\mathbb{R}^{4d}.
    \]
    The domain $T(\mathcal{A}^{K}_{N,n,d})$ is a nonempty open subset
    of $\mathbb R^R$.  The function $G$ is jointly real analytic in
    $(\theta,z)$, since $T$ is linear and
    all prefix softmax denominators are strictly positive.  Its values
    span $\mathbb{R}^R$ at every configuration by \Cref{prop:causal-rank}.
    Apply \Cref{thm:uniform-random-spanning-principle} with parameter
    dimension $4d$ and with both configuration and output dimensions
    equal to $R$.  It follows that $2R$ independent Gaussian-factor
    samples span simultaneously throughout this domain with probability
    one.  The same event covers all its connected components, for this
    fixed $K$.

    Fix a draw on this event and consider the signed family
    \[
        \{\pm F^{\mathrm c}_{W_j,V_j}:1\leq j\leq2R\}.
    \]
    This family is closed under negation and has
    pointwise full rank in $E_K$ at every point of the class.  Applying
    \Cref{cor:pointwise-full-rank-criterion} in the coordinates $T$ on each
    connected component gives exact interpolation on a horizon $[0,S]$
    depending on the endpoints, within that component.  Each control
    segment selects one of the same $2R$ sampled generators and its
    sign, so this is the
    relative universal interpolation property of
    \Cref{def:universal-interpolation}.  The signed dictionary contains
    at most $4R$ fields, all realized by the same architecture since
    $-F^{\mathrm c}_{W_j,V_j}=F^{\mathrm c}_{W_j,-V_j}$.

    For finite depth, apply \Cref{prop:finite-depth-interpolation} to
    the scalar-closed family
    $\{cF^{\mathrm c}_{W_j,V_j}:c\in\mathbb R,\ 1\leq j\leq2R\}$.
    A sufficiently fine Euler discretization of an interpolating trajectory
    first gives a network of $L_0$ residual layers, where $L_0$ is finite
    and may depend on the interpolation problem.  Appending at most $R$
    correction layers then yields exact interpolation, with total depth
    at most $L_0+R$ and scalar gains shared across the batch.
    Linearity in $V$ preserves
    the rank-at-most-one realization.  The sampled matrices are fixed
    before the endpoints are chosen, proving universal interpolation
    at finite depth with the same $2R$ fields.

\end{proof}


\end{document}